%% file: neurips_2024.tex
\documentclass{article}

    \usepackage[final]{neurips_2024}

\usepackage[utf8]{inputenc} 
\usepackage[T1]{fontenc}    
\usepackage{hyperref}       
\hypersetup{
	colorlinks=true,       
	linkcolor=blue,        
	citecolor=blue,        
	filecolor=magenta,     
	urlcolor=blue         
}
\usepackage{url}            
\usepackage{booktabs}       
\usepackage{amsfonts}       
\usepackage{nicefrac}       
\usepackage{microtype}      
\usepackage{xcolor}         
\usepackage{xspace}  
\usepackage{graphicx}
\usepackage{amsmath}
\usepackage{amssymb}
\usepackage{mathtools}
\usepackage{amsthm}
\usepackage{listings}
\usepackage{natbib}
\usepackage{enumitem}
\usepackage{silence}
\usepackage[capitalize,noabbrev]{cleveref}
\usepackage[textsize=tiny]{todonotes}
\usepackage{indentfirst} 
\usepackage{microtype}
\usepackage{graphicx}
\usepackage{subfigure}
\usepackage{wrapfig}
\usepackage{booktabs} 
\usepackage{hyperref}
\usepackage{multirow}
\usepackage{makecell}
\usepackage{subcaption}
\usepackage{etoc}

\newcommand{\method}{CultureLLM\xspace}
\newcommand{\llms}{LLMs\xspace}
\newcommand{\prompt}[1]{{\footnotesize \ttfamily #1}\xspace}

\title{CultureLLM: Incorporating Cultural Differences into Large Language Models}

\author{%
  Cheng Li\thanks{Work done during Cheng's internship at MSRA.} \\
  Institute of Software, CAS\\
  \texttt{chenglicat0228@gmail.com} \\
  \And
  Mengzhuo Chen \\
  Institute of Software, CAS \\
  \texttt{mengzhuo.happy@gmail.com} \\
  \And
  Jindong Wang\thanks{Corresponding author.} \\
  Microsoft Research \\
  \texttt{jindong.wang@microsoft.com} \\
  \And
  Sunayana Sitaram \\
  Microsoft Research \\
  \texttt{Sunayana.Sitaram@microsoft.com} \\
  \And
  Xing Xie \\
  Microsoft Research \\
  \texttt{xing.xie@microsoft.com} \\
}

\begin{document}
\etocdepthtag.toc{mtchapter}
\etocsettagdepth{mtchapter}{none}
\etocsettagdepth{mtappendix}{none}

\maketitle

\begin{abstract}
Large language models (LLMs) have been observed to exhibit bias towards certain cultures due to the predominance of training data obtained from English corpora. Considering that multilingual cultural data is often expensive to procure, existing methodologies address this challenge through prompt engineering or culture-specific pre-training. However, these strategies may neglect the knowledge deficiency of low-resource cultures and necessitate substantial computing resources. In this paper, we propose \textbf{\method}, a cost-effective solution to integrate cultural differences into \llms. \method employs the World Value Survey (WVS) as seed data and generates semantically equivalent training data through the proposed semantic data augmentation. Utilizing only $50$ seed samples from WVS with augmented data, we fine-tune culture-specific \llms as well as a unified model (\method-One) for $9$ cultures, encompassing both rich and low-resource languages. Extensive experiments conducted on $60$ culture-related datasets reveal that \method significantly surpasses various counterparts such as GPT-3.5 (by $8.1$\%) and Gemini Pro (by $9.5$\%), demonstrating performance comparable to or exceeding that of GPT-4. Our human study indicates that the generated samples maintain semantic equivalence to the original samples, offering an effective solution for \llms augmentation. Code is released at \url{https://github.com/Scarelette/CultureLLM}.
\end{abstract}

\section{Introduction}
\label{sec-intro}


Culture is a complex construct that encapsulates various identities, including, but not limited to, language, nationality, region, religion, and gender identity.
Cultural bias is prevalent worldwide and refers to the tendency to favor specific cultural perspectives, values, and norms, which results in subjective opinions and may offend individuals from other cultures.
For instance, according to the World Value Survey~\citep{wvs}, Arabic culture believes that men are better political leaders than women, while people in the United States maintain a contrary opinion.\footnote{We respect all opinions in different cultures.}
As large language models (\llms)~\citep{openai2023gpt4,gemini} gain prominence, they are reported to exhibit cultural bias and specifically show partiality towards Western culture, as English corpora dominate the training data~\citep{liu2023multilingual,cao2023assessing,masoud2023cultural,naous2023having,wang2023not,johnsonghost}.
Low-resource cultures are frequently underrepresented due to the insufficient training data available from these cultures.
\llms' cultural bias constitutes a significant bottleneck in human-AI collaboration and considerably impedes AI democracy.

Tackling cultural bias necessitates that a large language model (LLM) acknowledges cultural differences~\citep{hofstede1984culture}.
Kovavc et al. \cite{kovavc2023large} and Want et al. \cite{wang2023not} thought \llms have enough knowledge of all cultures and devised prompt engineering technologies to induce \llms to exhibit specific cultural perspectives.
However, they are not effective, especially in low-resource cultures with limited data.
Another line of work pre-trained culturally aware \llms and then fine-tuned on specific datasets~\citep{chan2023harmonizing,nguyen2023seallms,pipatanakul2023typhoon,abbasi2023persianllama,lin2023taiwan}.
They require the collection of large-scale pre-training and fine-tuning datasets and extensive computing resources, thus are not affordable to ordinary researchers and cannot handle low-resource culture.
To date, training culturally aware \llms at affordable costs remains a challenge.

\begin{wrapfigure}{r}{0.5\textwidth}
  \begin{center}
  \vspace{-.2in}
    \includegraphics[width=.5\textwidth]{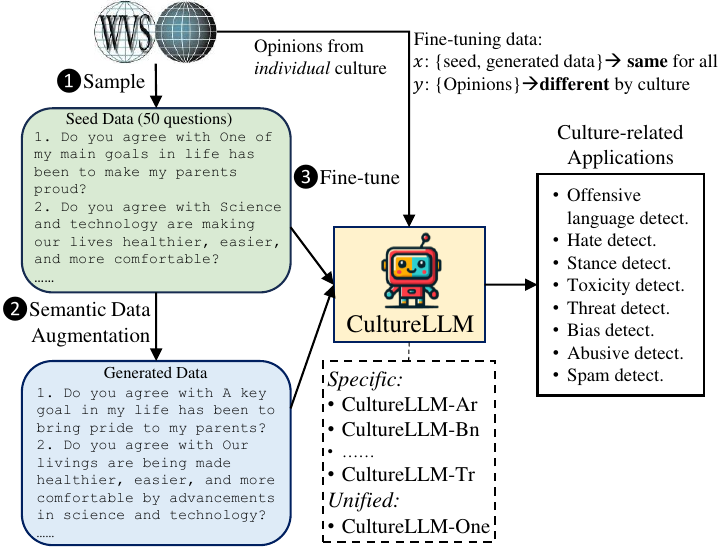}
  \end{center}
  \vspace{-.1in}
  \caption{Overview of \method. \method consists of three steps: sampling, semantic data augmentation, and fine-tuning. Culture-specific and unified \method can be fine-tuned.}
    \label{fig-overview}
\end{wrapfigure}

In this paper, we propose \textbf{\method}, a cost-effective\footnote{Fine-tuning a \method only costs \$6 via OpenAI API.} solution to incorporate cultural differences into \llms.
Technically speaking, \method is inspired by the well-known fact that LLMs are inevitably not robust to the style and format of the prompts~\citep{zhu2023promptbench}, indicating that we can further leverage such a weakness to further improve the performance of LLMs by enriching the prompts.
In particular, we focus on cultural values in this work.
As shown in \Cref{fig-overview}, \method consists of three steps: sampling, semantic data augmentation, and fine-tuning.
Inspired by Attitude-Behavior Consistency theory~\citep{fazio1981direct} which emphasizes that people's opinion is consistent with their behaviors, we use the World Values Survey (WVS)~\citep{wvs} as seed data.
Then, we devise a semantic data augmentation approach to generate semantically equivalent samples. The aim is to generate semantic equivalent inputs, thus we can get the ground-truth from seed data directly.
Finally, \method is obtained by fine-tuning both the seed and the generated data.
WVS is a public opinion poll that contains people's opinions on cultural topics from different countries.
To be specific, we select $50$ seed samples from WVS, covering $7$ topics: ``social values", ``migration", ``security", ``science and technology", ``religious values", ``ethical values and norms", and ``political interest and political participation".
Using these generated samples and answers from people in different cultures, we fine-tune specific and unified \llms: specific \llms are tailored for each culture such as \method-Ar for Arabic and \method-Tr for Turkish; unified \llms (\method-One) are one LLM that fits all cultures.\footnote{The bound of culture is unclear; we use the main spoken language to distinguish cultures~\citep{delanoy_2020}.}

We build $9$ specific \method and a \method-One covering both high- and low-resource cultures: Arabic culture, Bengali culture, Chinese culture, English culture, German culture, Korean culture, Portuguese culture, Spanish culture, and Turkish culture.
Then we evaluated them on $8$ culture-related downstream tasks: offensive language detection, hate speech detection, stance detection, toxicity detection, threat detection, bias detection, abusive detection, spam detection, and an open-ended generative task.
We have $60$ test sets of $68,672$ samples in total.
Experiments show that \method fine-tuned on GPT-3.5 significantly outperforms GPT-3.5 by $8.1$\% and outperforms Gemini pro~\citep{gemini} by $9.5$\% on average F1 score, achieving comparable or even better performance with GPT-4.
Our human study of $50$ people demonstrates that the augmentation method can generate semantically equivalent samples.
We further interpret the rationale behind its effectiveness by exploring the fine-tuning data size and case studies.
Finally, results on Big-Bench Hard~\citep{suzgun2022challenging} and GSM8K~\citep{cobbe2021training} indicate that \method is resistant to catastrophic forgetting.
\method also supports fine-tuning \llms of open-source models.

Our contributions are three-fold:
\begin{enumerate}[leftmargin=1em]
\setlength\itemsep{0em}
    \item We presented \method, a cost-effective fine-tuning solution to build culturally-aware \llms.
    \item We proposed semantic data augmentation, an augmentation approach to generate high-quality and diverse training data for \llms.
    \item We conducted extensive experiments in a wide range of cultures and \llms, showing that \llms performs consistently well in all downstream tasks.
\end{enumerate}

\section{Related Work}
\label{sec-related}

\subsection{Cultural Problem and Solution in \llms}

Previous efforts have shown that \llms exhibit the same cultural problems as in human society.
\citet{niszczota2023large} proved that GPT-4 can replicate the cross-cultural differences for each personality factor through large-scale experiments.
Meanwhile, other works also found that \llms can reflect cultural bias and dominance in human society~\citep{liu2023multilingual,cao2023assessing,masoud2023cultural,naous2023having,wang2023not,johnsonghost}, e.g., Western culture dominance, since the major training corpus such as Pile~\citep{gao2020pile} is in English.

The ideal solution is to improve the cultural awareness of \llms.
There are mainly two types of approach: prompt engineering and pre-training.
\citet{kovavc2023large,wang2023not} thought \llms as superpositions of cultural perspectives, which can be prompted to targeted cultural perspectives. while \citet{rao2023ethical} encoded cultural values in the prompts.
Although PE is cheap, its effectiveness is challenged, especially in low-resource cultures where \llms lack such cultural knowledge due to lack of representation in pre-training data.
Another line of research is pre-training and fine-tuning~\citep{chan2023harmonizing,nguyen2023seallms,pipatanakul2023typhoon,abbasi2023persianllama,lin2023taiwan} that trains culturally-aware \llms for different cultures by collecting large-scale pre-training datasets and then performed fine-tuning for better alignment.
While they achieved great performance, this approach is too expensive and time-consuming, thus it is difficult to apply to more cultures and countries.
They still suffer from a low-resource culture problem where the pre-training data are difficult to collect. MaLA-500~\citep{lin2024mala500} trained a new LLM on Llama 2 to cover $534$ languages, which is resource intensive.


\subsection{Data Augmentation for \llms}

Human-annotated data are high-quality but expensive.
Due to the strong generation ability of \llms, many works focused on leveraging data augmentation for \llms.
\citet{yu2023metamath,liu2023tinygsm} used \llms to augment the math data and then fine-tuned with those data.
\citet{li2023self} synthesized data with two designed modules: self-augmentation and self-curation.
\citet{chen2024self} introduced a self-play mechanism, where LLM generates its own training data from its previous iterations, refining its policy by discerning these self-generated responses from those obtained from human-annotated data. There are also other uses for synthetic data, such as knowledge distillation~\citep{wang2023let} and improving text embedding tasks~\citep{wang2023improving}.
Our data augmentation approach also adopts \llms for data generation, but we add controllable modules such as template editing, synonym replacement, and semantic filter to ensure the diversity and semantic equivalence of the generated samples.
It can also be used as a general augmentation method in other applications.

Efforts in cultural datasets~\citep{nguyen2023extracting,fung2022normsage} focus on cultural common sense and norms.
However, they generate data from only the English or Chinese corpus and thus may contain cultural bias toward other cultures. In contrast, World Values Survey (WVS)~\citep{wvs} is a large-scale pool that contains answers from people of different cultures, thus providing more objective cultural values from specific cultures.

This work is also related to value alignment~\citep{ji2023ai,shen2023large,yao2023instructions} to align the values of \llms with human's by designing algorithms for value measurement and behavior alignment.
In contrast, this work primarily emphasizes value understanding with the potential to be extended for value alignment.
For instance, semantic augmentation can be used to generate training data for alignment-related tasks.

\section{\method}
\label{sec-method}


\subsection{Overview}

Cultural differences are prevalent in various cultures and backgrounds, leading to an impact on outcomes in downstream applications such as hate speech and biased language.
To address the gap between low-source cultural data collection and its wide applications, we design \method by fine-tuning an LLM on data generated by our novel semantic data augmentation approach leveraging the sensitivity of \llms on prompts~\citep{zhu2023promptbench}.
\Cref{fig-overview} presents an overview of \method, where the first step is to sample a subset of data from an existing World Value Survey (WVS)~\citep{wvs} that represents different opinions (answers) towards the same value questions given by native users.
The adoption of WVS is inspired by Attitude-Behavior Consistency theory~\citep{fazio1981direct}, which emphasizes the strong relationship between attitude and behavior.
Therefore, WVS serves as an ideal seed for data augmentation.\footnote{WVS is one feasible option for seed and other surveys can also be used. But the cultural survey data are extremely rare and the WVS could be the most comprehensive one. More details about WVS are in \Cref{append-seeddata}.}
After sampling, the second step is to generate augmented data using our proposed semantic augmentation approach (\Cref{sec-sub-aug}) and then fine-tune a \method for each specific culture such as \method-Ar for Arabic culture and \method-Tr for Turkish culture.

Generally speaking, we use $\mathcal{D}_d=\{(x_j, y_j^d)\}_{j=1}^{n}$ to denote the seed and $\mathcal{D}^\prime_d=g(\mathcal{D}_d)=\{(x^\prime_j, y_j^d)\}_{j=1}^{n^\prime}$ as the augmented data with $g(\cdot)$ the augmentation algorithm.
Note that the question $x$ here is the \emph{same} in all cultures in WVS and $d$ is the cultural index denoting \emph{different} answers to the same question $x$.
For example, for a question $x$=``\prompt{Do you agree with on the whole, men make better political leaders than women do}?'', the answer $y=\text{\prompt{Disagree}}$ if $d=\text{\prompt{English}}$; and $y=\text{\prompt{Strongly agree}}$ if $d=\text{\prompt{Arabic}}$.
Therefore, we only augment the question $x$ to be $x^\prime$ but retain the same opinion $y$ as the original $x$.
We also denote vanilla LLM and \method as $f$ and $f^\star$, respectively.
Then, denoting $\ell$ as the loss function, our learning objective is formulated as: $
    f^\star=\arg \min_{f}\mathbb{E}_{\substack{(x_j, y_j^d) \in \{\mathcal{D}_d, g(\mathcal{D}_d)\}}}[\ell(f(x_j),y_j^d)].$

\subsection{Sampling}
\label{sec-sub-sample}


\begin{wrapfigure}{r}{0.5\textwidth}
  \begin{center}
  \vspace{-.2in}
    \includegraphics[width=.5\textwidth]{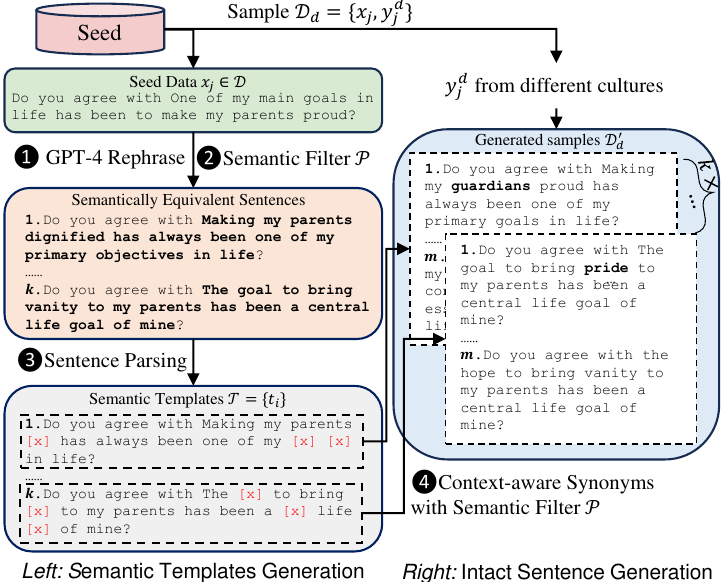}
  \end{center}
  \caption{Details of semantic data augmentation. First, semantic templates are generated via rephrasing, semantic filtering, and sentence parsing. Then, training samples are generated by context-aware synonyms replacement and semantic filtering.}
  \vspace{-.3in}
    \label{fig-dataAug}
\end{wrapfigure}

The sampling process should follow two principles: 1) cover as many cultural topics as possible and 2) sample questions that can be clearly answered by \llms.
Based on the two principles, we manually select $n=50$ questions and rewrite them in the Question-Answer (QA) format, covering $7$ topics, namely social values, security, science and technology, religious values, ethical values and norms, political interest and political participation, and migration. The details can be found in \Cref{append-seeddata}.


\subsection{Semantic Data Augmentation}
\label{sec-sub-aug}


Samples from WVS are not enough to fine-tune, which can be augmented by our semantic augmentation approach.
In a formal sense, semantic augmentation retains the original ground-truth opinions ($y_d$) from different cultures and only generates semantically equivalent questions ($x$).
A naive augmentation approach is to directly use strong \llms such as GPT-4 to generate new samples \citep{walters2023fabrication}, which could introduce mode collapse, as generation quality can only be controlled by prompts.
Furthermore, since LLMs could suffer from cultural bias, directly generating cultural data using prompts could lead to unexpected or even erroneous results.

As shown in \Cref{fig-dataAug}, the augmentation consists of two stages: semantic template generation and intact sentence generation.
The first stage generates several semantically equivalent but stylistically different sentences and parses them into semantic templates.
The second stage then generates samples by replacing certain words in the semantic templates.
Such an augmentation can naturally introduce more diversities: The first stage increases sentence-level diversities and the second improves the word-level diversities.

\paragraph{Semantic Template Generation}
This stage generates semantically equivalent question templates $\mathcal{T}=\{t_i\}_{i=1}^{k}$ based on $x \in \mathcal{D}_d$.
The generation process is nontrivial since there are two challenges ahead: 1) the naturalness and diversities and 2) the semantic preservation.
We solve the first challenge by using GPT-4 as the generator with certain prompts to ensure naturalness and diversity. 
Then, we solve the second challenge by introducing a semantic preservation filter $\mathcal{P}$ to measure the similarity between the original and generated sentences.\footnote{For computational efficiency, we use BERT embedding as $\mathcal{P}$ while other models can also be used.}

We first use the prompt ``\prompt{Could you please generate [$n$] sentences that (1) have different sentence structures and (2) have the same meaning with the following sentence: {$x_i$}}'' to generate $n$ sentences using GPT-4.
Then, we denote the embedding of the original sentence and the generated sentences as $z=\mathcal{P}(x)$ and $z^\prime=\mathcal{P}(t)$, respectively. Then we compute their similarity score $c=\cos(z,z^,)$.
If $c$ passes the threshold value $\tau$, the generated sentence will be reserved:$
    \mathcal{T}=\{t_i | \cos(\mathcal{P}(t_i), \mathcal{P}(x_j))>\tau\}, \forall x_j \in \mathcal{D}_d.$
Specifically, for sample ``\prompt{Do you agree with One of my main goals in life has been to make my parents proud}?'', we generate $m$ samples using GPT-4, which are then go through the semantic filter $\mathcal{P}$ to eventually retain $k (k \le m)$ semantically equivalent sentences, e.g. ``\prompt{Do you agree with Making my parents dignified has always been one of my primary objectives in life}?" and ``\prompt{Do you agree with The goal to bring vanity to my parents has been a central life goal of mine}?" 

  
  



To diversify the generated data, we parse the $n$ sentences to find the appropriate components to replace, which construct the templates.
For efficiency, we use NLTK~\citep{loper2002nltk} to find replaceable words, such as adjectives, adverbs, nouns, and verbs.
The semantic templates are like ``\prompt{Do you agree with The [x] to bring [x] to my parents has been a [x] life [x] of mine}?" where ``\prompt{[x]}'' is the replaceable part.
In total, we generate $k$ templates for each sample $x_j \in \mathcal{D}_d$.

\paragraph{Intact Sample Generation}
\label{sec-intact-gen}

This step is to replace synonyms in templates to generate fine-tuning samples.
We apply GPT-4 to generate context-aware synonyms in the templates and randomly replace some of them.
To further preserve semantics, we also use the semantic preservation filter.
After filtering, we generate $m$ samples for each template $t_i \in \mathcal{T}$, and get $n^\prime = mnk$ samples for all $x_j \in \mathcal{D}_d$ in total.
For example, intact samples for template ``\prompt{Do you agree with The [x] to bring [x] to my parents has been a [x] life [x] of mine}?" could be ``\prompt{Do you agree with The goal to bring pride to my parents has been a central life goal of mine}?" and ``\prompt{Do you agree with The hope to bring vanity to my parents has been a central life goal of mine}?"
Our human study in \Cref{sec-dis-human} shows that augmentation can generate high-quality and semantically equivalent sentences.

  
  
  
  

\subsection{Fine-tuning}
\label{sec-sub-ansAug}


Since culture is a complex construct, we use languages spoken by geographical cultures (represented by countries) in WVS to represent broader cultures and arrive at a set of $9$ cultures in total. In cases where a language is spoken by more than one geographical culture, we pick representative countries and use the average of all answers as groundtruth. Our final set of cultures represented as described above is Arabic (for which we select Jordan and Iraq), Spanish (for which we select Mexico and Argentina), Bengali, Chinese, English, German, Korean, Portuguese, and Turkish.



Finally, \method is obtained by fine-tuning an LLM on the combination of the seed and the generated data.
Specifically, we fine-tune two types of LLMs: 1) culture-specific LLMs for each language such as \method-Ar and \method-Bn, and 2) one unified LLM for all languages, denoted as \method-One.
Culture-specific LLMs are tailored by setting $d$, namely, $\{\mathcal{D}_{d}, \mathcal{D}^\prime_d\}$.
On the other hand, \method-One is trained on all datasets: $\{\mathcal{D}_d, \mathcal{D}^\prime_d\}_{d \in \text{all language}}$ to serve as a unified LLM for all cultures.
Note that since all languages have the same input question $x$ but different answers $y$, we need to manually write different prompts in the instruction to distinguish them.
For example, we add ``\prompt{You are an Arabic chatbot that knows Arabic very well}" before Arabic samples.
\method can be used in cultural downstream applications.
In the following, we use \method to denote specific LLM and \method-One for unified LLM.

\textbf{Remark:} Note that the WVS is all in English, where we focus on cultural differences in \emph{opinions} regardless of their native language.
Thus, we do not perform fine-tuning for other languages due to the shortage of their training data and rely on cross-lingual transfer.
Multilingual tasks for cultures can still benefit from fine-tuned models in English, as models can learn the basic values from the opinions~\citep{moussaid2013social,jin2023better}.
Our experiments in \Cref{sec-dis-multilingula} further demonstrate that fine-tuning on English data can outperform fine-tuning on native data that are translated from the original English version.




\section{Experiments}
\label{sec-exp}

We fine-tuned a \method-One and $9$ specific \method for $9$ languages: Arabic (Ar), Bengali (Bn), Chinese (Zh), English (En, United States), German (De), Korean (Ko), Portuguese (Pt), Spanish (Es), and Turkish (Tr).
These cultures are diverse and represent both high- and low-resource regimes and thus can serve as representative evaluation.

\subsection{Setup}
\label{setup}

\textbf{Datasets.}
We adopt culture-related public datasets in specific languages for evaluation.
In total, we have $59$ test sets, covering $9$ languages and containing $68,607$ test samples.
We test on $56$ binary classification and $3$ multi-classification tasks to detect: offensive language, hate speech, stance, toxicity, threat, bias, abusive, and spam in zero-shot evaluation.
For example, we ask \llms to judge whether the sentence contains offensive language, hate speech, or biased speech.
Details are shown in \Cref{sec-append-dataset}.
Furthermore, we generate an open-ended generation dataset for evaluation in \Cref{sec-exp-generation}.

\textbf{Baselines and details.}
We fine-tune \method using the GPT-3.5 (0613)~\citep{chatgpt} fine-tuning API due to its efficiency and compared with two state-of-the-art \llms, namely Gemini pro~\citep{gemini} and GPT-4 (1104)~\citep{openai2023gpt4}.
We further compare with cultural specific pre-trained models SeaLLM~\citep{nguyen2023seallms}, TaiwanLLM~\citep{lin2023taiwan} and CultureBank~\citep{shi2024culturebank}.
We also compare this with retrieval augmentation (RAG), which enhances \llms by searching for related information and adding it to context~\citep{lewis2020retrieval}.
To implement RAG, we search for information about each culture on Wikipedia and append them in a system prompt, as detailed in \Cref{sec-append-prompt-rag}.
Finally, we fine-tuned \method using Llama-2-70b-chat~\citep{touvron2023llama} as the base model for reproduction (\Cref{sec-dis-llama}).
As for prompt setup, since our goal is to make \llms better align with people from different cultures, we add a system prompt ``\prompt{You are an [x] chatbot that knows [x] very well}'' where \prompt{[x]} is a certain language before each input. 
For metrics, we use macro F1 score for all tasks except for CValues~\citep{xu2023cvalues} where we use the automatic evaluation script provided by the paper.
For data augmentation, we set $k=5$, $m=2$, and $\tau=0.8$.
Evaluation prompts are in \Cref{sec-append-promptevaluation}.



\begin{figure}[t!]
    \centering
    \subfigure[Main results. \label{fig-main-results}]{
        \includegraphics[width=0.48\linewidth]{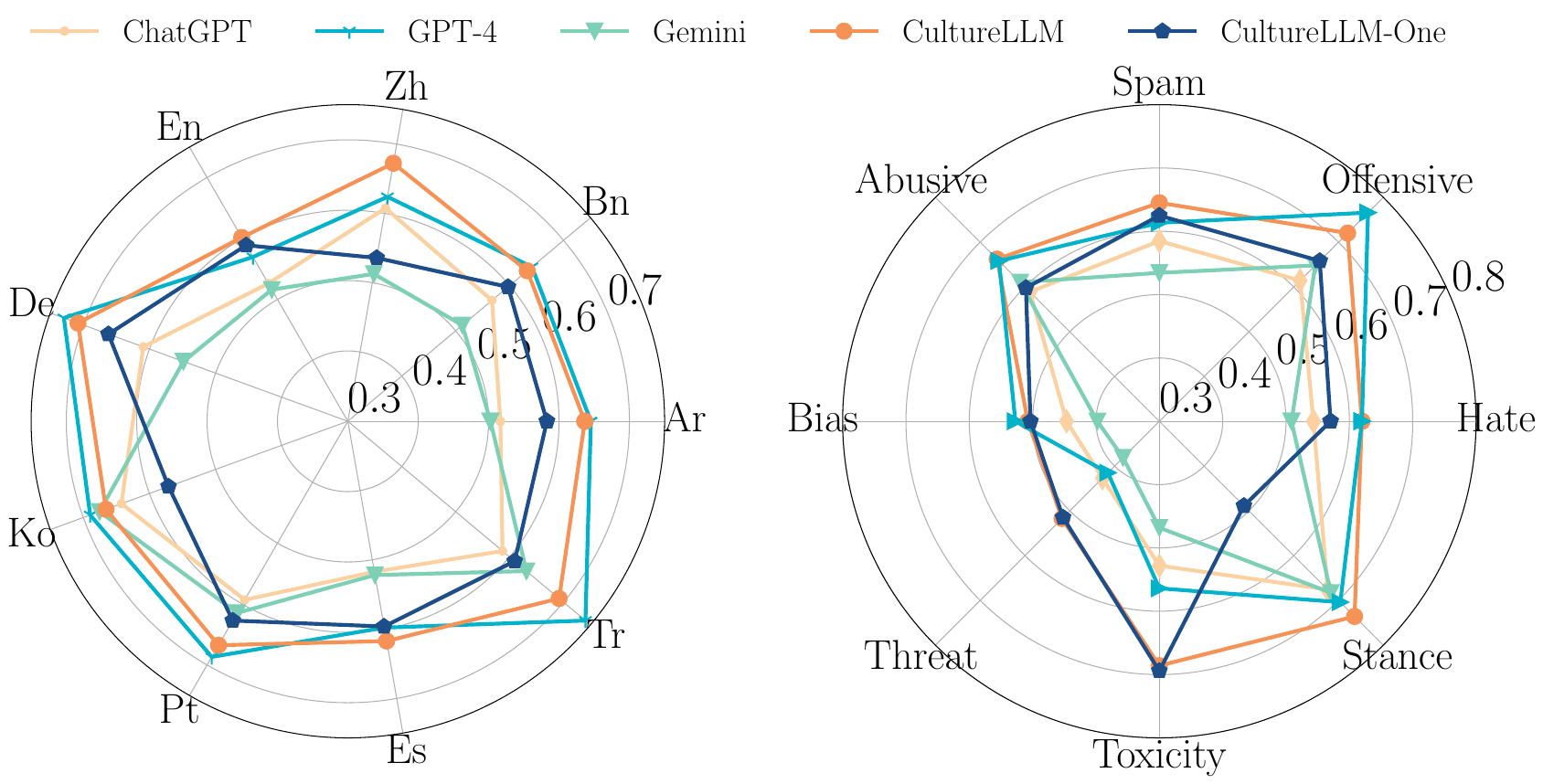}
    }
    \subfigure[Ablation study. \label{fig-ablation-study-data-aug}]{
        \includegraphics[width=0.48\linewidth]{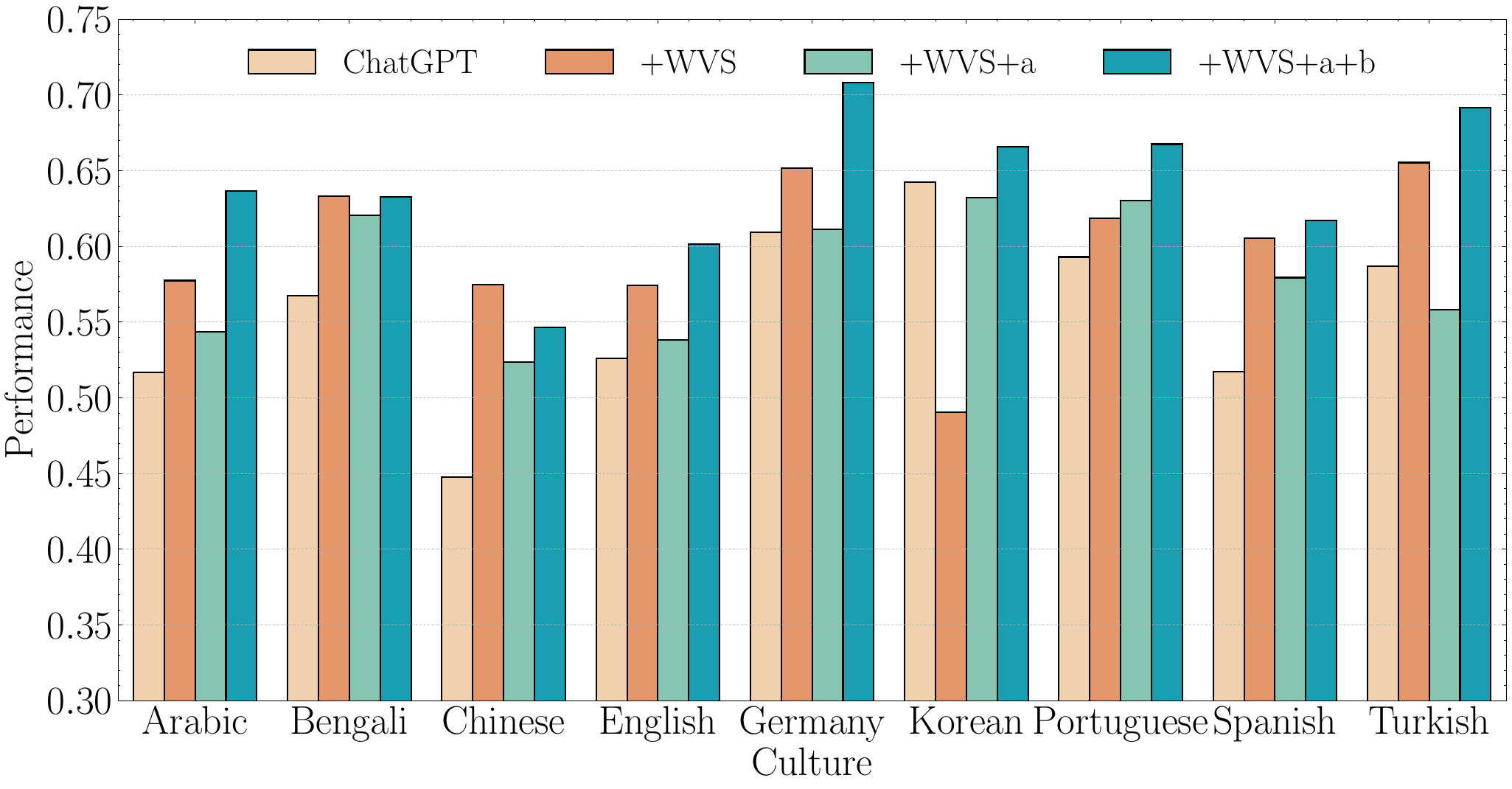}
    }
    \vspace{-.1in}
    \caption{(a) The main results averaged by cultures (left) and by tasks (right). Both \method and \method-One significantly outperform \method and Gemini with \method achieving the best performance comparable to GPT-4. (b) Ablation study. `+WVS' denotes the fine-tuned models using only the 50 samples from WVS, `+WVS+a' denotes fine-tuning using the WVS samples and the generated samples in step 1 of our data augmentation (i.e., using only GPT-4 to generate), and `+WVS+a+b' denotes the complete process of our algorithm. }
    \label{fig-main-ablation}
    \vspace{-.1in}
\end{figure}

\subsection{Main Results}

We present the average results for each culture and task in \Cref{fig-main-results} \footnote{Results on RAG are not shown since they are close to GPT-3.5. Since the metrics are not the same (e.g., accuracy for CValues and F1 for other tasks), we normalized each one and then averaged them.} and more detailed results are shown in Appendix~\ref{sec-append-exp}.
Our conclusions are as follows.
First, both specific and unified \method achieve a great improvement over other approaches, and specific \method achieves the best performance.
Concretely speaking, \method significantly outperforms GPT-3.5 (by $8.1$\%), Gemini (by $9.5$\%), and RAG (by $7.94$\%), achieving performance comparable to GPT-4 and even better on some tasks.
Second, \method-One exceeds GPT-3.5 by more than $4\%$ on $59$ tasks, while inferior to culture-specific models, suggesting that a single LLM might not be the best solution to solve cultural tasks with low resources, since data from different cultures could intertwine with each other.
Third, in terms of cultures, \method achieves the best performance in English, Chinese, and Spanish cultures while showing no obvious improvement in Korean culture, where all four models have a similar performance.
We infer that the reason could be that these base models have less exposure to Korean culture.

Then, we analyze the performance on both low-resource and high-resource language tasks.
As shown in \Cref{fig-resource-results}, \method shows excellent performance in both types of tasks and outperforms GPT-4 on a large scale in high-resource tasks.
Finally, we evaluated an extremely low-resource culture, Greek, in \Cref{sec-append-other-lowresource} and compared \method with other cultural-specific models SeaLLM~\citep{nguyen2023seallms}, TaiwanLLM~\citep{lin2023taiwan} and CultureBank~\citep{shi2024culturebank} in \Cref{sec-append-other-model}, which shows that our \method can also improve performance.
The correlation with the WVS data is in \Cref{sec-append-analysis-seed}, showing that the performance improvement does not come from the seed data in WVS.

\subsection{Results on Open-ended Generation Tasks}
\label{sec-exp-generation}

\input{tables/tb-gen-results}

To evaluate the performance of \method on open-ended tasks, we construct a dataset using GPT-4, containing $65$ open-ended questions, which cover the seven topics in WVS. The prompt setting for dataset generation can be found in \Cref{sec-append-prompt}.
We evaluated the outputs of GPT-3.5 and \method using Gemini Pro\footnote{We do not use GPT-4 to judge because it may prefer the response from GPT series models.}. We also devised a metric $\text{WinRate}=(s_{\method}-s_{ChatGPT})/65$, where $s$ represents the number of acceptances by Gemini Pro. Positive $\text{WinRate}$ means \method wins GPT-3.5 and vice versa.
As shown in \Cref{tb-gen-results}, \method performs better than GPT-3.5 on $8$ out of $9$ cultures, demonstrating its effectiveness in generation tasks.



\subsection{Ablation Study}

We evaluate the effectiveness of our semantic data augmentation approach by comparing it with the following variants: GPT-3.5, \method (WVS), \method (WVS+a), and \method (WVS+a+b), where \method (WVS) denotes the fine-tuned models using only the 50 samples from WVS, \method (WVS+a) denotes fine-tuning using 50 WVS samples and the generated samples in step 1 of our data augmentation (i.e., only using semantic templates), and \method (WVS+a+b) denotes the complete process of our algorithm.
Note that `WVS+a' denotes the naive baseline of only using GPT-4 to generate samples.

\Cref{fig-ablation-study-data-aug} shows that fine-tuning using the 50 seeds in WVS can inconsistently improve and impair performance on different tasks such as the decrease in Korean tasks.
While WVS data are of high quality, we see gains with our generated data which leads to improvements on most tasks. The average performance on Korean tasks also improves by \method.
\Cref{fig-ablation-study-data-aug} also demonstrates that the two steps in our semantic data augmentation approach are useful and necessary.
Ablation for \method-One is in \Cref{sec-append-ablation-one}, which also shows the effectiveness of our approach.

\begin{figure}[t!]
    \centering
    \subfigure[Influence of language. \label{fig-finetune-data}]{
        \includegraphics[width=0.48\linewidth]{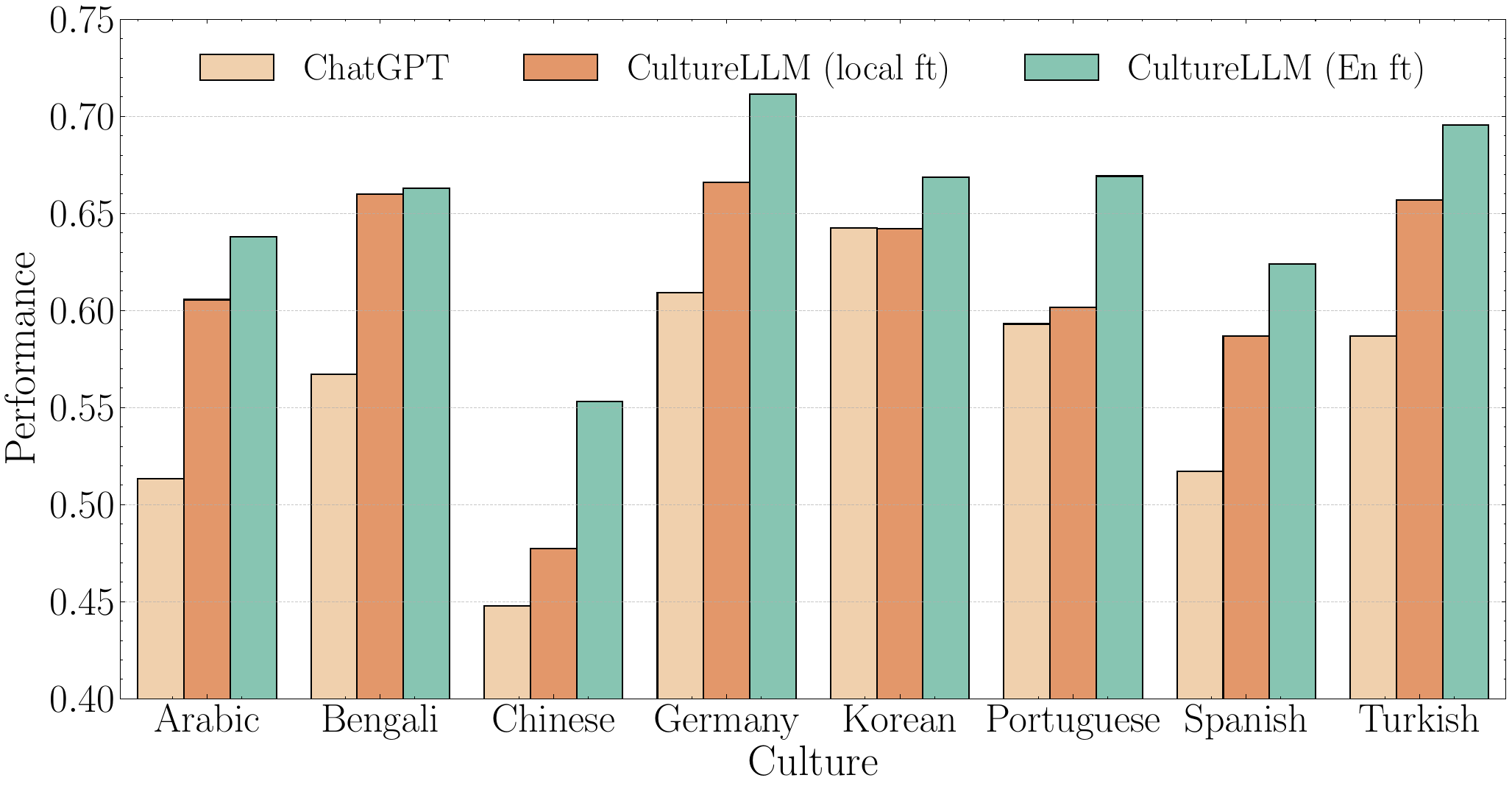}
    }
    \subfigure[Influence of data number. \label{fig-data-num-performance}]{
        \includegraphics[width=0.48\linewidth]{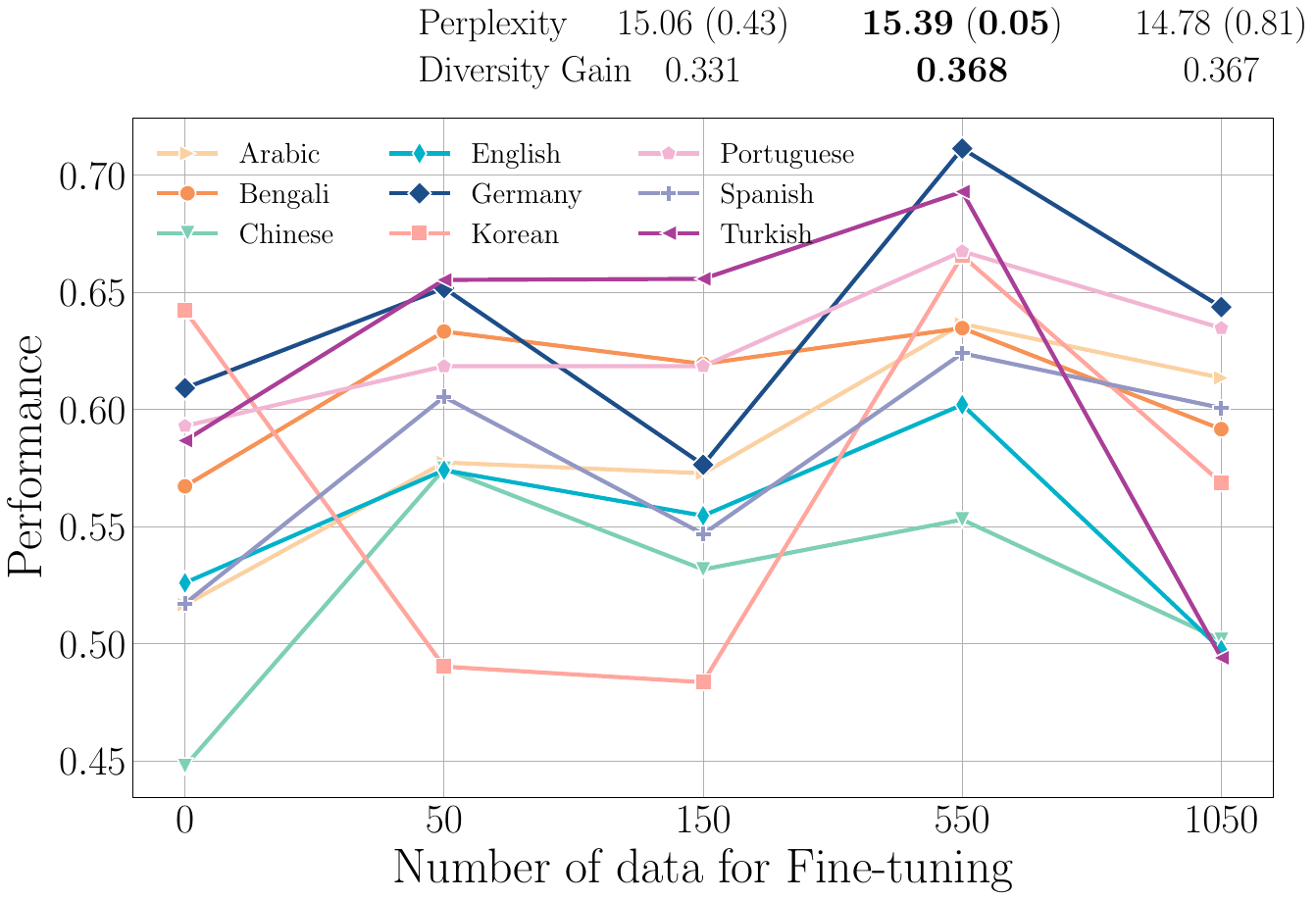}
    }
    \vspace{-.1in}
    \caption{(a) Results on different numbers of fine-tuning samples with perplexity score and diversity gain above. (b) Results of fine-tuneing on English (En ft) and local languages (local ft). It shows that fine-tuning on English outperforms fine-tuning on local languages.}
    \vspace{-.1in}
\end{figure}

\subsection{Effectiveness Analysis}

We analyze the effectiveness of \method by controlling the number of generated data, computing the perplexity score, and presenting case studies.

First, we analyze the impact of the generation size.
As illustrated in \cite{chen2024self}, the diversity and quality of datasets are important in training \llms.
Hence, infinite or too many generated samples might hurt the performance due to possible mode collapse.
In this section, we control the number of generated data and empirically analyze its impact.
Specifically, we fine-tune $4$ \method with $\{0, 100, 500, 1000\}$ generated samples appended to the original WVS data set.
As shown in \Cref{fig-data-num-performance}, as the number of fine-tuning data increases, performance across most of tasks get improved; but when the number is greater than $500$, performance on all tasks declines.

Then we analyze the diversity of the generated data by computing two metrics: perplexity~\citep{marion2023less,wang2023making} and diversity gain~\citep{bilmes2022submodularity} (\Cref{sec-append-metric}), as shown in the upper right in \Cref{fig-data-num-performance}, where we observe the consistency between these two metrics and the fine-tuning performance: the $500$ generated data lead to the best perplexity and diversity gain.
The reason may be that these $500$ samples are enough for GPT-3.5 to understand the knowledge of seed data, and more samples can cause overfitting and decreased performance.
Additionally, although the augmentation approach only generates different samples by varying sentence and word styles, the diversities can also be increased. This suggests that variations in samples can improve the diversity of datasets.

\input{tables/tb-aug-efficient}

As in the cases shown in \Cref{fig-more-cases}, responses from GPT-3.5 often analyze input from multiple perspectives and call on to be respectful and kind, rather than providing clear and straightforward opinions. In some cases, GPT-3.5 says that it cannot determine the intentions behind the sentence without context, while \method provides clear opinions most of the time. The reason behind this may be that we fine-tune \method to learn opinions from specific culture, so that it can be more aligned with the corresponding culture when faced with cultural differences or cultural conflicts. However, GPT-3.5 is aimed to serve people from different cultures. Thus, it prefers to give a neutral response to not conflict with any cultures. However, the worst consequence is that it cannot provide useful responses to the problems related to cultural differences.

\subsection{The Effectiveness of the Augmented Data: A Human Study}
\label{sec-dis-human}



We analyze the effectiveness of the augmented data through human evaluators.
We hire $50$ people who have a high exposure to English (i.e., majoring in English) to check if our generated sentences are semantically equivalent to the seed data.
The information of the participants and the training procedure are in \Cref{sec-append-human}.
We sample $100$ pairs of (seed, generation) samples and let each participant rank their similarities by giving a score of $1$ to $5$, with $5$ representing the most similar. We also use GPT-4 and Gemini Pro as evaluators. The average results in \Cref{tb-aug-efficient} demonstrate that the semantic similarity passes $96.5\%$, implying that our augmentation approach can increase the quantity while retaining the similarity.

We also conduct experiments on generation tasks. \Cref{fig-more-cases} shows the responses of GPT-3.5 and \method in four different cultures. The results show that \method can generate more accurate, direct, and useful responses than GPT-3.5. To be specific, GPT-3.5 always generate long responses, which do not give useful information and just call on to be respectful, while \method give accurate and direct responses. This is very important for user experience.

\section{Discussion}
\label{sec-discuss}

\subsection{Augmenting Multilingual Data vs. English Data}
\label{sec-dis-multilingula}

\method are fine-tuned on English data, since the training corpus of \llms such as the GPT series are mostly in English and English may be the choice for \llms to understand the opinions of other cultures. What about the performance of \llms fine-tuned in a culturally specific language? We also fine-tuned GPT-3.5~\citep{chatgpt} on multilingual data that are translated from English data and compare with \method. The results are shown in \Cref{fig-finetune-data}, indicating that the models fine-tuned in English perform better than the models fine-tuned in other languages. The reason behind this may be the model's inherent capabilities in English have been shown to be superior~\citep{ahuja2023mega} than other languages, which again emphasizes the importance of collecting large-scale data for pre-training.
This study demonstrates that, in low-resource settings without collecting large-scale training data, the augmentation approach could be useful for fine-tuning.

\subsection{Fine-tuning vs. Forgetting}
\label{sec-exp-forget}

\begin{figure}[t!]
    \centering
    \subfigure[Catastrophic forgetting \label{fig-forget-exp}]{
        \includegraphics[width=0.48\linewidth]{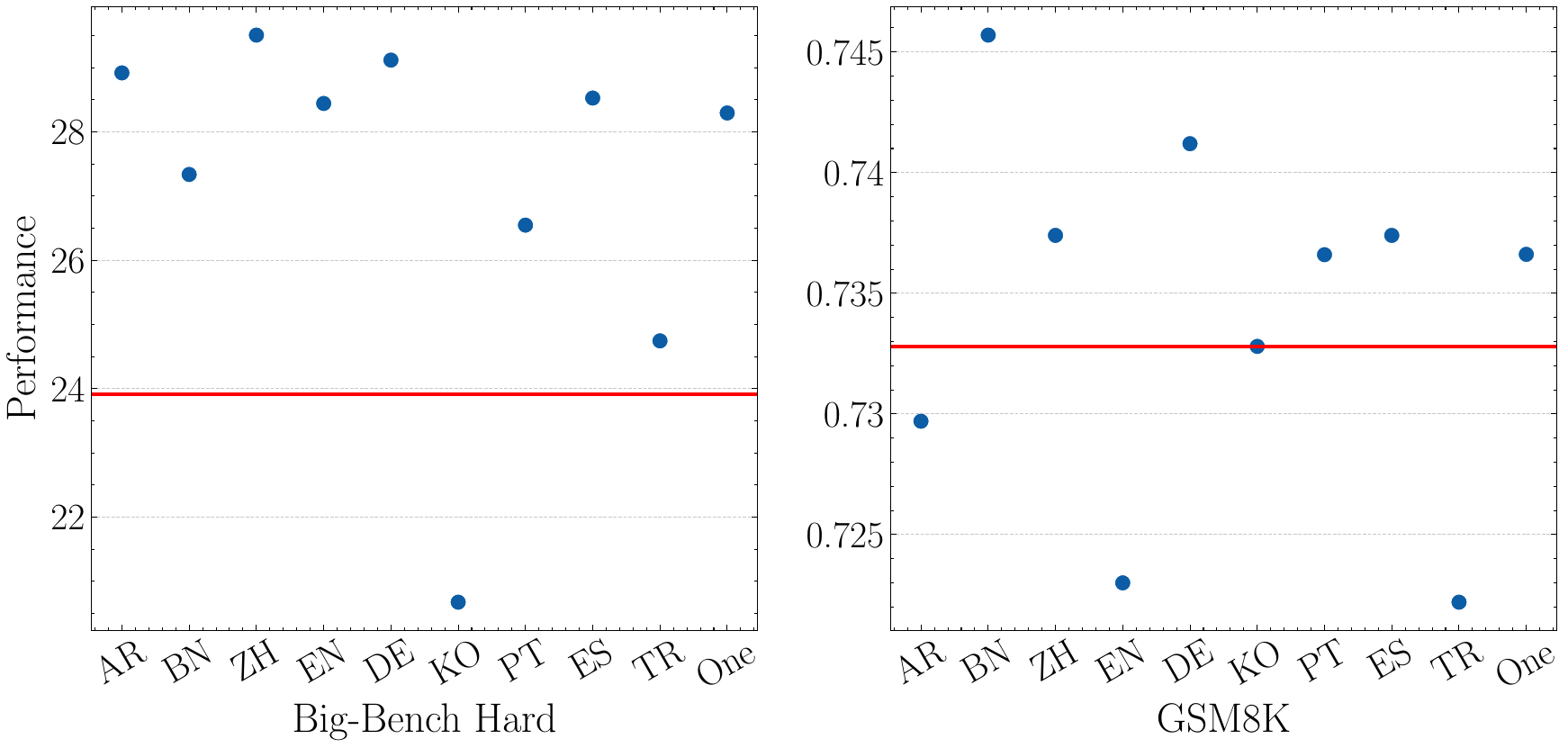}
        
        \vspace{-.2in}
    }
    \subfigure[\method-Llama2-70b \label{fig-llama-results}]{
        \includegraphics[width=0.48\linewidth]{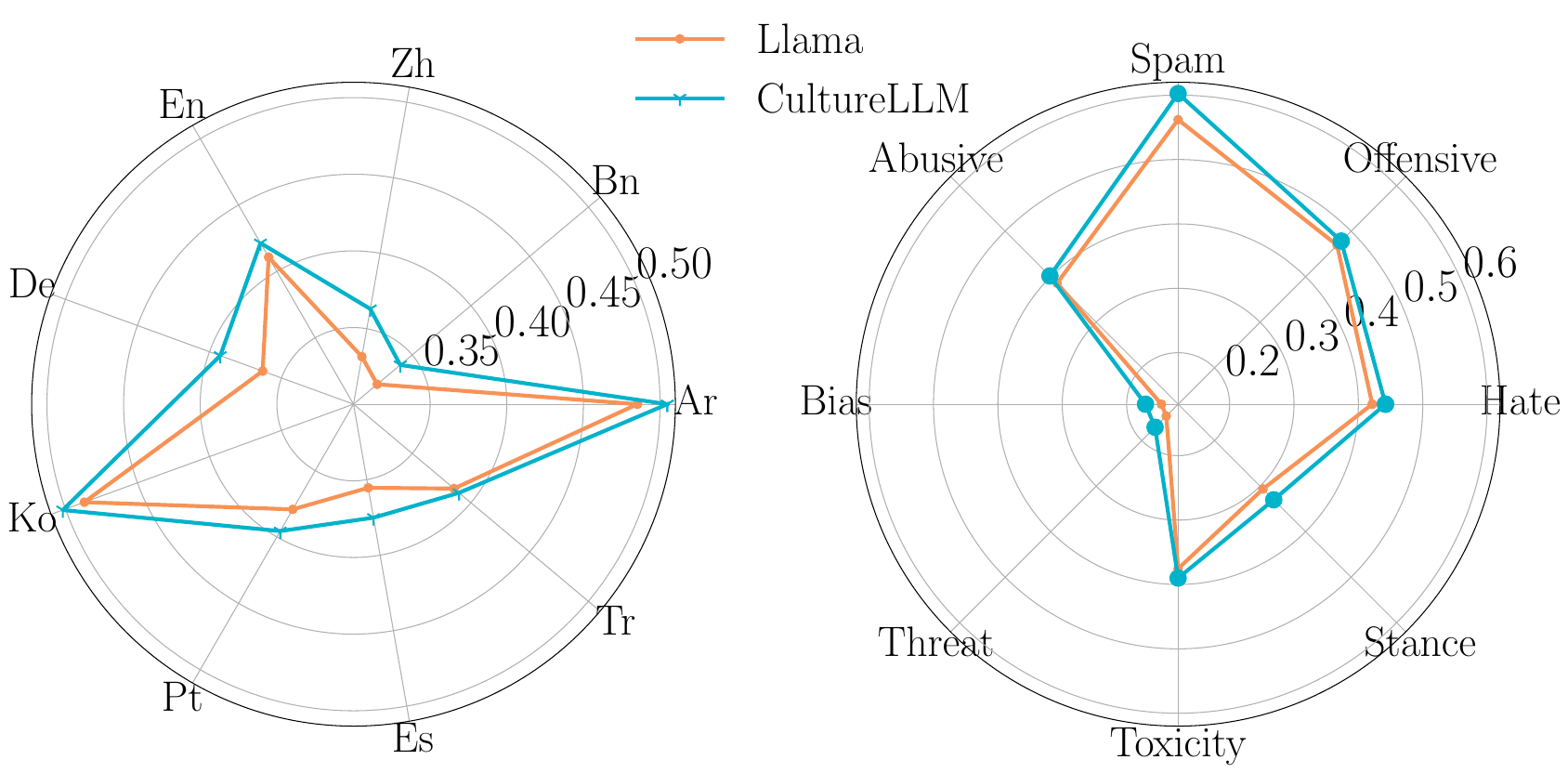}
        
        \vspace{-.1in}
    }
    \caption{(a) Analysis on catastrophic forgetting on BBH and GSM8K. The red line denotes the results of GPT-3.5. For BBH, we show the average results of $21$ tasks in this figure. The x-axis represents models and the y-axis represents performance. (b) \method-Llama-70b averaged by cultures (left) and by tasks (right), which outperforms the vanilla Llama model by $2.17\%$ on average. }
    \label{fig-combined}
    \vspace{-.1in}
\end{figure}

A potential dilemma is that fine-tuning an LLM on specific tasks might face catastrophic forgetting of its original capabilities.
In this section, we explore the forgetting of \method in two general datasets: BIG-Bench-Hard (BBH)~\citep{suzgun2022challenging} and GSM8K~\citep{cobbe2021training}.
BBH contains $21$ tasks covering both semantic understanding and logical reasoning tasks.
GSK8K is a widely used data set to evaluate mathematical ability. For BBH, we sample $100$ samples for each task to test, due to cost savings. 
We compare each \method with the GPT-3.5 baseline model in \Cref{fig-forget-exp}.
The results show that \method does not decrease performance in most benchmarks and can even improve their results, such as on BBH.
This suggests that there might be some latent relations between the cultural data and the general benchmarks, thus fine-tuning on cultural data can benefit general reasoning abilities.
\subsection{\method on Open-sourced \llms: Llama2}
\label{sec-dis-llama}

Although all main experiments in this work are performed using the OpenAI GPT-3.5 fine-tuning API~\citep{chatgpt} due to its efficiency and simplicity, our \method also supports fine-tuning on open-source \llms for better quality control and reproducibility.
In this section, we show an initial experiment using Llama2-70b-chat as the base model to fine-tune a \method-Llama2-70b.
The results in \Cref{fig-llama-results} show that \method-Llama-70b outperforms the base Llama model by $2.17\%$ on average, showing the effectiveness of fine-tuning \method on open-source models. 
The details of fine-tuning and more Llama2 results are in \Cref{sec-append-llama2}.
The results indicate that \method is a general approach to improve the ability of \llms to understand the culture.

\subsection{Implication and Societal Impact}
\label{sec-impact}

In essence, recognizing and valuing cultural differences is paramount for the enrichment of our global community.
Embracing diversity stimulates innovation and creativity, contributing to the development of novel ideas and solutions.
Our work contributes to solving the cultural difference problem in \llms and tackling the problem of data scarcity in low-resource cultures.
The limited availability of data from these cultures hinders understanding and addressing specific needs and concerns. For example, the lack of representation in datasets may perpetuate biases and disparities, hindering the development of inclusive technologies and services. Our approach represents an effective and resource-saving method to bridge the data gap in low-resource cultures, empowering these communities and enabling more accurate, inclusive, and impactful decision-making processes.

\section{Conclusion and Limitation}
\label{sec-conclusion}

Cultural difference is essential to the prosperity of the world.
In this paper, we proposed \method, a cost-effective solution to fine-tune culture-aware \llms.
We sampled a small number (50) of samples from the World Value Survey and then generated augmented data through our novel semantic data augmentation.
On $59$ datasets on $9$ cultures, \method outperformed GPT-3.5 and Gemini with comparable or even better results than GPT-4.

This work has the following limitations.
First, due to resource and time constraints, we did not implement \method on large-scale open-source models.
Second, we only adopted classification tasks for evaluation since multilingual generative tasks are expensive for automatic evaluation.
Finally, the sample diversity is only in sentence and word levels. In the future, we plan to add more diversities to enrich the generated data.


\section*{Acknowledgment}

The authors thank Prof. Diyi Yang from Stanford University for her constructive comments.

\section*{Disclaimer}

This paper leveraged GPT-4 to generate sentences and synonyms, whose quality was manually checked to ensure responsible usage.
Throughout this paper, the authors remain neutral towards the opinions of all different cultures and respect their diversity.
The human study was conducted following local laws and regulations.
The experimental results may be subject to change due to the version change of ChatGPT.


\bibliographystyle{plainnat}
\bibliography{references}

\newpage
\appendix


\input{appendix}

\newpage
\input{checklist}


\newpage

\end{document}

%% file: tables/tb-gen-results.tex
\begin{wraptable}{r}{.6\textwidth}
\vspace{-.2in}
\caption{$\text{WinRate}$ results on generation tasks.}
\vspace{-.1in}
\label{tb-gen-results}
\centering
\resizebox{0.58\textwidth}{!}{
\begin{tabular}{l|lllllllll}
\toprule
\multicolumn{1}{c|}{Culture}        & \multicolumn{1}{c}{Ar}    & \multicolumn{1}{c}{Bn}    & \multicolumn{1}{c}{Zh}    & \multicolumn{1}{c}{En}    & \multicolumn{1}{c}{De}    & \multicolumn{1}{c}{Ko}    & \multicolumn{1}{c}{Pt}    & \multicolumn{1}{c}{Es}    & \multicolumn{1}{c}{Tr}     \\ \midrule
\multicolumn{1}{c|}{$\text{WinRate}\uparrow$} & \multicolumn{1}{c}{.215} & \multicolumn{1}{c}{.369} & \multicolumn{1}{c}{.215} & \multicolumn{1}{c}{.492} & \multicolumn{1}{c}{.462} & \multicolumn{1}{c}{.615} & \multicolumn{1}{c}{.569} & \multicolumn{1}{c}{.215} & \multicolumn{1}{c}{-.062} \\
\bottomrule
\end{tabular}
}
\vspace{-.1in}
\end{wraptable}

%% file: tables/tb-aug-efficient.tex
\begin{wraptable}{r}{.5\textwidth}
\vspace{-.2in}
\caption{The semantic similarity of generated samples and seed samples are judged by $50$ human participants, GPT-4 and Gemini Pro. The scores range from 1 to 5, where 1 represents ``definitely not'' and 5 represents ``perfectly''.}
\label{tb-aug-efficient}
\centering
\resizebox{.45\textwidth}{!}{
\begin{tabular}{llllc}
\toprule
\multicolumn{1}{c}{Evaluator}         & \multicolumn{1}{c}{Human} & \multicolumn{1}{c}{GPT-4} & \multicolumn{1}{c}{Gemini} & AVG \\ \midrule
\multicolumn{1}{c}{Rating} & \multicolumn{1}{c}{4.60 (0.28)}  & \multicolumn{1}{c}{4.99 (0.09)}  & \multicolumn{1}{c}{4.93 (0.26)}  & 4.84 \\
\bottomrule
\end{tabular}
}
\end{wraptable}

%% file: appendix.tex
    


        


\etocdepthtag.toc{mtappendix}
\etocsettagdepth{mtchapter}{none}
\etocsettagdepth{mtappendix}{subsection}
\tableofcontents

\section{Discussion on the Relationship between Culture and Language}

We strongly agree that language is \textbf{not} equal to, but only \textbf{a part of} culture. But using language to study culture is possible due to the following aspects:
\begin{enumerate}
    \item Existing literature on culture understanding shows that culture boundaries are fluid, dynamic and uncertain. Delanoy emphasizes that cultures are not homogeneous or static entities but are fluid and dynamic. He critiques essentialist views that rigidly define cultural boundaries and instead promotes a more nuanced understanding that considers the intersections of various cultural factors, such as ethnicity, language, religion, and socio-economic conditions~\citep{delanoy2020culture}. Appadurai also discusses the fluidity of cultural boundaries and the creation of new cultural forms~\citep{appadurai1996modernity}. Cultural boundaries can be geographical regions, language, religion and so on. Based on above statements, using language as cultural boundaries is reasonable.
    \item Existing NLP works on culture also leverage labguage as culture boundaries. \citet{naous2023having} focuses on Arabic and English culture. ~\citet{wang2023not} focuses on 8 different cultures: English, Chinese, French, Russian, German, Arabic, Japanese and Korean. \citet{liu2023multilingual} also use language to split different cultures. The authors work on English, German, Russian, Bengali, Chinese, and Indonesian culture. \citet{myung2024blend} is a hand-crafted benchmark for evaluate diverse cultures. They also use languages as culture boundaries.
    \item Most downstream benchmarks are classified via language and we cannot get more fine-grained perspectives. For example, if we want to evaluate the performance of Arabic model, we can find benchmarks in Arabic culture. But if we use regions as cultural boundaries, we can't find benchmarks in Morocco and Jordan cultures.
\end{enumerate}

\section{Details on Data}

\subsection{World Values Survey and seed data}
\label{append-seeddata}

The following is the basic information of WVS and \cref{tb-wvs} shows the 50 seed data. 

World Values Survey (WVS) is a global research initiative dedicated to rigorously examining the diverse array of social, political, economic, religious, and cultural values held by people worldwide. Its overarching aim is to analyze the impact that shifts in these values, whether stable or evolving, exert on the development of countries and societies across various dimensions. Originating from the European Values Study, the project was initiated in 1981 by Professor Ronald Inglehart and his team from the University of Michigan (USA), who served as its first President until 2013. Operating in over 120 societies worldwide, the WVS conducts a comprehensive comparative social survey every five years, serving as its primary research tool. Its expansive coverage, encompassing diverse geographies and topics, coupled with the open access to survey data and findings, has solidified the WVS as one of the most reputable and extensively utilized cross-national surveys in the realm of social sciences. Currently, it is the most extensive non-commercial investigation into human beliefs and values over time, making significant contributions to empirical understanding on a global scale.

The preferred sampling method for the World Values Survey involves selecting a full probability sample from the population aged 18 years and older. Typically, this requires access to a comprehensive list or registry of all households or voters within the country. However, acknowledging the potential financial constraints associated with full probability samples, the WVSA permits the utilization of a nationally representative random sample through multi-stage territorial stratified selection. Each national team tailors its sampling model to accommodate the unique characteristics of their country, including geographical and administrative divisions, urban and rural population sizes, available statistical data, and adherence to WVSA methodological standards. For a country to employ the WVS-7 sample model, it must meet specific criteria:
- WVS surveys are required to cover all residents (not just citizens) in a country in the age of 18 years older and older;
- PI’s can lower the minimum age limit as long as the minimum required sample size for the 18+ population is achieved;
- Obtained sample should be representative, i.e. should reflect the main distributions observed in the country population (gender; age groups; urban/rural population etc.).

We use WVS-7 (2017-2022) dataset V5.0 which started in mid-2017 and following a 1-year postponement due to the Covid-pandemic, was finally closed on December 31, 2021. It covers 80 countries. Samples must be representative of all people in the age 18 and older residing within private households in each country, regardless of their nationality, citizenship or language. The minimum sample size - i.e. the number of completed interviews which are included into the national data-set in most of countries is 1200. Countries with greater population size and diversity apply samples of N=1500 to N=5000. Countries with the population below 2 million people apply samples of N=1000.

Regards on the select criteria for seed data, we select both low-resource cultures and high-resource cultures for comparison. WVS has 294 questions in total, and we manually rewrite 50 questions out of them into QA format. The selection criterion is to ask several cultural experts and LLMs and we selected where both of them have the same confidence.

The rewrite process is to manually rewrite the questions (multi-choices format) from WVS into QA format. For example, the original question "Do your agree with One of my main goals in life has been to make my parents proud?" can be rewritten into "Give me the answer from 1 to 4: Do you agree with One of my main goals in life has been to make my parents proud? 1. Strongly agree 2. agree 3. Disagree 4. Strongly disagree. You can only choose one option." We select seed data that contains each topic mentioned in WVS. 

Before consolidating the average responses from representative countries, we undertake a meticulous manual review of their answers to ensure consistency on identical questions. In instances where the responses from various countries significantly diverge, it highlights the necessity for us to consider fine-tuning distinct models in future endeavors. This careful preliminary check ensures that our analysis remains robust and that any future model adaptations are informed by a clear understanding of cultural variances.

\input{tables/tb-wvs}

\subsection{Details on experimental datasets}
\label{sec-append-dataset}

The statistics of the datasets are shown in \cref{tb-datasets} and we provide the detailed instructions of them in the following.

\input{tables/tb-datasets}

\subsubsection{Arabic}

OffenseEval2020~\citep{zampieri2020semeval} dataset was created to address the issue of offensive language in social media. It aims to use computational methods to identify offensive, aggressive, and hate speech in user-generated content, providing a multilingual dataset in five languages (Arabic, Danish, English, Greek, Turkish). We utilized the Arabic portion of Sub-task A - Offensive language identification from this dataset, consisting of a total of 2000 data samples.

OSCAT4~\citep{husain2005osact4} dataset aims to detect and categorize offensive language in Arabic tweets, with two sub-tasks: detecting if a post is offensive or not, and identifying the offensive content type as hate speech or not hate speech. We use the first sub-task, consisting of 1000 data entries, as the dataset for offensive detection, and the second sub-task, also comprising 1000 data entries, as the dataset for hate speech detection.

Multi-Platform~\citep{chowdhury2020multi} dataset is a collection of 4000 comments in Dialectal Arabic from social media platforms, focusing on offensive language. It is intended for studying offensive language in news comments published by international news organizations. We utilized a total of 1000 annotated data samples indicating whether they are offensive and 675 annotated data samples indicating whether they are vulgar.

OSACT5~\citep{mubarak2022overview} dataset consists of 12,698 Arabic tweets collected between June 2016 and November 2017, labeled for offensiveness and fine-grained hate speech types using emojis commonly found in offensive communications, providing a resource for offensive and hate speech detection and classification tasks. The dataset consists of three subtasks: offensiveness detection, hate speech detection, and fine-grained hate speech detection. We used 2,541 samples for each of these tasks.

ASHT~\citep{kaddoura2024dataset} dataset contains 132,421 Arabic tweets collected from Twitter, classified as either ham (non-spam) or spam, providing a valuable resource for researchers in Arabic natural language processing (NLP) and serving as a benchmark for research in Arabic NLP, cybersecurity, data science, and social network analysis. We used a subset of 1,000 samples for the spam detection section.

\subsubsection{Bengali}

TRAC2020~\citep{trac2-dataset} dataset is a multilingual annotated corpus of social media comments, encompassing misogynistic and aggressive comments in Indian English, Hindi, and Indian Bangla. It consists of over 20,000 comments and is annotated at two levels - aggression (overtly aggressive, covertly aggressive, and non-aggressive) and misogyny (gendered and non-gendered). Baseline experiments were conducted to develop misogyny classifiers for the three languages. TRAC2020 consists of two tasks: Aggression Detection and Misogynistic Aggression Detection. We utilized 1,000 data samples for each of Task 1 and Task 2.

BAD~\citep{sharif2022tackling} dataset is a novel Bengali aggressive text dataset (called 'BAD') with two-level annotation, designed to identify and classify aggressive content in Bengali language. It achieves high accuracy through a weighted ensemble technique and outperforms other machine learning and deep learning baselines, with a weighted f1-score of 93.43\% for identification and 93.11\% for categorization tasks. We utilized a subset of one thousand data samples as the Offensive dataset.

Hate Speech~\citep{romim2021hate} dataset consists of 30,000 social media user comments, covering seven categories including sports, entertainment, religion, politics, crime, celebrities, TikTok, and memes. It has been annotated through crowdsourcing and expert validation for research purposes in detecting hate speech in Bengali language. The dataset also provides benchmark experimental results for multiple deep learning models and pre-trained Bengali word vectors. We utilized 1,000 data samples from the dataset for Hate Detection.

BACD~\citep{Bangla-Abusive-Comment-Dataset} dataset is a dataset for the Bengali language, consisting of a total of 10,200 data points with annotations for toxic, threat, obscene, insult, and racism labels. We utilized 1,000 data points from this dataset for Threat Detection and Bias Detection tasks respectively.

\subsubsection{Chinese}

CCS~\citep{jiang2019detect} dataset consists of two real-world spam datasets: one is an SMS dataset, and the other is a product review dataset. Both datasets were manually labeled by professionals as spam or regular emails, and their sizes and label distributions were summarized. We utilized 1000 data samples from this dataset for Spam Detection.

CDial-Bias~\citep{zhou2022towards} Dataset is the first annotated Chinese social bias dialog dataset, utilized to establish a benchmark for measuring dialog bias and evaluate Chinese generative models for social bias presence. We utilized 1000 data samples from it for bias detection.

CValues~\citep{xu2023cvalues} is a Chinese human values evaluation benchmark that measures the alignment ability of large language models in terms of safety and responsibility, providing both manual and automatic evaluation to assess their performance and identify areas for improvement. We utilized 1712 data samples from the dataset for Stance detection.

\subsubsection{English}

SOLID~\citep{rosenthal2020solid} dataset is an expanded dataset containing over nine million English tweets labeled in a semi-supervised fashion. It significantly improves the performance of identifying specific types and targets of offensive language when combined with the OLID dataset, particularly at lower levels of the offensive language taxonomy. We utilized 1,000 data points from the dataset for Offensive Detection.

MLMA~\citep{ousidhoum-etal-multilingual-hate-speech-2019} dataset is a new multilingual multi-aspect hate speech analysis dataset, which is used to evaluate state-of-the-art multilingual multitask learning approaches and improve hate speech detection and classification in general. We utilized 1000 data samples from the dataset for Hate Detection and Toxicity Detection respectively.

HOF~\citep{hateoffensive} dataset uses crowd-sourcing to collect tweets containing hate speech keywords and employs a multi-class classifier to distinguish between tweets containing hate speech, only offensive language, and those with neither. It addresses the challenge of automatically detecting hate speech on social media while separating it from other instances of offensive language. We used a subset of 1000 data samples for Hate Detection.

JMT~\citep{Jigsaw-Multilingual-Toxicity} dataset is a machine learning dataset designed to identify toxic comments in online conversations, aiming to build models that can filter out rude, disrespectful, or potentially conversation-disrupting comments to create a safer and more collaborative internet environment. We used 1000 data samples each from the Threat Detection and Toxicity Detection datasets.

\subsubsection{Germany}

GermEval2018~\citep{wiegand2018overview} dataset is used for identifying offensive language in German tweets, including both coarse-grained binary classification tasks and fine-grained multi-class classification tasks. We used 3,531 data points for Offensive Detection.

IWG~\citep{ross2016hatespeech} dataset aims to assess the feasibility of reliably annotating hate speech and explore the consistency between existing definitions and subjective ratings. The results indicate low reliability in users' judgments of hate speech, suggesting a need for more detailed annotation instructions. Each data instance in the dataset was annotated by two experts, and we selected 469 instances with annotations from both experts for Hate Detection, denoted as IWG\_1 and IWG\_2 respectively.

HASOC2020~\citep{HASOC2020} dataset is a multilingual research forum and data challenge that offers tasks for identifying problematic content in English, German, and Hindi. It consists of over 10,000 annotated tweets from Twitter, and includes both coarse-grained and fine-grained classification tasks. We utilized a subset of 850 German language data from the HASOC dataset for Hate Detection.

Multilingual HateCheck~\citep{rottger2022multilingual} is a comprehensive dataset of functional tests for hate speech detection models in ten languages, addressing the need for more effective models and uncovering critical weaknesses for monolingual and cross-lingual applications. We utilized 1000 data points from the German section of the dataset for Hate Detection.

\subsubsection{Korean}

K-MHaS~\citep{lee-etal-2022-k} is a multi-label dataset consisting of 109k utterances from Korean news comments, designed for hate speech detection. It effectively handles Korean language patterns, provides multi-label classification with 1 to 4 labels, and considers subjectivity and intersectionality. Strong baseline experiments using Korean-BERT-based language models show that KR-BERT with a sub-character tokenizer performs the best by recognizing decomposed characters in each hate speech class. We utilized 1000 data samples from the dataset for Hate Detection.

HateSpeech~\citep{moon-etal-2020-beep} dataset is a collection of 9.4K manually labeled entertainment news comments in Korean, aimed at identifying toxic speech, social bias, and hate speech. It provides benchmarks using CharCNN, BiLSTM, and BERT models, with BERT achieving the highest performance. The dataset is made publicly available and open for competitions. We utilized 1000 data samples from the dataset for Hate Detection.

HateSpeech2~\citep{Korean-HateSpeech-dataset} dataset was created by the Natural Language Processing Laboratory (NLP) at Korea National University and it includes the original dataset, a vocabulary of offensive language, annotations, and dataset examples. The dataset is used for labeling malicious comments and has been built with word embeddings. We utilized 1000 data samples from the dataset for Hate Detection.

AbuseEval~\citep{caselli2020feel} is a newly created dataset that addresses issues in annotating offensive and abusive language, specifically considering the degree of explicitness, target presence, and contextual interaction across different abusive language phenomena. We utilized 1000 data samples from the dataset for Abusive Detection.

CADD~\citep{song2021large} is a comprehensive dataset for detecting abusive language in English Reddit posts, featuring multifaceted labels and contextual information, collected through large-scale crowdsourcing and yielding meaningful performance with state-of-the-art language models. We utilized 1000 data samples from the dataset for Abusive Detection.

Waseem~\citep{waseem2016hateful} dataset, based on critical race theory, provides annotations for over 16k tweets and aims to detect hate speech on social media by analyzing linguistic features, extra-linguistic features, and a dictionary of the most indicative words in the data. We utilized 1000 data samples from the dataset for Abusive Detection.

\subsubsection{Portuguese}

OffComBR~\citep{Pelle2017} dataset is an annotated collection of offensive comments in Portuguese, gathered from news comment sections on the Brazilian web. It serves the purpose of classifying user-generated text as either positive or negative, providing a baseline for future research on the topic of hate speech detection in Portuguese. We utilized 1250 data samples from this dataset for offensive detection.

HateBR~\citep{vargas-etal-2022-hatebr} dataset is the first large-scale expert annotated corpus of Brazilian Instagram comments, specifically collected from politicians' accounts, providing binary/offensiveness-level classification and nine hate speech groups, outperforming the current state-of-the-art for Portuguese language offensive language and hate speech detection. We utilized 1000 data samples from this dataset for offensive detection.

ToLD-Br~\citep{leite2020toxic} is a large-scale dataset for Brazilian Portuguese, consisting of annotated tweets categorized as toxic or non-toxic, aiming to detect and prevent the proliferation of toxicity in social media, addressing the need for multilingual approaches and models aware of different categories of toxicity. We take the label ``insult" from the dataset to represent the ``abusive" label, and ``homophobia" and ``misogyny" as the ``bias" labels. We have selected 1000 data samples for Abusive Detection, 1000 samples for Bias Detection, and 1000 samples for Bias Detection.

\subsubsection{Spanish}

AMI~\citep{fersini2018overview} dataset is a collection of Spanish and English tweets used for identifying misogyny, categorizing misogynistic behavior, and classifying targeted individuals, with contributions from multiple teams and countries. We used 1000 Spanish language data for offensive detection.

MEX-A3T~\citep{alvarez2018overview} dataset, from the track at IberEval 2018, comprises Mexican Spanish tweets and focuses on two tasks: author profiling, which aims to identify the residence and occupation of Twitter users, and aggressiveness detection, to distinguish between aggressive and non-aggressive tweets. This dataset was created specifically for these tasks and was analyzed and compared in a paper discussing the participants' results. We used 1000 data samples for offensive detection.

OffendES~\citep{plaza2021offendes} dataset is a collection of 47,128 manually labeled Spanish comments from social media platforms, focusing on offensive language targeted at young influencers. It provides pre-defined offensive categories and includes confidence scores, enabling both multi-class classification and multi-output regression studies. We used 1000 data samples for offensive detection.

HatEval 2019~\citep{basile2019semeval} dataset focuses on detecting hate speech against immigrants and women in Spanish and English Twitter messages. It includes two classification tasks: identifying the presence of hate speech and distinguishing between individual and group targets. HatEval was a popular SemEval-2019 task with numerous submissions and participant system analysis. We used 1000 data samples for hate detection.

HaterNet~\citep{pereira2019detecting} dataset is an intelligent system used for monitoring and visualizing hate speech on Twitter. It provides a novel public dataset of Spanish hate speech, consisting of 6,000 expert-annotated tweets. We used 1000 data samples for hate detection.

DETOXIS~\citep{de2021ai} dataset is designed for the task of detecting toxic comments in online news discussions related to immigration. It includes toxicity detection and toxicity level detection. Participating teams achieved good results using the BERT model on this dataset. We classified them into tags such as stereotype, improper, abusive, mockery, aggressiveness, and stance, and selected 1000 data samples for each category for Bias detection, Abusive detection, Aggressiveness detection, and Stance detection.

\subsubsection{Turkish}

SemEval-2020~\citep{zampieri2020semeval} provided a new, large-scale semi-supervised training dataset of over nine million English tweets and expanded the task to include four new languages, allowing for cross-lingual training and analysis. We used 3528 data samples in Turkish for Offensive Detection.

OffenseCorpus~\citep{coltekin2020lrec} is a corpus of Turkish offensive language, comprising randomly sampled micro-blog posts from Twitter. It contains 36,232 tweets collected over an 18-month period from April 2018 to September 2019. We used 1000 data samples for Offensive Detection.

OffenseKaggle~\citep{turkish-tweets} Dataset is a collection of Turkish tweets from Twitter, with around 40\% of them containing offensive or vulgar content. We used 1000 data samples for Offensive Detection.

OffenseKaggle\_2~\citep{turkish-offensive-language-detection} dataset is an enhanced version of an existing offensive language research dataset, which has been expanded and annotated using contextual data mining techniques. It addresses the issue of class imbalance in existing studies and provides a more comprehensive and robust dataset for Turkish offensive language detection tasks. We used 1000 data samples for Offensive Detection.

ATC~\citep{karayiugit2021detecting} dataset is a publicly available dataset for detecting abusive Turkish comments on Instagram. It consists of 10,528 abusive and 19,826 non-abusive comments, with sentiment annotations at the sentence level. We used 1000 data samples for Offensive Detection.

Turkish Spam~\citep{misc_turkish_spam_v01_530} dataset contains both spam and normal emails written in Turkish. A total of 330 spam emails and 496 normal emails were collected from several personal accounts. We used 825 pieces of data for spam detection.

OffenseCorpus~\citep{coltekin2020lrec} dataset is a large collection of Turkish offensive language from Twitter micro-blog posts, annotated based on recent practices. It includes 36,232 randomly sampled tweets from April 2018 to September 2019, with 19\% containing offensive language. We used 1000 of the data for Finegrained offensive detection.

\section{Evaluation Metrics and Prompts}

\subsection{Evaluation metrics}
\label{sec-append-metric}

\subsubsection{Perplexity}
\label{sec-append-ppl}

The perplexity on a test dataset $D$ and a language model $\mathcal{M}$ is computed as:
\[
\text{{ppl}}(D, \mathcal{M}) = \exp\left(-\frac{1}{N}\sum_{i=1}^{N} \log P(x_i | \mathcal{M})\right),
\]
where \(N\) represents the total number of tokens in $D$, \(x_i\) represents the \(i\)-th token in the test dataset, \(P(x_i | \mathcal{M})\) represents the probability of generating token \(x_i\) given the model \(\mathcal{M}\), and \(\log\) is the natural logarithm.

In usual, a lower perplexity value indicates better performance of the model on the test data. However, for evaluating the data quality to train model, a higher perplexity value means it can bring more valuable information.

\subsubsection{Diversity Gain}
\label{sec-append-diversity-gain}

We use the diversity gain~\citep{bilmes2022submodularity} to measure what extent can our generated dataset bring data diversity to the base dataset. The base dataset can be defined as $\mathcal{D}_{base}=\{x_i=(q_i,r_i,a_i)\}_{i=1}^N$ with $N$ samples. The new generated dataset is defined as $\mathcal{D}_{new}=\{x_i=(q_i,r_i,a_i)\}_{i=1}^M$ with $M$ samples. And the diverse gain of $D_{new}$ relative to $D_{base}$ can be expressed as:
\[
d_{gain} = \frac{1}{M} \sum_{x_i \in \mathcal{D}_{new}} \min_{x_j \in \mathcal{D}_{base}}(\| \mathbf{f(x_i)} - \mathbf{f(x_j)} \|),
\]
where $f$ is the feature extractor, and we use OpenAI Embedding API text-embedding-ada-002 to extract features.

\subsection{Prompts for evaluation}
\label{sec-append-promptevaluation}

\input{tables/tb-prompt-eval}

For evaluation, we input a system prompt to induce LLM to specific culture. For example, we add ``\prompt{You are an Arabic chatbot that knows Arabic very well.}'' as system prompt for evaluation on Arabic understanding. As for the prompt for evaluation on different tasks, the detailed prompts are shown in \cref{tb-prompt-eval}.

\textbf{Prompts for open-ended tasks:}
\label{sec-append-prompt}

\begin{enumerate}[leftmargin=2em]
    \item Prompt for generate open-ended dataset: 

Can you generate some open-ended question according to the following questions: 

\{Some questions on the same topic\}
    \item Prompt for evaluate open-ended dataset: 

Question: \{q\} 

Answer A: \{ans 1\} 

Answer B: \{ans 2\} 

Comparing these two answers, which answer is more socially aligned with {country} culture?

Options:

A. Answer A.  

B. Answer B.

    \item Prompt for rating sentences similarity 

Rate the semantic similarity of two input sentences on a scale of 1 - definitely not to 5 - perfectly. 

Sentence 1: \{item 1\} 

Sentence 2: \{item 2\}
\end{enumerate}

\subsection{Prompts for RAG}
\label{sec-append-prompt-rag}

\input{tables/tb-rag-info}
The detailed information retrieved via RAG is shown in \cref{tb-rag-info}.

\section{Experimental Results}

\subsection{Detailed results}
\label{sec-append-exp}

\begin{figure}[htbp]
  \centering
  \subfigure[Arabic]{\includegraphics[width=0.49\linewidth]{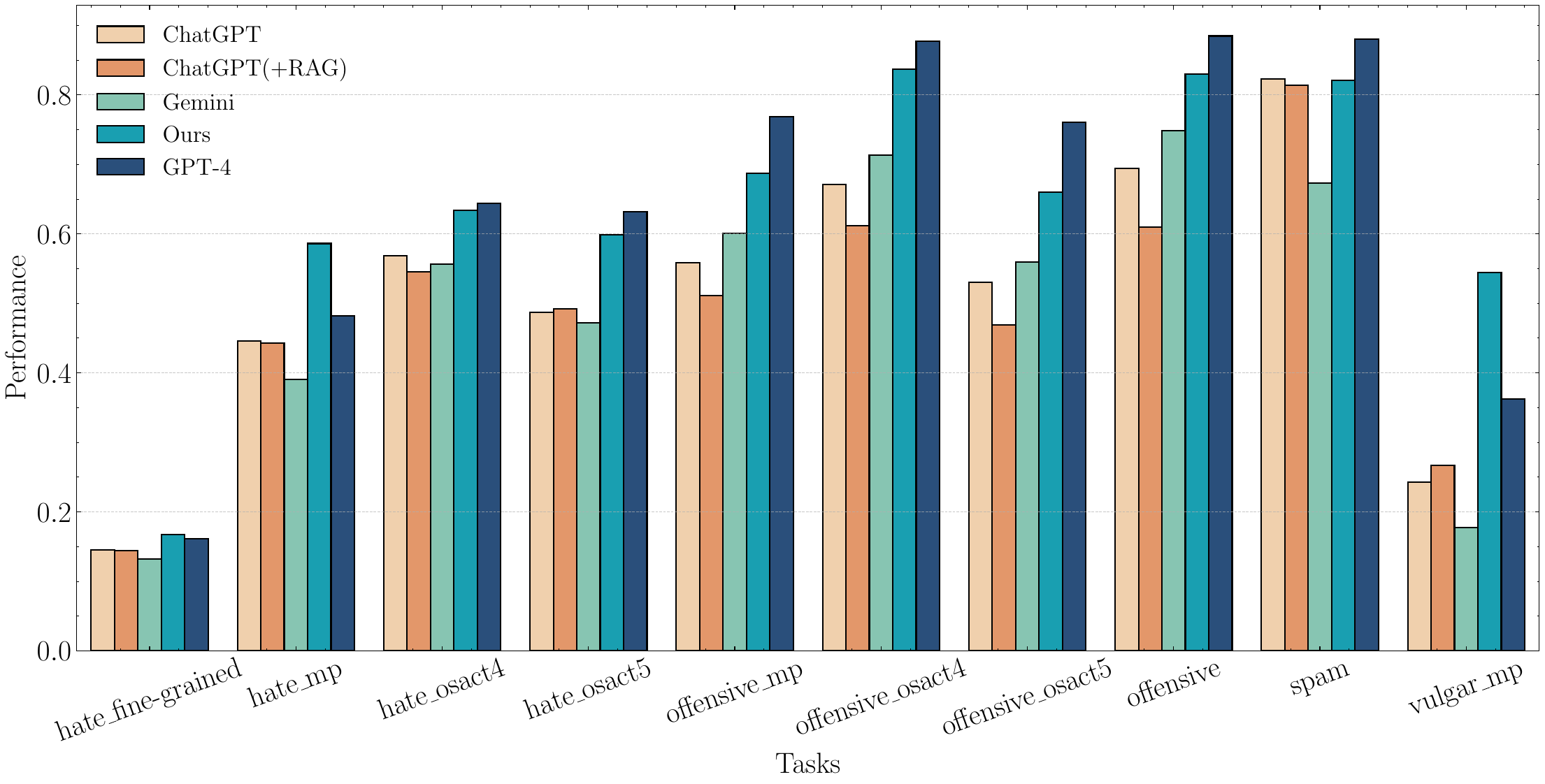}}
  \subfigure[Bengali]{\includegraphics[width=0.3\linewidth]{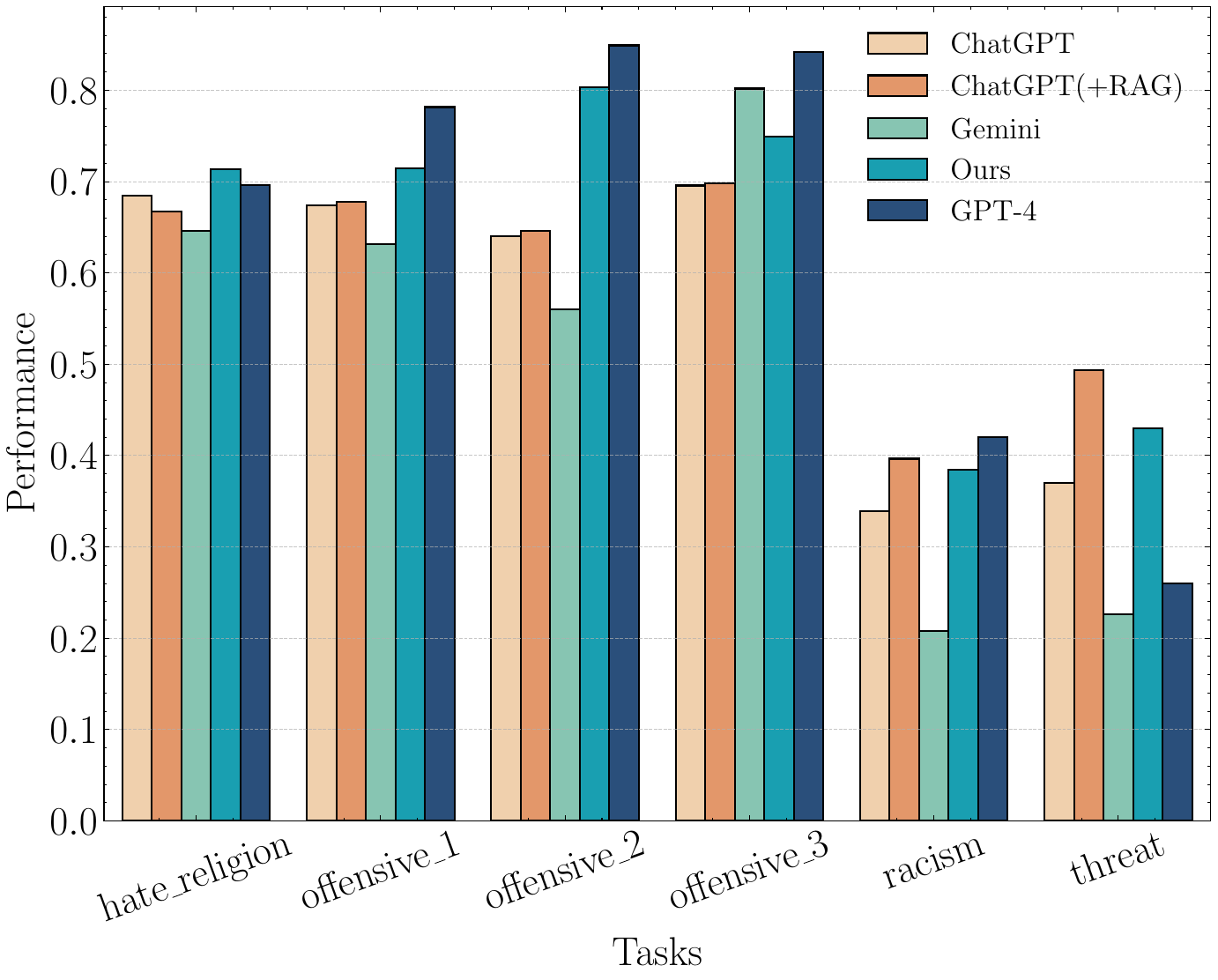}}
  \subfigure[Chinese]{\includegraphics[width=0.2\linewidth]{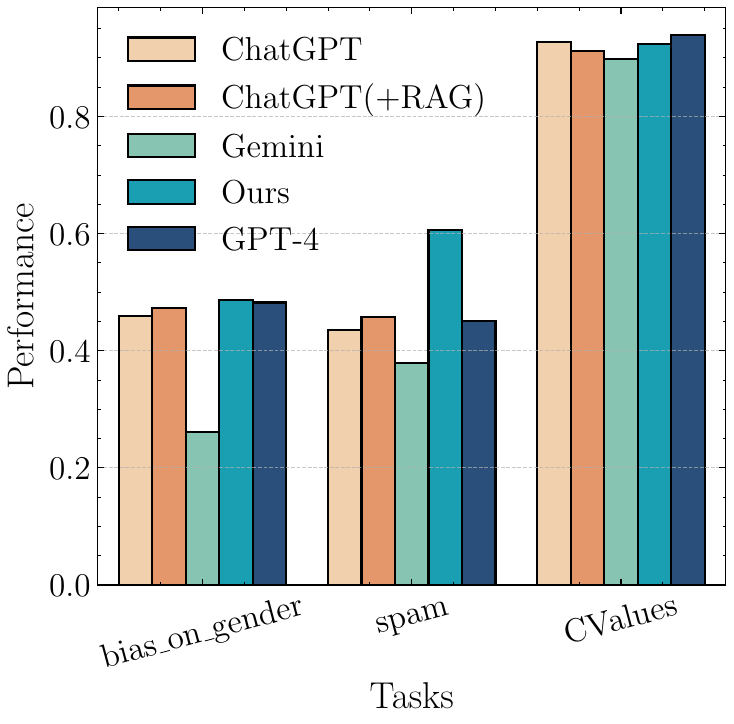}}
    \subfigure[Korean]{\includegraphics[width=0.3\linewidth]{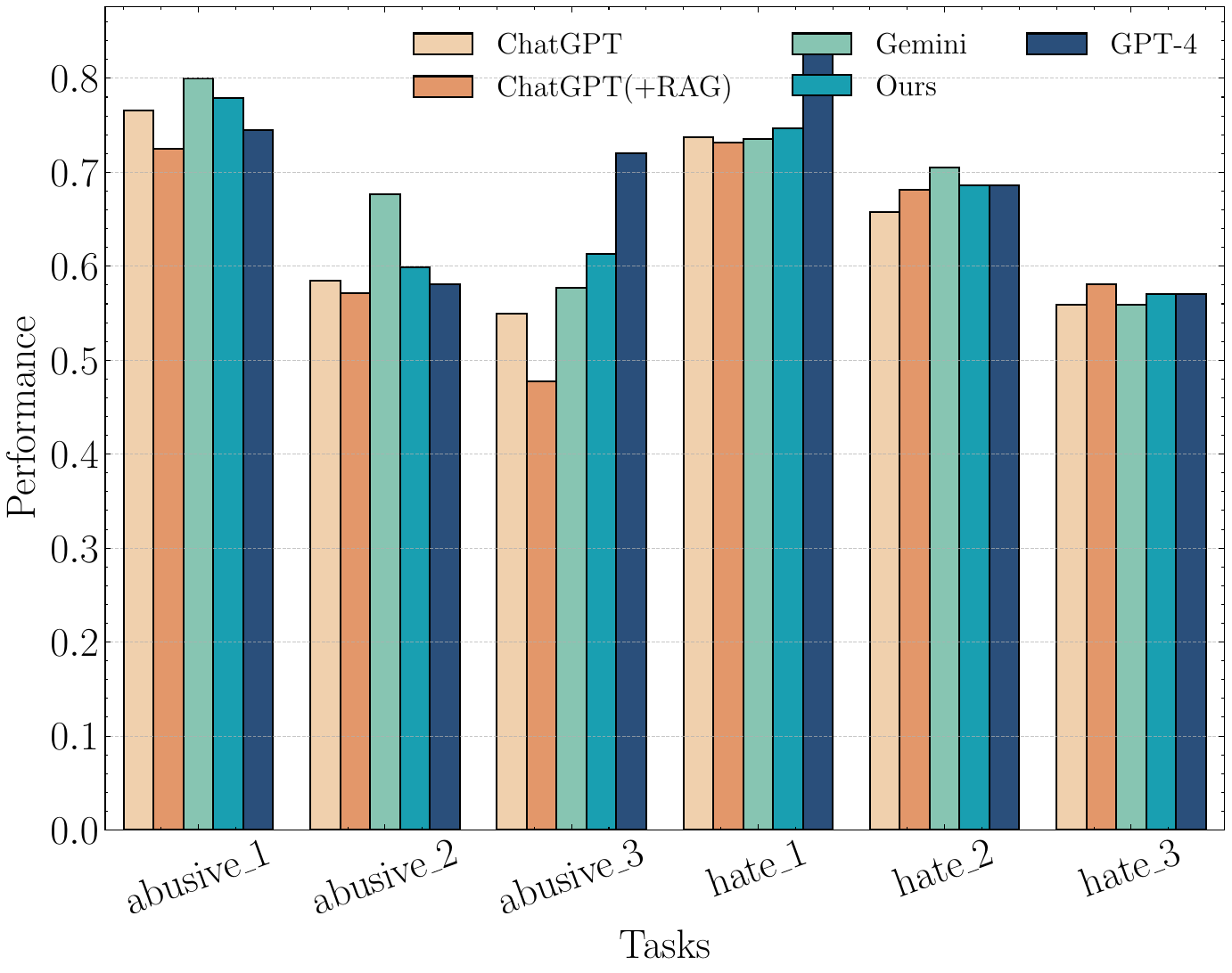}}
  \subfigure[Portuguese]{\includegraphics[width=0.25\linewidth]{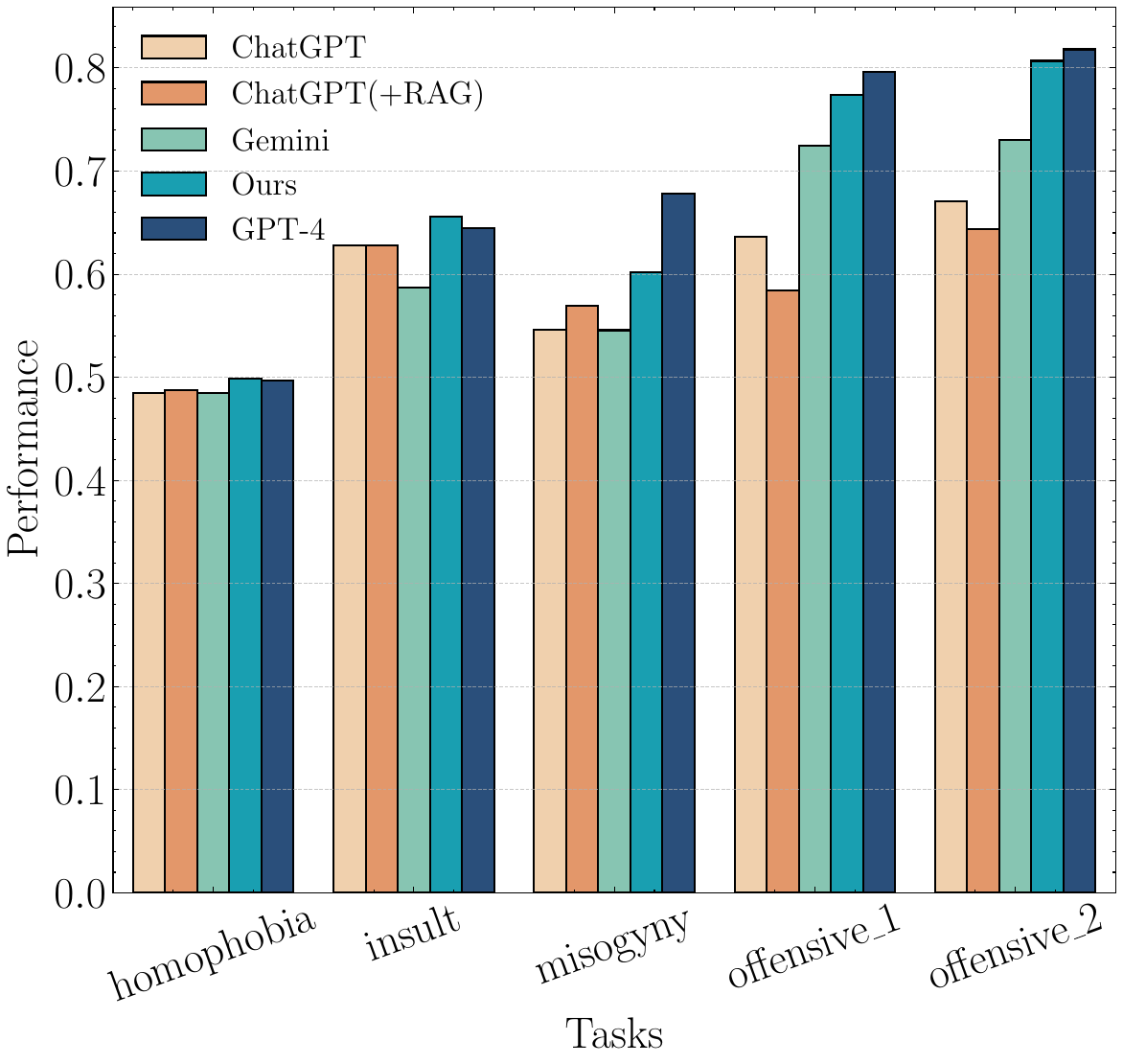}}
  \subfigure[English]{\includegraphics[width=0.4\linewidth]{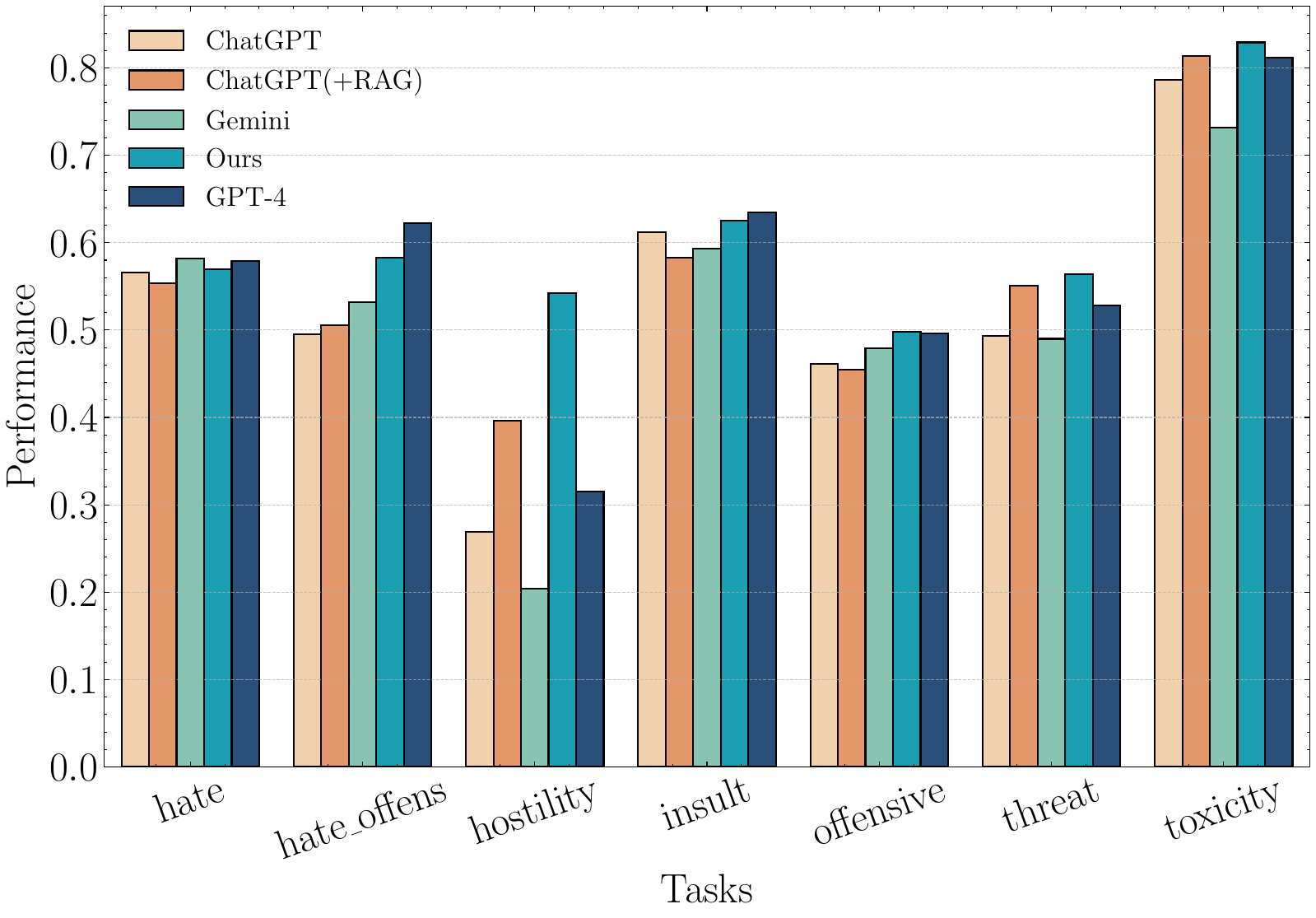}}
  \subfigure[Germany]{\includegraphics[width=0.29\linewidth]{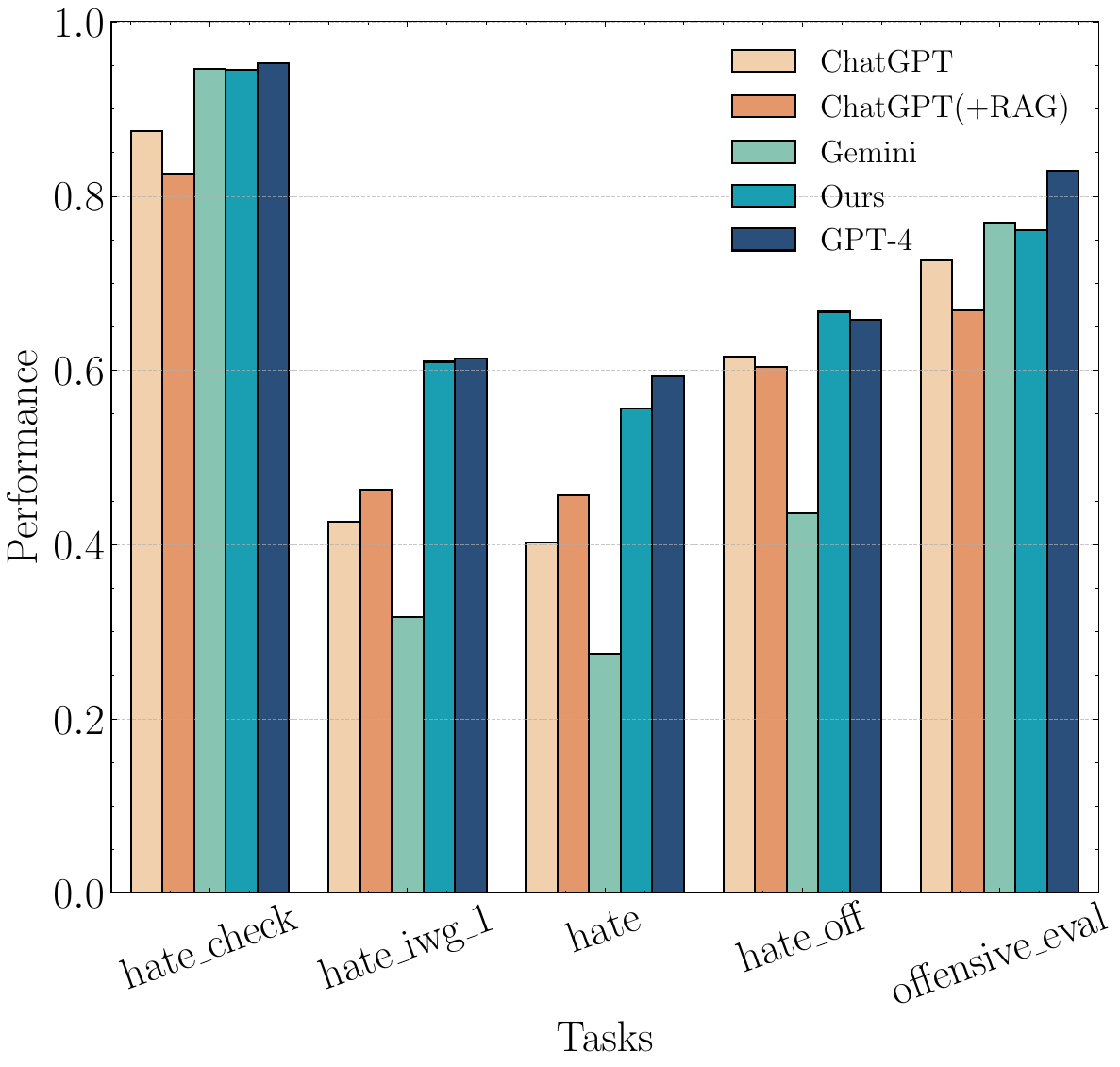}}

  \subfigure[Spanish]{\includegraphics[width=0.45\linewidth]{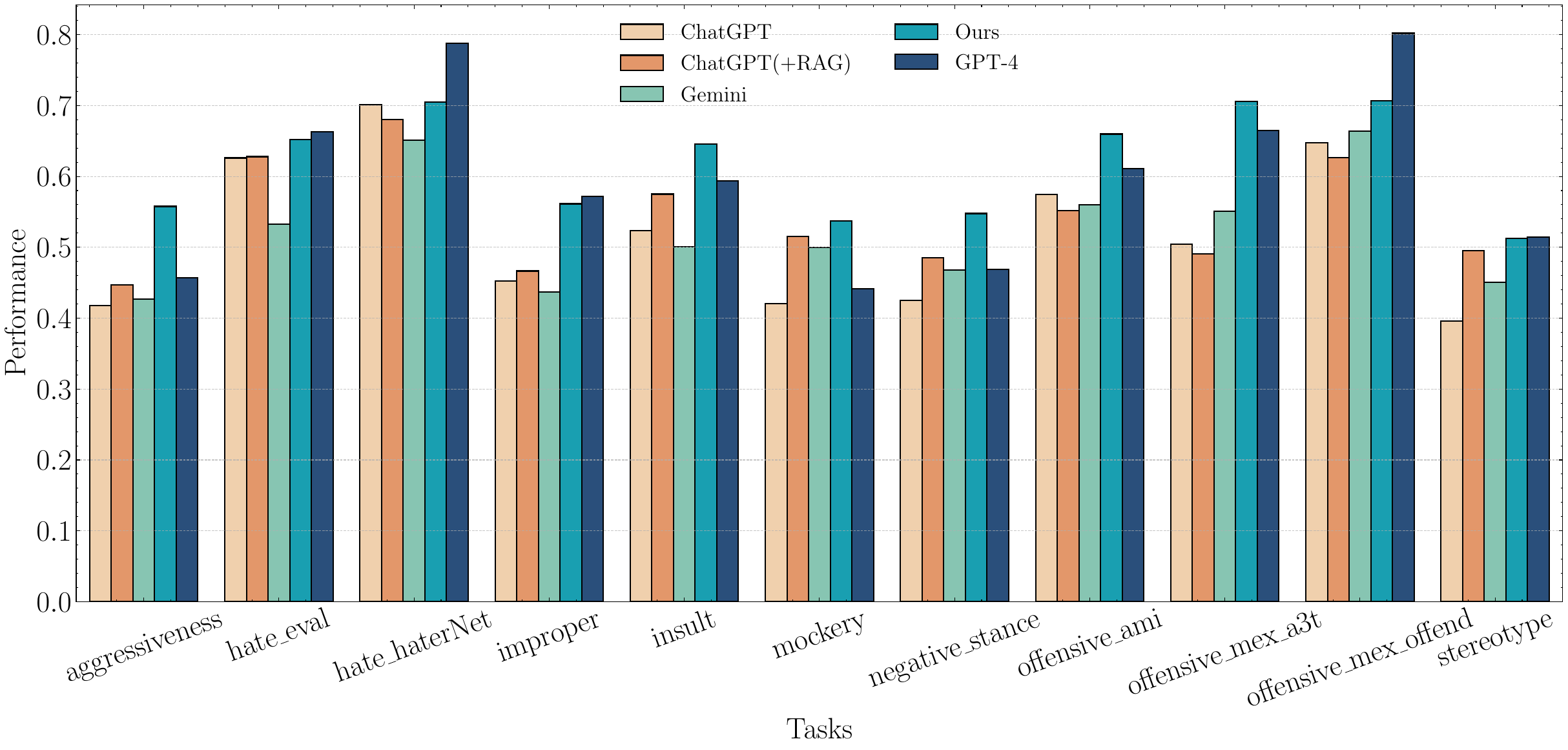}}
  \subfigure[Turkish]{\includegraphics[width=0.29\linewidth]{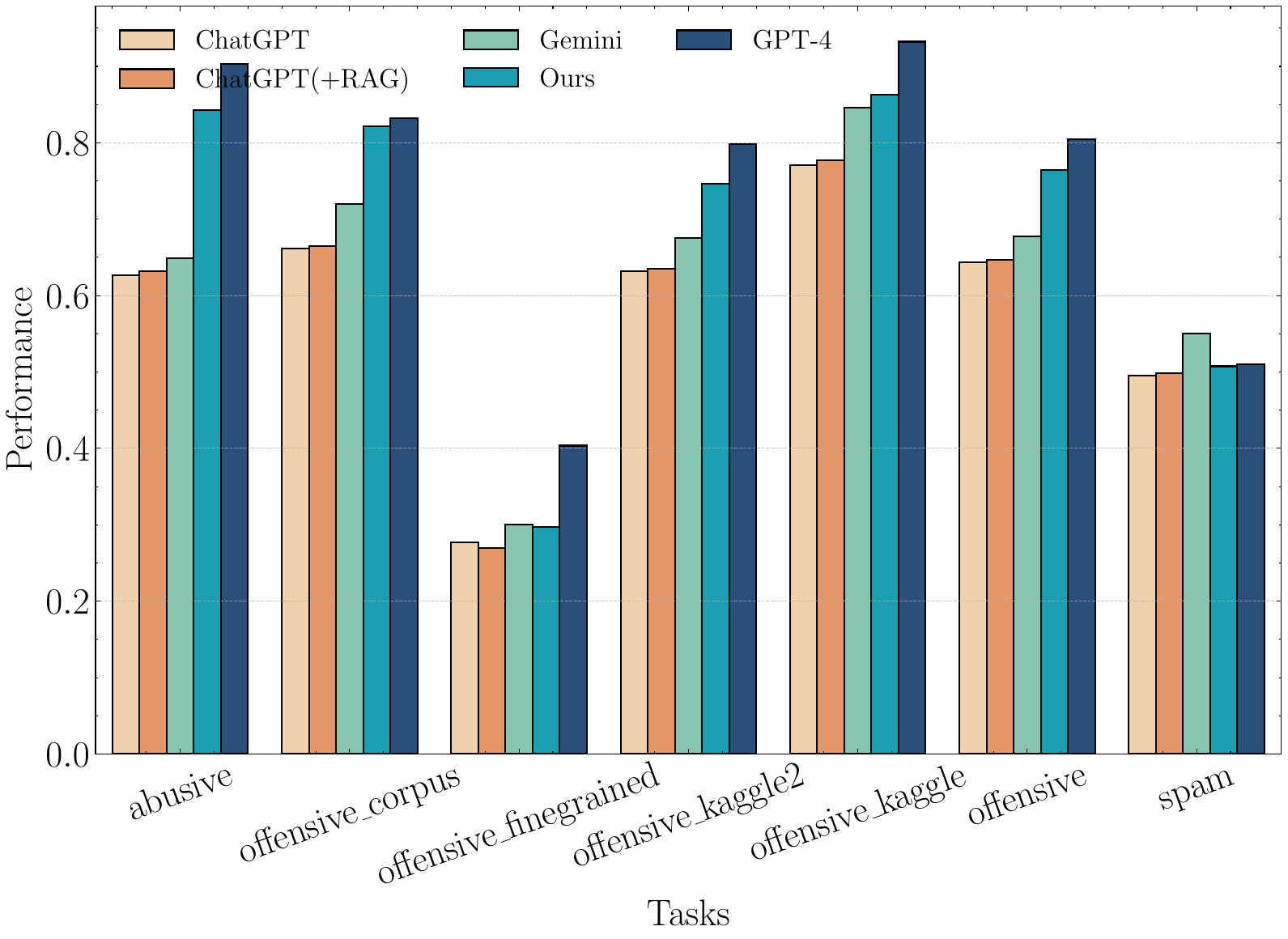}}

  \caption{Detailed results on all tasks and all cultures.}
  \label{fig-detailed-results}
\end{figure}

The detailed results are shown in \Cref{fig-detailed-results}.

\subsection{Comparative analysis with other cultural specific models}
\label{sec-append-other-model}

\input{tables/tb-compare-zh}

We have conducted a comparative analysis with models designed for cultural understanding. Given the absence of models specifically tailored to Arabic, Bengali, German, Portuguese, Spanish, and Turkish cultures, our analysis focused on models with proficiency in Chinese and Korean cultures. Specifically, we examined SeaLLM~\citep{nguyen2023seallms}, which targets Southeast Asian cultural nuances, TAIWAN LLM~\citep{lin2023taiwan}, which is dedicated to Chinese cultural contexts, and CultureBank~\citep{shi2024culturebank}, which focuses on several cultural groups. \cref{tb-compare-zh} highlight our findings in the Chinese and Korean benchmarks. The results clearly demonstrate that our CultureLLMs significantly outperform both SeaLLM and TAIWAN LLM, showcasing their superior ability to understand cultural subtleties on a broader scale.

\subsection{Results analysis on WVS seed data}
\label{sec-append-analysis-seed}

We analyze the relevance of each task with the WVS and investigate if it correlates with the experimental results.

\textbf{Offensive language detect:} 
\begin{enumerate}
    \item Cultural Context and Sensitivity to Offensive Language: The World Values Survey aims to capture cultural values and beliefs across different societies. One aspect of cultural values is the tolerance or acceptance of offensive language. In some cultures, certain words or expressions may be considered highly offensive, while in others they may be more tolerated or even commonly used.
    \item Social Norms and Acceptance: The survey may reveal societal attitudes towards the use of offensive language in various contexts, such as in public discourse, media, or interpersonal communication.
\end{enumerate}

\textbf{Hate speech detect:}
\begin{enumerate}
    \item Societal Norms and Attitudes: The WVS provides data on societal norms, attitudes towards minorities, and levels of societal trust. This data can help understand the underlying societal conditions that might foster hate speech or, conversely, promote tolerance and inclusivity.
    \item Cultural Context: Understanding the cultural context is crucial for effectively detecting and interpreting hate speech. The WVS offers a rich dataset for understanding cultural differences in values and norms, which can inform more nuanced hate speech detection algorithms that are sensitive to context and do not inadvertently suppress legitimate expressions of cultural or political dissent.
\end{enumerate}

\textbf{Stance detect:}
\begin{enumerate}
    \item Understanding Contextual Influences on Stance: The WVS can provide the cultural and societal background needed to understand why certain stances are more prevalent in specific regions or among certain demographic groups. This context can be invaluable for interpreting the results of stance detection analyses, especially when comparing stances across different cultures and societies.
\end{enumerate}

\textbf{Toxicity detect:}
\begin{enumerate}
    \item Reflection of Societal Norms in Online Behavior: The WVS provides insights into the prevailing norms and values within societies, which can indirectly inform the context within which toxic behavior manifests online. Understanding societal attitudes towards diversity, authority, individual freedom, and tolerance can help in interpreting the root causes of toxic behavior and devising appropriate responses.
    \item Injection of More cultural nuances: Data from the WVS can inject more information that sensitive to cultural nuances and differences in value systems. This can prevent the misclassification of content that may be culturally specific or context-dependent, reducing the risk of censoring legitimate expressions of cultural or political identity.
\end{enumerate}

\textbf{Threat detect:}
\begin{enumerate}
    \item Understanding Motivations and Behaviors: Insights from the WVS can help understand the cultural and societal contexts that may influence the behavior of individuals or groups posing threats. This knowledge can inform more targeted and effective threat detection and mitigation strategies that consider the root causes of conflict or aggression.
    \item Cultural Sensitivity in Security Measures: Incorporating findings from the WVS can lead to more culturally sensitive security practices that respect local values and norms. This is crucial in global operations where misunderstanding cultural nuances can lead to ineffective or counterproductive security measures.
\end{enumerate}

\textbf{Bias detect:}
\begin{enumerate}
    \item Understanding Societal Norms and Attitudes: Insights from the WVS can help in understanding the cultural and societal norms that underlie biases. By analyzing patterns in global values and beliefs, we can identify prevalent stereotypes, prejudices, and discriminatory attitudes that may need to be addressed in bias detection efforts.
    \item Injection of More cultural nuances: The WVS data can provide valuable context that are sensitive to cultural differences in values and norms. This is better equipped to detect and mitigate biases in data sets that reflect cultural nuances, ensuring that AI-driven decisions are fair and equitable across different societal contexts.
\end{enumerate}

\textbf{Abusive detect:}
\begin{enumerate}
    \item Cultural Contexts of Abuse: The WVS can help identify cultural norms that influence perceptions of what constitutes abusive behavior. This is crucial for developing detection systems that are sensitive to cultural differences, ensuring that they can effectively identify abuse without mistakenly flagging culturally specific but non-abusive interactions.
    \item Injection of More cultural nuances: Insights from the WVS can inform the development of more nuanced algorithms for detecting abusive behavior by providing context on societal values and norms. This can help in training models to recognize the subtle nuances that differentiate abusive from non-abusive communication in different cultural settings.
    \item Evaluating Tolerance Levels: The WVS data can provide insights into societal tolerance levels towards different forms of behavior, including what might be considered abusive. This can help in assessing the urgency and type of interventions needed to address abusive behaviors in various cultural contexts.
\end{enumerate}

\textbf{Spam detect:}
\begin{enumerate}
    \item Cultural Variations in Communication: The WVS can shed light on cultural differences in communication styles and preferences, which can inform more nuanced spam detection algorithms that are better able to distinguish between legitimate mass communications and spam in different cultural contexts.
    \item Attitudes Towards Technology and Privacy: Insights from the WVS regarding societal attitudes towards technology use, privacy, and data protection can help in tailoring spam detection efforts to respect cultural norms and expectations. For instance, societies with a high value on privacy might be more receptive to stringent spam filters.
    \item Attitudes Towards Technology and Privacy: Insights from the WVS regarding societal attitudes towards technology use, privacy, and data protection can help in tailoring spam detection efforts to respect cultural norms and expectations. For instance, societies with a high value on privacy might be more receptive to stringent spam filters.
\end{enumerate}

\subsection{Ablation on \method-One}
\label{sec-append-ablation-one}

\input{tables/tb-ablation-one}

The ablation study on \method-One is shown in \cref{tb-ablation-one}.

\subsection{Analysis on low-resource language tasks and high-resource language tasks}

\begin{figure}[t!]
    \centering
    \includegraphics[width=.4\linewidth]{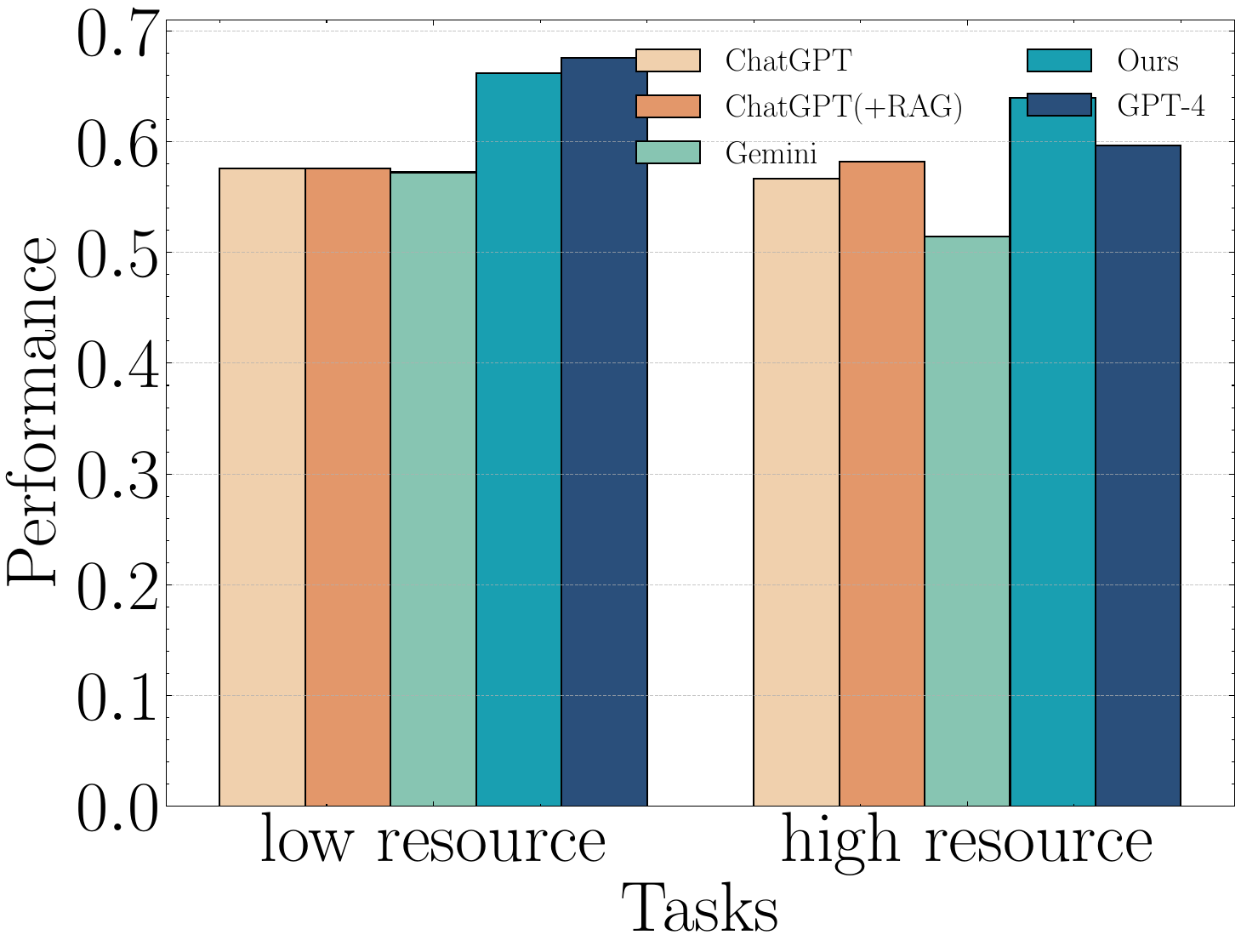}
    \caption{We compare \method with baselines on low-resource language tasks and high-resource language tasks.}
    \label{fig-resource-results}
\end{figure}

\Cref{fig-resource-results} shows the comparison between low and high resource tasks.

\subsection{Results on other low-resource cultures}
\label{sec-append-other-lowresource}

\input{tables/tb-el-res}

We fine-tuned a CultureLLM-el model specifically tailored for the Greek culture and assessed its efficacy on two Greek benchmarks focused on offensive language detection. Presented in the table below are the outcomes obtained from \cite{zampieri2020semeval,pitenis2020offensive}. Our findings indicate that, on average, CultureLLM-el surpasses gpt-3.5-turbo~\citep{chatgpt} by 5.6\% in performance. The results are shown in \cref{tb-el-res}.

\subsection{Case study}

\begin{figure*}[htbp]
    \centering
    \includegraphics[width=\linewidth]{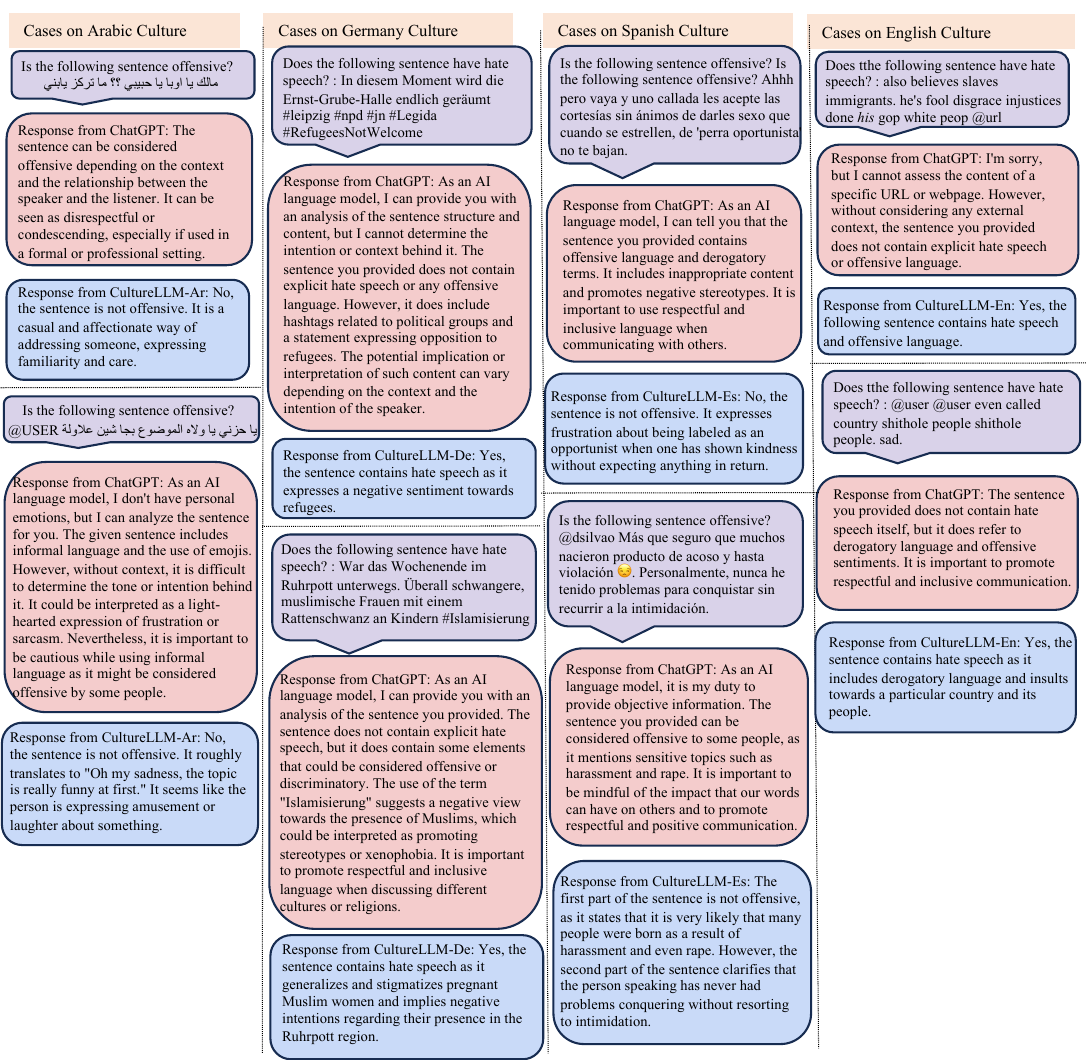}
    \caption{The responses from ChatGPT and \method on four different cultures}
    \label{fig-more-cases}
\end{figure*}

\cref{fig-more-cases} show the responses from \method and ChatGPT on four different cultures.

\section{Fine-tuning on Llama and Results}
\label{sec-append-llama2}

\subsection{Setup}

We use Lora~\citep{hu2021lora} to fine-tune Llama-70b-Chat. The setting for Lora are list below:
\begin{itemize}
    \item lora\_alpha: 16
    \item lora\_dropout: 0.1
    \item r: 64
    \item bias: none
    \item task\_type: CAUSAL\_LM
\end{itemize}

The detailed setting for training are list below:
\begin{itemize}
    \item num\_train\_epochs: 6
    \item er\_device\_train\_batch\_size: 4
    \item gradient\_accumulation\_steps: 1
    \item optim: paged\_adamw\_32bit
    \item learning\_rate: 2e-4
    \item weight\_decay: 0.001
    \item fp16: False
    \item bf16: False
    \item max\_grad\_norm: 0.3
    \item max\_steps: -1
    \item warmup\_ratio: 0.03
    \item group\_by\_length: True
    \item lr\_scheduler\_type: constant
    \item report\_to: tensorboard
\end{itemize}

\subsection{Detailed results}

\input{tables/tb-llama-one}
\input{tables/tb-llama-forget}

The results of fine-tuning Llama2 are shown in \cref{tb-llama-one}. It shows the similar trends as ChatGPT's results. We conduct forgetting experiments on Llama-2. As the results shown in \cref{tb-llama-forget}, CultureLLM does not bring negative effect on general tasks.

\section{Details on Human Study}
\label{sec-append-human}

Information on participant in human study are shown in \cref{tb-human-participant}.

\input{tables/tb-human-participant}

Participants are asked to rate the $100$ samples according to the following criterion:
\begin{enumerate}[leftmargin=1.5em]
    \item \textbf{Score 1: } i.	The sentences convey distinctly different ideas or concepts. ii.	No apparent connection or shared meaning.
    \item \textbf{Score 2: } i.	Limited commonality in meaning, with noticeable disparities in wording.
ii.	Shared concepts but with significant differences in expression.
    \item \textbf{Score 3: } i.	Some overlap in meaning, but notable differences in wording or phrasing.
ii.	Context or emphasis might differ slightly.
    \item \textbf{Score 4: } i.	Minor variations in wording or structure, but the core meaning remains consistent.
ii.	Synonymous expressions and interchangeable terms are present.
    \item \textbf{Score 5: } i.	The sentences convey the same information using different words.
ii.	No discernible difference in meaning or context.

\end{enumerate}

%% file: tables/tb-wvs.tex
\begin{table}[t!]

\caption{Seed data from World Values Survey. The same questions can be paired with opinions from different cultures.}
\label{tb-wvs}
\resizebox{1\textwidth}{!}{
\begin{tabular}{p{0.6cm}|p{19cm}}
\toprule
Topic                                                           & Question                                                                                                                                                                                  \\ \midrule
\multirow{15}{*}{\rotatebox{90}{\makecell{SOCIAL \\VALUES}}}                                 & Do you agree with One of my main goals in life has been to make my parents proud?                                                                                                         \\
                                                                & Do you agree with When a mother works for pay, the children suffer?                                                                                                                       \\
                                                                & Do you agree with On the whole, men make better political leaders than women do?                                                                                                          \\
                                                                & Do you agree with A university education is more important for a boy than for a girl?                                                                                                     \\
                                                                & Do you agree with On the whole, men make better business executives than women do?                                                                                                        \\
                                                                & Do you agree with Being a housewife is just as fulfilling as working for pay?                                                                                                             \\
                                                                & Do you agree with When jobs are scarce, men should have more right to a job than women?                                                                                                   \\
                                                                & Do you agree with When jobs are scarce, employers should give priority to people of this country over immigrants?                                                                         \\
                                                                & Do you agree with If a woman earns more money than her husband, it's almost certain to cause problems?                                                                                    \\
                                                                & Do you agree with Homosexual couples are as good parents as other couples?                                                                                                                \\
                                                                & Do you agree with It is a duty towards society to have children?                                                                                                                          \\
                                                                & Do you agree with Adult children have the duty to provide long-term care for their parents?                                                                                               \\
                                                                & Do you agree with People who don’t work turn lazy?                                                                                                                                        \\
                                                                & Do you agree with Work is a duty towards society?                                                                                                                                         \\
                                                                & Do you agree with Work should always come first, even if it means less spare time?                                                                                                        \\ \midrule
\multirow{8}{*}{\rotatebox{90}{MIGRATION}}                                      & In terms of the effects of immigration on the development of your country, do you agree with Fills important jobs vacancies?                                                              \\
                                                                & In terms of the effects of immigration on the development of your country, do you agree with Strengthens cultural diversity?                                                              \\
                                                                & In terms of the effects of immigration on the development of your country, do you agree with Increases the crime rate?                                                                    \\
                                                                & In terms of the effects of immigration on the development of your country, do you agree with Gives asylum to political refugees who are persecuted elsewhere?                             \\
                                                                & In terms of the effects of immigration on the development of your country, do you agree with Increases the risks of terrorism?                                                            \\
                                                                & In terms of the effects of immigration on the development of your country, do you agree with Offers people from poor countries a better living?                                           \\
                                                                & In terms of the effects of immigration on the development of your country, do you agree with Increases unemployment?                                                                      \\
                                                                & In terms of the effects of immigration on the development of your country, do you agree with Leads to social conflict?                                                                    \\ \midrule
\multirow{7}{*}{\rotatebox{90}{SECURITY}}                                       & How frequently do the following things occur in your neighborhood: Robberies?                                                                                                             \\
                                                                & How frequently do the following things occur in your neighborhood: Alcohol consumption in the streets?                                                                                    \\
                                                                & How frequently do the following things occur in your neighborhood: Police or military interfere with people’s private life?                                                               \\
                                                                & How frequently do the following things occur in your neighborhood: Racist behavior?                                                                                                       \\
                                                                & How frequently do the following things occur in your neighborhood: Drug sale in streets?                                                                                                  \\
                                                                & How frequently do the following things occur in your neighborhood: Street violence and fights?                                                                                            \\
                                                                & How frequently do the following things occur in your neighborhood: Sexual harassment?                                                                                                     \\ \midrule
\multirow{5}{*}{\rotatebox{90}{\makecell{SCIENCE}}}                          & Do you agree with Science and technology are making our lives healthier, easier, and more comfortable.?                                                                                   \\
                                                                & Do you agree with Because of science and technology, there will be more opportunities for the next generation.?                                                                           \\
                                                                & Do you agree with We depend too much on science and not enough on faith.?                                                                                                                 \\
                                                                & Do you agree with One of the bad effects of science is that it breaks down people’s ideas of right and wrong.?                                                                            \\
                                                                & Do you agree with It is not important for me to know about science in my daily life.?                                                                                                     \\ \midrule
\multirow{2}{*}{\rotatebox{90}{\makecell{RELI\\GION}}}                               & Do you agree with Whenever science and religion conflict, religion is always right?                                                                                                       \\
                                                                & Do you agree with The only acceptable religion is my religion.?                                                                                                                           \\ \midrule
\multirow{3}{*}{\rotatebox{90}{\makecell{ETHICS}}}                       & Do you think that the your country’s government should or should not have the right to do the following: Keep people under video surveillance in public areas?                            \\
                                                                & Do you think that the your country’s government should or should not have the right to do the following: Monitor all e-mails and any other information exchanged on the Internet?         \\
                                                                & Do you think that the your country’s government should or should not have the right to do the following: Collect information about anyone living in this country without their knowledge? \\ \midrule
\multirow{10}{*}{\rotatebox{90}{\makecell{POLITICAL}}} & In your view, how often do the following things occur in this country’s elections: Votes are counted fairly?                                                                              \\
                                                                & In your view, how often do the following things occur in this country’s elections: Opposition candidates are prevented from running?                                                      \\
                                                                & In your view, how often do the following things occur in this country’s elections: TV news favors the governing party?                                                                    \\
                                                                & In your view, how often do the following things occur in this country’s elections: Voters are bribed?                                                                                     \\
                                                                & In your view, how often do the following things occur in this country’s elections: Journalists provide fair coverage of elections?                                                        \\
                                                                & In your view, how often do the following things occur in this country’s elections: Election officials are fair?                                                                           \\
                                                                & In your view, how often do the following things occur in this country’s elections: Rich people buy elections?                                                                             \\
                                                                & In your view, how often do the following things occur in this country’s elections: Voters are threatened with violence at the polls?                                                      \\
                                                                & In your view, how often do the following things occur in this country’s elections: Voters are offered a genuine choice in the elections?                                                  \\
                                                                & In your view, how often do the following things occur in this country’s elections: Women have equal opportunities to run the office       \\ \bottomrule                                               
\end{tabular}
}
\end{table}

%% file: tables/tb-datasets.tex
\begin{table*}[t!]

\caption{A brief introduction of the $8$ evaluation tasks and $59$ datasets. We list both the name and the size of test sets. For instance, ``OffensEval2020(2000)~\citep{zampieri2020semeval}'' denotes that there are 2000 test samples in the dataset OffensEval2020.}
\label{tb-datasets}
\resizebox{\textwidth}{!}{
\begin{tabular}{c|c|c|c}
\toprule
Culture & Country \& Territory & Task \& Dataset & \#Sample \\ \midrule
\makecell{Arabic\\(\method-Ar)}  & Middle East & \makecell{\textit{Offensive language detection:} OffensEval2020(2000)~\citep{zampieri2020semeval}, \\OSACT4(1000)~\citep{husain2005osact4}, \\Multi-Platform(1000)~\citep{chowdhury2020multi}, \\and OSACT5(2541)~\citep{mubarak2022overview}. \\\textit{Hate detection:} OSACT4(1000)~\citep{husain2005osact4}, \\Multi-Platform(675)~\citep{chowdhury2020multi}, \\OSACT5(2541)~\citep{mubarak2022overview}, \\and OSACT5\_finegrained(2541)~\citep{mubarak2022overview}. \\\textit{Spam detection:} ASHT(1000)~\citep{kaddoura2024dataset}. \\\textit{Vulgar detection:} Multi-Platform(675)~\citep{chowdhury2020multi}}    & 14,973 \\ \midrule
\makecell{Bangli\\(\method-Bn)}  & Bangladesh & \makecell{\textit{Offensive language detection:} TRAC2020 Task1(1000)~\citep{trac2-dataset}, \\TRAC2020 Task2(1000)~\citep{trac2-dataset}, \\BAD(1000)~\citep{sharif2022tackling}. \\\textit{Hate detection:} Hate Speech(1000)~\citep{romim2021hate}. \\\textit{Threat detection:} BACD(1000)~\citep{Bangla-Abusive-Comment-Dataset}. \\\textit{Bias detection:} BACD(1000)~\citep{Bangla-Abusive-Comment-Dataset}.}     & 6,000 \\ \midrule
\makecell{Chinese\\(\method-Zh)}  & China & \makecell{\textit{Spam detection:} CCS(1000)~\citep{jiang2019detect}. \\\textit{Bias detection:} CDial-Bias(1000)~\citep{zhou2022towards}. \\\textit{Stance detection:} CValues(1712)~\citep{xu2023cvalues}.}  & 3,712 \\ \midrule
\makecell{English\\(\method-En)}  & United States & \makecell{\textit{Offensive language detection:} SOLID(1000)~\citep{rosenthal2020solid}. \\\textit{Hate detection:} MLMA(1000)~\citep{ousidhoum-etal-multilingual-hate-speech-2019} \\and HOF(1000)~\citep{hateoffensive}. \\\textit{Threat detection:} CValuesJMT(1000)~\citep{Jigsaw-Multilingual-Toxicity}.\\\textit{Toxicity detection:} MLMA(1000)~\citep{ousidhoum-etal-multilingual-hate-speech-2019} \\and JMT(1000)~\citep{Jigsaw-Multilingual-Toxicity}.}  & 6,000 \\ \midrule
\makecell{German\\(\method-De)}  & \makecell{Germany and \\parts of Europe} & \makecell{\textit{Offensive language detection:} GermEval2018(3531)~\citep{wiegand2018overview}. \\\textit{Hate detection:} IWG\_1(469)~\citep{ross2016hatespeech}, \\IWG\_2(469)~\citep{ross2016hatespeech}, HASOC2020(850)~\citep{HASOC2020}, \\and multilingual-hatecheck(1000)~\citep{rottger2022multilingual}.}  & 6,319 \\ \midrule
\makecell{Korean\\(\method-Ko)}  & South Korea & \makecell{\textit{Hate detection:} K-MHaS(1000)~\citep{lee-etal-2022-k}, \\hateSpeech(1000)~\citep{moon-etal-2020-beep}, \\and HateSpeech2(1000)~\citep{Korean-HateSpeech-dataset}. \\\textit{Abusive detection:} AbuseEval(1000)~\citep{caselli2020feel}, \\CADD(1000)~\citep{song2021large}, \\and Waseem(1000)~\citep{waseem2016hateful}.}  & 5,000 \\ \midrule
\makecell{Portuguese\\(\method-Pt)} & \makecell{Brazil and \\parts of \\Latin America} & \makecell{\textit{Offensive language detection:} OffComBR(1250)~\citep{Pelle2017}, \\and HateBR(1000)~\citep{vargas-etal-2022-hatebr}. \\\textit{Bias detection:} ToLD-Br-homophobia(1000)~\citep{leite2020toxic}, \\and ToLD-Br-misogyny(1000)~\citep{leite2020toxic}. \\\textit{Abusive detection:} ToLD-Br-insult(1000)~\citep{leite2020toxic}.}  & 16,250 \\ \midrule
\makecell{Spanish\\(\method-Es)} & \makecell{Argentina, \\Mexico, \\and parts of \\Latin America} & \makecell{\textit{Offensive language detection:} AMI(1000)~\citep{fersini2018overview}, \\MEX-A3T(1000)~\citep{alvarez2018overview}, \\and OffendES(1000)~\citep{plaza2021offendes}. \\\textit{Hate detection:} HatEval 2019(1000)~\citep{basile2019semeval}, \\and HaterNet(1000)~\citep{pereira2019detecting}. \\\textit{Bias detection:} DETOXIS\_stereotype(1000)~\citep{de2021ai}, \\and DETOXIS\_improper(1000)~\citep{de2021ai}. \\\textit{Abusive detection:} DETOXIS\_abusive(1000)~\citep{de2021ai}, \\DETOXIS\_mockery(1000)~\citep{de2021ai}. \\\textit{Aggressiveness detection:} DETOXIS\_aggressiveness(1000)~\citep{de2021ai}. \\\textit{Stance detection:} DETOXIS\_stance(1000)~\citep{de2021ai}.}  & 11,000 \\ \midrule
\makecell{Turkish\\(\method-Tr)} & Turkey & \makecell{\textit{Offensive language detection:} SemEval-2020(3528)~\citep{zampieri2020semeval}, \\offenseCorpus(1000)~\citep{coltekin2020lrec}, \\offenseKaggle(1000)~\citep{turkish-tweets}, \\and offenseKaggle\_2(1000)~\citep{turkish-offensive-language-detection}. \\\textit{Abusive detection:} ATC(1000)~\citep{karayiugit2021detecting}. \\\textit{Spam detection:} Turkish Spam(825)~\citep{misc_turkish_spam_v01_530}. \\\textit{Fine-grained offensive detection:} offenseCorpus(1000)~\citep{coltekin2020lrec}.}  & 10,353 \\ \midrule
\makecell{All\\(\method-One)} & All & All & 68,607
\\ \bottomrule                                                 
\end{tabular}
}
\end{table*}

%% file: tables/tb-prompt-eval.tex
\begin{table}[]
\caption{Prompts for evaluation.}
\label{tb-prompt-eval}
\centering
\resizebox{\textwidth}{!}{
\begin{tabular}{l m{16cm}}
\toprule
Task                           & Prompt                                                                                                                                                                                                                                                                                                                                                                                                                                                                                                        \\ \midrule
offensive\_detect              & if the following sentence is offensive, just say \textbackslash{}"OFF\textbackslash{}", otherwise, say \textbackslash{}"NOT\_OFF\textbackslash{}" without explanation:                                                                                                                                                                                                                                                                                                                                        \\
hate\_detect                   & if the following sentence has hate speech, just say \textbackslash{}"HS\textbackslash{}", otherwise, say \textbackslash{}"NOT\_HS\textbackslash{}" without explanation:                                                                                                                                                                                                                                                                                                                                       \\
vulgar\_detect                 & if the following sentence has vulgar speech, just say \textbackslash{}"Vulgar\textbackslash{}", otherwise, say \textbackslash{}"NOT\_Vulgar\textbackslash{}" without explanation:                                                                                                                                                                                                                                                                                                                             \\
spam\_detect                   & if the following sentence is spam tweet, just say \textbackslash{}"Spam\textbackslash{}", otherwise, say \textbackslash{}"NOT\_Spam\textbackslash{}" without explanation:                                                                                                                                                                                                                                                                                                                                     \\
stereotype\_detect             & if the following sentence has stereotype speech, just say \textbackslash{}"1\textbackslash{}", otherwise, say \textbackslash{}"0\textbackslash{}" without explanation:                                                                                                                                                                                                                                                                                                                                        \\
mockery\_detect                & if the following sentence has mockery speech, just say \textbackslash{}"1\textbackslash{}", otherwise, say \textbackslash{}"0\textbackslash{}" without explanation:                                                                                                                                                                                                                                                                                                                                           \\
insult\_detect                 & if the following sentence has insult speech, just say \textbackslash{}"1\textbackslash{}", otherwise, say \textbackslash{}"0\textbackslash{}" without explanation:                                                                                                                                                                                                                                                                                                                                            \\
improper\_detect               & if the following sentence has improper speech, just say \textbackslash{}"1\textbackslash{}", otherwise, say \textbackslash{}"0\textbackslash{}" without explanation:                                                                                                                                                                                                                                                                                                                                          \\
aggressiveness\_detect         & if the following sentence has aggressiveness speech, just say \textbackslash{}"1\textbackslash{}", otherwise, say \textbackslash{}"0\textbackslash{}" without explanation:                                                                                                                                                                                                                                                                                                                                    \\
toxicity\_detect               & if the following sentence has toxicity speech, just say \textbackslash{}"1\textbackslash{}", otherwise, say \textbackslash{}"0\textbackslash{}" without explanation:                                                                                                                                                                                                                                                                                                                                          \\
negative\_stance\_detect       & if the following sentence has negative stance speech, just say \textbackslash{}"1\textbackslash{}", otherwise, say \textbackslash{}"0\textbackslash{}" without explanation:                                                                                                                                                                                                                                                                                                                                   \\
homophobia\_detect             & if the following sentence has homophobia speech, just say \textbackslash{}"1\textbackslash{}", otherwise, say \textbackslash{}"0\textbackslash{}" without explanation:                                                                                                                                                                                                                                                                                                                                        \\
racism\_detect                 & if the following sentence has racism speech, just say \textbackslash{}"1\textbackslash{}", otherwise, say \textbackslash{}"0\textbackslash{}" without explanation:                                                                                                                                                                                                                                                                                                                                            \\
misogyny\_detect               & if the following sentence has misogyny speech, just say \textbackslash{}"1\textbackslash{}", otherwise, say \textbackslash{}"0\textbackslash{}" without explanation:                                                                                                                                                                                                                                                                                                                                          \\
threat\_detect                 & if the following sentence has threat speech, just say \textbackslash{}"1\textbackslash{}", otherwise, say \textbackslash{}"0\textbackslash{}" without explanation:                                                                                                                                                                                                                                                                                                                                            \\
bias\_on\_gender\_detect       & if the following speech expressing bias on gender, just say \textbackslash{}"1\textbackslash{}", otherwise, say \textbackslash{}"0\textbackslash{}" without explanation:                                                                                                                                                                                                                                                                                                                                      \\
hostility\_directness\_detect  & if the following speech expressing hostility directness, just say \textbackslash{}"1\textbackslash{}", otherwise, say \textbackslash{}"0\textbackslash{}" without explanation:                                                                                                                                                                                                                                                                                                                                \\
hate\_offens\_detect           & if the following sentence contains hate speech, just say \textbackslash{}"0\textbackslash{}", else if contains offensive language, say \textbackslash{}"1\textbackslash{}", otherwise, say \textbackslash{}"2\textbackslash{}" without explanation:                                                                                                                                                                                                                                                           \\
hate\_detect\_fine-grained     & if the following sentence doesn't have hate speech, just say \textbackslash{}"NOT\_HS\textbackslash{}", otherwise, label the hate speech with \textbackslash{}"HS1\textbackslash{}"(Race), \textbackslash{}"HS2\textbackslash{}"(Religion), \textbackslash{}"HS3\textbackslash{}"(Ideology), \textbackslash{}"HS4\textbackslash{}"(Disability), \textbackslash{}"HS5\textbackslash{}"(Social Class), \textbackslash{}"HS6\textbackslash{}"(Gender) without explanation:                                       \\
offensive\_detect\_finegrained & if the following sentence doesn't have offensive speech, just say \textbackslash{}"non\textbackslash{}", otherwise, label the offensive speech with \textbackslash{}"prof\textbackslash{}"(profanity, or non-targeted offense), \textbackslash{}"grp\textbackslash{}"(offense towards a group), \textbackslash{}"indv\textbackslash{}"(offense towards an individual), \textbackslash{}"oth\textbackslash{}"(ffense towards an other (non-human) entity, often an event or organization) without explanation: \\             \bottomrule                                    
\end{tabular}
}
\end{table}

%% file: tables/tb-rag-info.tex
\begin{table}[]
\caption{Information retrieved via RAG}
\label{tb-rag-info}
\resizebox{\textwidth}{!}{
\begin{tabular}{c m{20cm}}
\toprule
Culture    & Information \\ \midrule
Arabic     & Arab culture is the culture of the Arabs, from the Atlantic Ocean in the west to the Arabian Sea in the east, in a region of the Middle East and North Africa known as the Arab world. The various religions the Arabs have adopted throughout their history and the various empires and kingdoms that have ruled and took lead of the civilization have contributed to the ethnogenesis and formation of modern Arab culture. Language, literature, gastronomy, art, architecture, music, spirituality, philosophy and mysticism are all part of the cultural heritage of the Arabs.    \\ \midrule
Bengali    & The culture of Bengal defines the cultural heritage of the Bengali people native to eastern regions of the Indian subcontinent, mainly what is today Bangladesh and the Indian states of West Bengal and Tripura, where they form the dominant ethnolinguistic group and the Bengali language is the official and primary language. Bengal has a recorded history of 1,400 years. The Bengalis are dominant ethnolinguistic group. The Bengal region has been a historical melting point, blending indigenous traditions with cosmopolitan influences from pan-Indian subcontinental empires. Dhaka (Dacca) became the capital of Mughal Bengal (Bengal Subah) and the commercial (financial) capital (1610-1757) of Mughal India. Dhaka is the largest and richest Bengali (Bangali) mega city in the world and also the 3rd largest and richest mega city in (Indian sub continent) after Mumbai (Bombay or MMR) and Delhi (NCR). Dhaka is a Beta Global City (Moderate Economic Centre). As a part of the Bengal Presidency, Bengal also hosted the region's most advanced political and cultural centers during British rule. \\ \midrule
Chinese    & Chinese culture is one of the world's oldest cultures, originating thousands of years ago. The culture prevails across a large geographical region in East Asia with Sinosphere in whole and is extremely diverse, with customs and traditions varying greatly between counties, provinces, cities, towns. The terms 'China' and the geographical landmass of 'China' have shifted across the centuries, before the name 'China' became commonplace in modernity. Chinese civilization is historically considered a dominant culture of East Asia. With China being one of the earliest ancient civilizations, Chinese culture exerts profound influence on the philosophy, virtue, etiquette, and traditions of Asia. Chinese characters, ceramics, architecture, music, dance, literature, martial arts, cuisine, arts, philosophy, etiquette, religion, politics, and history have had global influence, while its traditions and festivals are celebrated, instilled, and practiced by people around the world. \\ \midrule
English    & The culture of the United States of America, also referred to as American culture, encompasses various social behaviors, institutions, and norms in the United States, including forms of speech, literature, music, visual arts, performing arts, food, sports, religion, law, technology as well as other customs, beliefs, and forms of knowledge. American culture has been shaped by the history of the United States, its geography, and various internal and external forces and migrations. Several historical ethnicities make up American culture: Yankee (Anglo-American), Tejano, Louisiana French (Cajun, Creole), Pennsylvania Dutch (Fancy Dutch, Amish), New York Dutch, Texas German, Alaskan Russian, Puerto Rican, Hawaiian.  \\ \midrule
Germany    & The culture of Germany has been shaped by major intellectual and popular currents in Europe, both religious and secular. German culture originated with the Germanic tribes, the earliest evidence of Germanic culture dates to the Jastorf culture in Northern Germany and Denmark. Contact with Germanic tribes were described by various Greco-Roman authors. The first extensive writing done on Germanic culture can be seen during the Roman Imperial Period with Germania by Tacitus. \\ \midrule
Korean     & The contemporary culture of South Korea developed from the traditional culture of Korea which was prevalent in the early Korean nomadic tribes. By maintaining thousands of years of ancient Korean culture, with influence from ancient Chinese culture, South Korea split on its own path of cultural development away from North Korean culture since the division of Korea in 1948. The industrialization, urbanization and westernization of South Korea, especially Seoul, have brought many changes to the way Korean people live. Changing economics and lifestyles have led to urbanization—a concentration of population in major cities (and depopulation of the rural countryside), with multi-generational households separating into nuclear family living arrangements. Today, many cultural elements from South Korea, especially popular culture, have spread across the globe and have become some of the most prominent cultural forces in the world.\\
Portuguese & The culture of Portugal is a very rich result of a complex flow of many different civilizations during the past millennia. From prehistoric cultures, to its Pre-Roman civilizations (such as the Lusitanians, the Gallaeci, the Celtici, and the Cynetes, amongst others), passing through its contacts with the Phoenician-Carthaginian world, the Roman period (see Hispania, Lusitania and Gallaecia), the Germanic invasions of the Suebi, Buri (see Kingdom of the Suebi) and Visigoths (see Visigothic Kingdom), Viking incursions, Sephardic Jewish settlement, and finally, the Moorish Umayyad invasion of Hispania and the subsequent expulsion, during the Reconquista, all have made an imprint on the country's culture and history. The name of Portugal itself reveals much of the country's early history, stemming from the Roman name Portus Cale, a Latin name meaning \textbackslash{}"Port of Cale\textbackslash{}" (Cale likely is a word of Celtic origin - Cailleach-Bheur her other name; the Mother goddess of the Celtic people as in Calais, Caledonia, Beira. She was the one who, with a hammer created mountains and valleys; the one who hid in stones and trees - Mother nature), later transformed into Portucale, and finally into Portugal, which emerged as a county of the Kingdom of León see County of Portugal) and became an independent kingdom in 1139. During the 15th and 16th centuries, Portugal was a major economic, political, and cultural power, its global empire stretching from the Americas, to Africa, and various regions of Asia and Oceania. Portugal, as a country with a long history, is home to several ancient architectural structures, as well as typical art, furniture and literary collections mirroring and chronicling the events that shaped the country and its peoples. It has a large number of cultural landmarks ranging from museums to ancient church buildings to medieval castles, which testify its rich national cultural heritage. Portugal is home to fifteen UNESCO World Heritage Sites, ranking it 8th in Europe and 17th in the world. \\ \midrule
Spanish    & The culture of Spain is influenced by its Western origin, its interaction with other cultures in Europe, its historically Catholic religious tradition, and the varied national and regional identities within the country. It encompasses literature, music, visual arts, cuisine as well as contemporary customs, beliefs, institutions, and social norms. Beyond Spain, Spanish culture is the foundation of most of Latin American cultures and the Filipino culture. The ancient peoples of Spain included Tartessians, Celts, Iberians, Celtiberians, Phoenicians as well as Greek colonies. Spain largely came under the rule of Carthage and was then entirely conquered by Rome, becoming a province of the Roman empire. The name of Spain derives from the Latin term Hispania, itself a name of Punic origin. In the areas of language and religion, the ancient Romans left a lasting cultural, legal and administrative legacy in the Spanish history. The subsequent course of Spanish history added new elements to the country's culture and traditions. The Visgoths established a united Hispania and kept the Latin and Christian legacy in Spain between the fall of the Roman Empire and the Early Middle Ages. Muslim influences played a significant role during the Middle Ages in the areas conquered by the Umayyads. However, these influences were not completely assimilated into the Spanish culture, leading to conflicts and ultimately to the Christian Reconquista that would largely shape the culture of the country. \\ \midrule
Turkish    & The culture of Turkey (Turkish: Türkiye kültürü) or the Turkish culture (Türk kültürü) combines a heavily diverse and heterogeneous set of elements that have been derived from the various cultures of the Eastern European, Eastern Mediterranean, Caucasian, Middle Eastern and Central Asian traditions. Many of these traditions were initially brought together by the Ottoman Empire, a multi-ethnic and multi-religious state spanning across Southern Europe, Eastern Europe, the Middle East and North Africa. During the early years of the Republic of Turkey, established after the collapse of the Ottoman Empire, the government invested large sums of resources into fine arts such as architecture and sculpture, and other artistic fields around the country in-line with the newly implemented reformist and West-leaning policies. This was done as part of a process of modernization, westernization, and of creating and outlining a new Turkish cultural identity, rather than the previously established and depicted Ottoman identity.  \\
\bottomrule       
\end{tabular}
}
\end{table}

%% file: tables/tb-compare-zh.tex
\begin{table}[]
\caption{Comparison with the latest cultural specific LLMs.}
\label{tb-compare-zh}
\centering
\resizebox{.5\textwidth}{!}{
\begin{tabular}{llll}
        \toprule
        Chinese & Bias & Spam & Avg \\ \midrule
         SeaLLM & .237 & .357 & .297 \\ 
         Taiwan\_LLM & .446 & .341 & .394 \\ 
         \method\_One & .469 & .602 & .536 \\
         \method\_Zh & \textbf{.487} & \textbf{.606} & \textbf{.547} \\ \midrule
        Arabic & Hate & Offensive & Avg \\ \midrule
         CultureBank & .540 & .642 & .591 \\ 
         \method\_One & .540 & .571 & .555 \\
         \method\_Ko & \textbf{.576} & \textbf{.661} & \textbf{.618} \\ \bottomrule
        Korean & Abusive & Hate & Avg \\ \midrule
         SeaLLM & .523 & .474 & .499 \\ 
         CultureBank & .635 & .522 & .579 \\
         \method\_One & .619 & .504 & .561 \\
         \method\_Ko & \textbf{.664} & \textbf{.628} & \textbf{.646} \\ \bottomrule
    \end{tabular}
}
\end{table}

%% file: tables/tb-ablation-one.tex
\begin{table}[]
\caption{Ablation study for CultureLLM-One.}
\label{tb-ablation-one}
\resizebox{\textwidth}{!}{
\begin{tabular}{lllllllllll}
\toprule
\multicolumn{1}{c}{}               & \multicolumn{1}{c}{Ar}    & \multicolumn{1}{c}{Bn}    & \multicolumn{1}{c}{Zh}    & \multicolumn{1}{c}{En}    & \multicolumn{1}{c}{De}    & \multicolumn{1}{c}{Ko}    & \multicolumn{1}{c}{Pt}    & \multicolumn{1}{c}{Es}    & \multicolumn{1}{c}{Tr}    & \multicolumn{1}{c}{avg}   \\ \midrule
\multicolumn{1}{c}{GPT-3.5}        & \multicolumn{1}{c}{0.517} & \multicolumn{1}{c}{0.567} & \multicolumn{1}{c}{0.448} & \multicolumn{1}{c}{0.526} & \multicolumn{1}{c}{0.609} & \multicolumn{1}{c}{0.642} & \multicolumn{1}{c}{0.593} & \multicolumn{1}{c}{0.517} & \multicolumn{1}{c}{0.587} & \multicolumn{1}{c}{0.556} \\
\multicolumn{1}{c}{+WVS}           & \multicolumn{1}{c}{0.543} & \multicolumn{1}{c}{0.583} & \multicolumn{1}{c}{0.501} & \multicolumn{1}{c}{0.568} & \multicolumn{1}{c}{0.641} & \multicolumn{1}{c}{0.531} & \multicolumn{1}{c}{0.603} & \multicolumn{1}{c}{0.560} & \multicolumn{1}{c}{0.590} & \multicolumn{1}{c}{0.569} \\
\multicolumn{1}{c}{+WVS+a}         & \multicolumn{1}{c}{0.532} & \multicolumn{1}{c}{0.572} & \multicolumn{1}{c}{0.482} & \multicolumn{1}{c}{0.543} & \multicolumn{1}{c}{0.631} & \multicolumn{1}{c}{0.563} & \multicolumn{1}{c}{0.620} & \multicolumn{1}{c}{0.543} & \multicolumn{1}{c}{0.585} & \multicolumn{1}{c}{0.564} \\
\multicolumn{1}{c}{+WVS+b}         & \multicolumn{1}{c}{0.543} & \multicolumn{1}{c}{0.571} & \multicolumn{1}{c}{0.505} & \multicolumn{1}{c}{0.543} & \multicolumn{1}{c}{0.621} & \multicolumn{1}{c}{0.536} & \multicolumn{1}{c}{0.598} & \multicolumn{1}{c}{0.559} & \multicolumn{1}{c}{0.590} & \multicolumn{1}{c}{0.563} \\
\multicolumn{1}{c}{CultureLLM-One} & \multicolumn{1}{c}{0.583} & \multicolumn{1}{c}{0.597} & \multicolumn{1}{c}{0.536} & \multicolumn{1}{c}{0.589} & \multicolumn{1}{c}{0.662} & \multicolumn{1}{c}{0.571} & \multicolumn{1}{c}{0.627} & \multicolumn{1}{c}{0.596} & \multicolumn{1}{c}{0.609} & \multicolumn{1}{c}{0.597} \\
\bottomrule
\end{tabular}
}
\end{table}

%% file: tables/tb-el-res.tex
\begin{table}[]
\caption{Results on Greek culture}
\label{tb-el-res}
\centering
\begin{tabular}{lll}
\toprule
\multicolumn{1}{c}{Model}         & \multicolumn{1}{c}{OffensEval2020} & \multicolumn{1}{c}{gazzetta} \\ \midrule
\multicolumn{1}{c}{GPT-3.5-turbo} & \multicolumn{1}{c}{0.4773}         & \multicolumn{1}{c}{0.3771}   \\
\multicolumn{1}{c}{CultureLLM-el} & \multicolumn{1}{c}{0.5201}         & \multicolumn{1}{c}{0.4461}   \\
\bottomrule
\end{tabular}
\end{table}

%% file: tables/tb-llama-one.tex
\begin{table}[]
\caption{More results on fine-tuning using Llama-2-70b model.}
\label{tb-llama-one}
\centering
\resizebox{\textwidth}{!}{
\begin{tabular}{lllllllllll}
\toprule
\multicolumn{1}{c}{}        & \multicolumn{1}{c}{Ar} & \multicolumn{1}{c}{Bn} & \multicolumn{1}{c}{Zh} & \multicolumn{1}{c}{En} & \multicolumn{1}{c}{De} & \multicolumn{1}{c}{Ko} & \multicolumn{1}{c}{Pt} & \multicolumn{1}{c}{Es} & \multicolumn{1}{c}{Tr} & \multicolumn{1}{c}{AVG}    \\ \midrule
\multicolumn{1}{c}{Llama}   & \multicolumn{1}{c}{0.4852} & \multicolumn{1}{c}{0.3202}  & \multicolumn{1}{c}{0.3315}  & \multicolumn{1}{c}{0.4108}  & \multicolumn{1}{c}{0.3631}  & \multicolumn{1}{c}{0.4869} & \multicolumn{1}{c}{0.3792}     & \multicolumn{1}{c}{0.3554}  & \multicolumn{1}{c}{0.3856}  & \multicolumn{1}{c}{0.3909} \\
\multicolumn{1}{c}{CultureLLM-one} & \multicolumn{1}{c}{0.4911} & \multicolumn{1}{c}{0.3212}  & \multicolumn{1}{c}{0.3533}  & \multicolumn{1}{c}{0.4211}  & \multicolumn{1}{c}{0.3827}  & \multicolumn{1}{c}{0.4923} & \multicolumn{1}{c}{0.3892}     & \multicolumn{1}{c}{0.3611}  & \multicolumn{1}{c}{0.3855}  & \multicolumn{1}{c}{0.3997} \\
\multicolumn{1}{c}{CultureLLM}    & \multicolumn{1}{c}{0.5047} & \multicolumn{1}{c}{0.3398}  & \multicolumn{1}{c}{0.3631}  & \multicolumn{1}{c}{0.4214}  & \multicolumn{1}{c}{0.3925}  & \multicolumn{1}{c}{0.5019} & \multicolumn{1}{c}{0.3956}     & \multicolumn{1}{c}{0.3752}  & \multicolumn{1}{c}{0.3899}  & \multicolumn{1}{c}{0.4093} \\ \bottomrule           
\end{tabular}
}
\end{table}

%% file: tables/tb-llama-forget.tex
\begin{table}[]
\caption{Results on forgetting experiments on Llama-2-70b}
\label{tb-llama-forget}
\centering
\begin{tabular}{lll}
\toprule
\multicolumn{1}{c}{Model}               & \multicolumn{1}{c}{GSM8K} & \multicolumn{1}{c}{BBH}  \\ \midrule
\multicolumn{1}{c}{Llama-2-70b}        & \multicolumn{1}{c}{56.8}  & \multicolumn{1}{c}{51.2} \\
\multicolumn{1}{c}{CultureLLM-Ar}  & \multicolumn{1}{c}{56.9}  & \multicolumn{1}{c}{53.2} \\
\multicolumn{1}{c}{CultureLLM-Bn} & \multicolumn{1}{c}{56.8}  & \multicolumn{1}{c}{52.1} \\
\multicolumn{1}{c}{CultureLLM-Zh} & \multicolumn{1}{c}{56.8}  & \multicolumn{1}{c}{56.6} \\
\multicolumn{1}{c}{CultureLLM-En} & \multicolumn{1}{c}{56.8}  & \multicolumn{1}{c}{48.8} \\
\multicolumn{1}{c}{CultureLLM-De} & \multicolumn{1}{c}{56.9}  & \multicolumn{1}{c}{51.8} \\
\multicolumn{1}{c}{CultureLLM-Ko} & \multicolumn{1}{c}{56.8}  & \multicolumn{1}{c}{53.6} \\
\multicolumn{1}{c}{CultureLLM-Pt} & \multicolumn{1}{c}{56.8}  & \multicolumn{1}{c}{55.1} \\
\multicolumn{1}{c}{CultureLLM-Es} & \multicolumn{1}{c}{56.7}  & \multicolumn{1}{c}{53.2} \\
\multicolumn{1}{c}{CultureLLM-Tr} & \multicolumn{1}{c}{56.8}  & \multicolumn{1}{c}{55.5} \\
\multicolumn{1}{c}{Avg}            & \multicolumn{1}{c}{56.8}  & \multicolumn{1}{c}{53.3} \\ \bottomrule         
\end{tabular}
\end{table}

%% file: tables/tb-human-participant.tex
\begin{table}[htbp]

\caption{Information on participants in human study}
\label{tb-human-participant}
\centering
\begin{tabular}{c|cc|cc}
\toprule
\multicolumn{1}{c|}{Gender}    & \multicolumn{1}{c}{Male}     & \multicolumn{1}{c|}{25} & \multicolumn{1}{c}{Female} & \multicolumn{1}{c}{25} \\ \midrule
\multicolumn{1}{c|}{Education} & \multicolumn{1}{c}{Bachelor} & \multicolumn{1}{c|}{26} & \multicolumn{1}{c}{Master} & \multicolumn{1}{c}{24} \\ \midrule
\multicolumn{1}{c|}{Age}       & \multicolumn{1}{c}{22}       & \multicolumn{1}{c}{11} &                            &                        \\
                              & \multicolumn{1}{c}{23}       & \multicolumn{1}{c}{15} &                            &                        \\
                              & \multicolumn{1}{c}{24}       & \multicolumn{1}{c}{13} &                            &                        \\
                              & \multicolumn{1}{c}{25}       & \multicolumn{1}{c}{9}  &                            &                        \\
                              & \multicolumn{1}{c}{26}       & \multicolumn{1}{c}{2}  &                            &                        \\ \bottomrule
\end{tabular}
\end{table}

%% file: checklist.tex
\section*{NeurIPS Paper Checklist}

The checklist is designed to encourage best practices for responsible machine learning research, addressing issues of reproducibility, transparency, research ethics, and societal impact. Do not remove the checklist: {\bf The papers not including the checklist will be desk rejected.} The checklist should follow the references and precede the (optional) supplemental material.  The checklist does NOT count towards the page
limit. 

Please read the checklist guidelines carefully for information on how to answer these questions. For each question in the checklist:
\begin{itemize}
    \item You should answer \answerYes{}, \answerNo{}, or \answerNA{}.
    \item \answerNA{} means either that the question is Not Applicable for that particular paper or the relevant information is Not Available.
    \item Please provide a short (1–2 sentence) justification right after your answer (even for NA). 
\end{itemize}

{\bf The checklist answers are an integral part of your paper submission.} They are visible to the reviewers, area chairs, senior area chairs, and ethics reviewers. You will be asked to also include it (after eventual revisions) with the final version of your paper, and its final version will be published with the paper.

The reviewers of your paper will be asked to use the checklist as one of the factors in their evaluation. While "\answerYes{}" is generally preferable to "\answerNo{}", it is perfectly acceptable to answer "\answerNo{}" provided a proper justification is given (e.g., "error bars are not reported because it would be too computationally expensive" or "we were unable to find the license for the dataset we used"). In general, answering "\answerNo{}" or "\answerNA{}" is not grounds for rejection. While the questions are phrased in a binary way, we acknowledge that the true answer is often more nuanced, so please just use your best judgment and write a justification to elaborate. All supporting evidence can appear either in the main paper or the supplemental material, provided in appendix. If you answer \answerYes{} to a question, in the justification please point to the section(s) where related material for the question can be found.

IMPORTANT, please:
\begin{itemize}
    \item {\bf Delete this instruction block, but keep the section heading ``NeurIPS paper checklist"},
    \item  {\bf Keep the checklist subsection headings, questions/answers and guidelines below.}
    \item {\bf Do not modify the questions and only use the provided macros for your answers}.
\end{itemize}


\begin{enumerate}

\item {\bf Claims}
    \item[] Question: Do the main claims made in the abstract and introduction accurately reflect the paper's contributions and scope?
    \item[] Answer: \answerYes{} 
    \item[] Justification: Yes, they do.
    \item[] Guidelines:
    \begin{itemize}
        \item The answer NA means that the abstract and introduction do not include the claims made in the paper.
        \item The abstract and/or introduction should clearly state the claims made, including the contributions made in the paper and important assumptions and limitations. A No or NA answer to this question will not be perceived well by the reviewers. 
        \item The claims made should match theoretical and experimental results, and reflect how much the results can be expected to generalize to other settings. 
        \item It is fine to include aspirational goals as motivation as long as it is clear that these goals are not attained by the paper. 
    \end{itemize}

\item {\bf Limitations}
    \item[] Question: Does the paper discuss the limitations of the work performed by the authors?
    \item[] Answer: \answerYes{} 
    \item[] Justification: We discuss the limitations in \Cref{sec-conclusion}.
    \item[] Guidelines:
    \begin{itemize}
        \item The answer NA means that the paper has no limitation while the answer No means that the paper has limitations, but those are not discussed in the paper. 
        \item The authors are encouraged to create a separate "Limitations" section in their paper.
        \item The paper should point out any strong assumptions and how robust the results are to violations of these assumptions (e.g., independence assumptions, noiseless settings, model well-specification, asymptotic approximations only holding locally). The authors should reflect on how these assumptions might be violated in practice and what the implications would be.
        \item The authors should reflect on the scope of the claims made, e.g., if the approach was only tested on a few datasets or with a few runs. In general, empirical results often depend on implicit assumptions, which should be articulated.
        \item The authors should reflect on the factors that influence the performance of the approach. For example, a facial recognition algorithm may perform poorly when image resolution is low or images are taken in low lighting. Or a speech-to-text system might not be used reliably to provide closed captions for online lectures because it fails to handle technical jargon.
        \item The authors should discuss the computational efficiency of the proposed algorithms and how they scale with dataset size.
        \item If applicable, the authors should discuss possible limitations of their approach to address problems of privacy and fairness.
        \item While the authors might fear that complete honesty about limitations might be used by reviewers as grounds for rejection, a worse outcome might be that reviewers discover limitations that aren't acknowledged in the paper. The authors should use their best judgment and recognize that individual actions in favor of transparency play an important role in developing norms that preserve the integrity of the community. Reviewers will be specifically instructed to not penalize honesty concerning limitations.
    \end{itemize}

\item {\bf Theory Assumptions and Proofs}
    \item[] Question: For each theoretical result, does the paper provide the full set of assumptions and a complete (and correct) proof?
    \item[] Answer: \answerYes{} 
    \item[] Justification: We conduct experiments on $60$ downstream tasks and verify the effectiveness of CultureLLMs. The results are shown in \Cref{sec-exp}
    \item[] Guidelines:
    \begin{itemize}
        \item The answer NA means that the paper does not include theoretical results. 
        \item All the theorems, formulas, and proofs in the paper should be numbered and cross-referenced.
        \item All assumptions should be clearly stated or referenced in the statement of any theorems.
        \item The proofs can either appear in the main paper or the supplemental material, but if they appear in the supplemental material, the authors are encouraged to provide a short proof sketch to provide intuition. 
        \item Inversely, any informal proof provided in the core of the paper should be complemented by formal proofs provided in appendix or supplemental material.
        \item Theorems and Lemmas that the proof relies upon should be properly referenced. 
    \end{itemize}

    \item {\bf Experimental Result Reproducibility}
    \item[] Question: Does the paper fully disclose all the information needed to reproduce the main experimental results of the paper to the extent that it affects the main claims and/or conclusions of the paper (regardless of whether the code and data are provided or not)?
    \item[] Answer: \answerYes{} 
    \item[] Justification: We discuss the setup and details in \Cref{sec-exp}.
    \item[] Guidelines:
    \begin{itemize}
        \item The answer NA means that the paper does not include experiments.
        \item If the paper includes experiments, a No answer to this question will not be perceived well by the reviewers: Making the paper reproducible is important, regardless of whether the code and data are provided or not.
        \item If the contribution is a dataset and/or model, the authors should describe the steps taken to make their results reproducible or verifiable. 
        \item Depending on the contribution, reproducibility can be accomplished in various ways. For example, if the contribution is a novel architecture, describing the architecture fully might suffice, or if the contribution is a specific model and empirical evaluation, it may be necessary to either make it possible for others to replicate the model with the same dataset, or provide access to the model. In general. releasing code and data is often one good way to accomplish this, but reproducibility can also be provided via detailed instructions for how to replicate the results, access to a hosted model (e.g., in the case of a large language model), releasing of a model checkpoint, or other means that are appropriate to the research performed.
        \item While NeurIPS does not require releasing code, the conference does require all submissions to provide some reasonable avenue for reproducibility, which may depend on the nature of the contribution. For example
        \begin{enumerate}
            \item If the contribution is primarily a new algorithm, the paper should make it clear how to reproduce that algorithm.
            \item If the contribution is primarily a new model architecture, the paper should describe the architecture clearly and fully.
            \item If the contribution is a new model (e.g., a large language model), then there should either be a way to access this model for reproducing the results or a way to reproduce the model (e.g., with an open-source dataset or instructions for how to construct the dataset).
            \item We recognize that reproducibility may be tricky in some cases, in which case authors are welcome to describe the particular way they provide for reproducibility. In the case of closed-source models, it may be that access to the model is limited in some way (e.g., to registered users), but it should be possible for other researchers to have some path to reproducing or verifying the results.
        \end{enumerate}
    \end{itemize}

\item {\bf Open access to data and code}
    \item[] Question: Does the paper provide open access to the data and code, with sufficient instructions to faithfully reproduce the main experimental results, as described in supplemental material?
    \item[] Answer: \answerYes{} 
    \item[] Justification: Our code is released in https://anonymous.4open.science/r/CultureLLM-DEE0.
    \item[] Guidelines:
    \begin{itemize}
        \item The answer NA means that paper does not include experiments requiring code.
        \item Please see the NeurIPS code and data submission guidelines (\url{https://nips.cc/public/guides/CodeSubmissionPolicy}) for more details.
        \item While we encourage the release of code and data, we understand that this might not be possible, so “No” is an acceptable answer. Papers cannot be rejected simply for not including code, unless this is central to the contribution (e.g., for a new open-source benchmark).
        \item The instructions should contain the exact command and environment needed to run to reproduce the results. See the NeurIPS code and data submission guidelines (\url{https://nips.cc/public/guides/CodeSubmissionPolicy}) for more details.
        \item The authors should provide instructions on data access and preparation, including how to access the raw data, preprocessed data, intermediate data, and generated data, etc.
        \item The authors should provide scripts to reproduce all experimental results for the new proposed method and baselines. If only a subset of experiments are reproducible, they should state which ones are omitted from the script and why.
        \item At submission time, to preserve anonymity, the authors should release anonymized versions (if applicable).
        \item Providing as much information as possible in supplemental material (appended to the paper) is recommended, but including URLs to data and code is permitted.
    \end{itemize}

\item {\bf Experimental Setting/Details}
    \item[] Question: Does the paper specify all the training and test details (e.g., data splits, hyperparameters, how they were chosen, type of optimizer, etc.) necessary to understand the results?
    \item[] Answer: \answerYes{} 
    \item[] Justification: We specify all the details in \Cref{setup}.
    \item[] Guidelines:
    \begin{itemize}
        \item The answer NA means that the paper does not include experiments.
        \item The experimental setting should be presented in the core of the paper to a level of detail that is necessary to appreciate the results and make sense of them.
        \item The full details can be provided either with the code, in appendix, or as supplemental material.
    \end{itemize}

\item {\bf Experiment Statistical Significance}
    \item[] Question: Does the paper report error bars suitably and correctly defined or other appropriate information about the statistical significance of the experiments?
    \item[] Answer: \answerYes{} 
    \item[] Justification: We report the statistical analysis in \Cref{fig-main-ablation}.
    \item[] Guidelines:
    \begin{itemize}
        \item The answer NA means that the paper does not include experiments.
        \item The authors should answer "Yes" if the results are accompanied by error bars, confidence intervals, or statistical significance tests, at least for the experiments that support the main claims of the paper.
        \item The factors of variability that the error bars are capturing should be clearly stated (for example, train/test split, initialization, random drawing of some parameter, or overall run with given experimental conditions).
        \item The method for calculating the error bars should be explained (closed form formula, call to a library function, bootstrap, etc.)
        \item The assumptions made should be given (e.g., Normally distributed errors).
        \item It should be clear whether the error bar is the standard deviation or the standard error of the mean.
        \item It is OK to report 1-sigma error bars, but one should state it. The authors should preferably report a 2-sigma error bar than state that they have a 96\% CI, if the hypothesis of Normality of errors is not verified.
        \item For asymmetric distributions, the authors should be careful not to show in tables or figures symmetric error bars that would yield results that are out of range (e.g. negative error rates).
        \item If error bars are reported in tables or plots, The authors should explain in the text how they were calculated and reference the corresponding figures or tables in the text.
    \end{itemize}

\item {\bf Experiments Compute Resources}
    \item[] Question: For each experiment, does the paper provide sufficient information on the computer resources (type of compute workers, memory, time of execution) needed to reproduce the experiments?
    \item[] Answer: \answerYes{} 
    \item[] Justification: We provide those information in \Cref{setup,sec-dis-llama}.
    \item[] Guidelines:
    \begin{itemize}
        \item The answer NA means that the paper does not include experiments.
        \item The paper should indicate the type of compute workers CPU or GPU, internal cluster, or cloud provider, including relevant memory and storage.
        \item The paper should provide the amount of compute required for each of the individual experimental runs as well as estimate the total compute. 
        \item The paper should disclose whether the full research project required more compute than the experiments reported in the paper (e.g., preliminary or failed experiments that didn't make it into the paper). 
    \end{itemize}
    
\item {\bf Code Of Ethics}
    \item[] Question: Does the research conducted in the paper conform, in every respect, with the NeurIPS Code of Ethics \url{https://neurips.cc/public/EthicsGuidelines}?
    \item[] Answer: \answerYes{} 
    \item[] Justification: We conform to the NeurIPS Code of Ethics in every respect.
    \item[] Guidelines:
    \begin{itemize}
        \item The answer NA means that the authors have not reviewed the NeurIPS Code of Ethics.
        \item If the authors answer No, they should explain the special circumstances that require a deviation from the Code of Ethics.
        \item The authors should make sure to preserve anonymity (e.g., if there is a special consideration due to laws or regulations in their jurisdiction).
    \end{itemize}

\item {\bf Broader Impacts}
    \item[] Question: Does the paper discuss both potential positive societal impacts and negative societal impacts of the work performed?
    \item[] Answer: \answerYes{} 
    \item[] Justification: We discuss in \Cref{sec-impact}.
    \item[] Guidelines:
    \begin{itemize}
        \item The answer NA means that there is no societal impact of the work performed.
        \item If the authors answer NA or No, they should explain why their work has no societal impact or why the paper does not address societal impact.
        \item Examples of negative societal impacts include potential malicious or unintended uses (e.g., disinformation, generating fake profiles, surveillance), fairness considerations (e.g., deployment of technologies that could make decisions that unfairly impact specific groups), privacy considerations, and security considerations.
        \item The conference expects that many papers will be foundational research and not tied to particular applications, let alone deployments. However, if there is a direct path to any negative applications, the authors should point it out. For example, it is legitimate to point out that an improvement in the quality of generative models could be used to generate deepfakes for disinformation. On the other hand, it is not needed to point out that a generic algorithm for optimizing neural networks could enable people to train models that generate Deepfakes faster.
        \item The authors should consider possible harms that could arise when the technology is being used as intended and functioning correctly, harms that could arise when the technology is being used as intended but gives incorrect results, and harms following from (intentional or unintentional) misuse of the technology.
        \item If there are negative societal impacts, the authors could also discuss possible mitigation strategies (e.g., gated release of models, providing defenses in addition to attacks, mechanisms for monitoring misuse, mechanisms to monitor how a system learns from feedback over time, improving the efficiency and accessibility of ML).
    \end{itemize}
    
\item {\bf Safeguards}
    \item[] Question: Does the paper describe safeguards that have been put in place for responsible release of data or models that have a high risk for misuse (e.g., pretrained language models, image generators, or scraped datasets)?
    \item[] Answer: \answerYes{} 
    \item[] Justification: We check the data manually to make sure the safety of data.
    \item[] Guidelines:
    \begin{itemize}
        \item The answer NA means that the paper poses no such risks.
        \item Released models that have a high risk for misuse or dual-use should be released with necessary safeguards to allow for controlled use of the model, for example by requiring that users adhere to usage guidelines or restrictions to access the model or implementing safety filters. 
        \item Datasets that have been scraped from the Internet could pose safety risks. The authors should describe how they avoided releasing unsafe images.
        \item We recognize that providing effective safeguards is challenging, and many papers do not require this, but we encourage authors to take this into account and make a best faith effort.
    \end{itemize}

\item {\bf Licenses for existing assets}
    \item[] Question: Are the creators or original owners of assets (e.g., code, data, models), used in the paper, properly credited and are the license and terms of use explicitly mentioned and properly respected?
    \item[] Answer: \answerYes{} 
    \item[] Justification: We properly credited the the creators or original owners of assets and the license and terms of use explicitly are mentioned and properly respected.
    \item[] Guidelines:
    \begin{itemize}
        \item The answer NA means that the paper does not use existing assets.
        \item The authors should cite the original paper that produced the code package or dataset.
        \item The authors should state which version of the asset is used and, if possible, include a URL.
        \item The name of the license (e.g., CC-BY 4.0) should be included for each asset.
        \item For scraped data from a particular source (e.g., website), the copyright and terms of service of that source should be provided.
        \item If assets are released, the license, copyright information, and terms of use in the package should be provided. For popular datasets, \url{paperswithcode.com/datasets} has curated licenses for some datasets. Their licensing guide can help determine the license of a dataset.
        \item For existing datasets that are re-packaged, both the original license and the license of the derived asset (if it has changed) should be provided.
        \item If this information is not available online, the authors are encouraged to reach out to the asset's creators.
    \end{itemize}

\item {\bf New Assets}
    \item[] Question: Are new assets introduced in the paper well documented and is the documentation provided alongside the assets?
    \item[] Answer: \answerYes{} 
    \item[] Justification: Our code and data are released. And it is well documented in README.md.
    \item[] Guidelines:
    \begin{itemize}
        \item The answer NA means that the paper does not release new assets.
        \item Researchers should communicate the details of the dataset/code/model as part of their submissions via structured templates. This includes details about training, license, limitations, etc. 
        \item The paper should discuss whether and how consent was obtained from people whose asset is used.
        \item At submission time, remember to anonymize your assets (if applicable). You can either create an anonymized URL or include an anonymized zip file.
    \end{itemize}

\item {\bf Crowdsourcing and Research with Human Subjects}
    \item[] Question: For crowdsourcing experiments and research with human subjects, does the paper include the full text of instructions given to participants and screenshots, if applicable, as well as details about compensation (if any)? 
    \item[] Answer: \answerYes{} 
    \item[] Justification: The details are described in \Cref{sec-append-human}.
    \item[] Guidelines:
    \begin{itemize}
        \item The answer NA means that the paper does not involve crowdsourcing nor research with human subjects.
        \item Including this information in the supplemental material is fine, but if the main contribution of the paper involves human subjects, then as much detail as possible should be included in the main paper. 
        \item According to the NeurIPS Code of Ethics, workers involved in data collection, curation, or other labor should be paid at least the minimum wage in the country of the data collector. 
    \end{itemize}

\item {\bf Institutional Review Board (IRB) Approvals or Equivalent for Research with Human Subjects}
    \item[] Question: Does the paper describe potential risks incurred by study participants, whether such risks were disclosed to the subjects, and whether Institutional Review Board (IRB) approvals (or an equivalent approval/review based on the requirements of your country or institution) were obtained?
    \item[] Answer: \answerYes{} 
    \item[] Justification: The details are described in \Cref{sec-append-human}.
    \item[] Guidelines:
    \begin{itemize}
        \item The answer NA means that the paper does not involve crowdsourcing nor research with human subjects.
        \item Depending on the country in which research is conducted, IRB approval (or equivalent) may be required for any human subjects research. If you obtained IRB approval, you should clearly state this in the paper. 
        \item We recognize that the procedures for this may vary significantly between institutions and locations, and we expect authors to adhere to the NeurIPS Code of Ethics and the guidelines for their institution. 
        \item For initial submissions, do not include any information that would break anonymity (if applicable), such as the institution conducting the review.
    \end{itemize}

\end{enumerate}

%% file: neurips_2024.bbl
\begin{thebibliography}{93}
\providecommand{\natexlab}[1]{#1}
\providecommand{\url}[1]{\texttt{#1}}
\expandafter\ifx\csname urlstyle\endcsname\relax
  \providecommand{\doi}[1]{doi: #1}\else
  \providecommand{\doi}{doi: \begingroup \urlstyle{rm}\Url}\fi

\bibitem[mis(2019)]{misc_turkish_spam_v01_530}
{Turkish Spam V01}.
\newblock UCI Machine Learning Repository, 2019.
\newblock {DOI}: https://doi.org/10.24432/C5WG7F.

\bibitem[Abbasi et~al.(2023)Abbasi, Ghafouri, Firouzmandi, Naderi, and Bidgoli]{abbasi2023persianllama}
Mohammad~Amin Abbasi, Arash Ghafouri, Mahdi Firouzmandi, Hassan Naderi, and Behrouz~Minaei Bidgoli.
\newblock Persianllama: Towards building first persian large language model.
\newblock \emph{arXiv preprint arXiv:2312.15713}, 2023.

\bibitem[Ahuja et~al.(2023)Ahuja, Hada, Ochieng, Jain, Diddee, Maina, Ganu, Segal, Axmed, Bali, et~al.]{ahuja2023mega}
Kabir Ahuja, Rishav Hada, Millicent Ochieng, Prachi Jain, Harshita Diddee, Samuel Maina, Tanuja Ganu, Sameer Segal, Maxamed Axmed, Kalika Bali, et~al.
\newblock Mega: Multilingual evaluation of generative ai.
\newblock \emph{arXiv preprint arXiv:2303.12528}, 2023.

\bibitem[aimansnigdha(2018)]{Bangla-Abusive-Comment-Dataset}
aimansnigdha.
\newblock Bangla-abusive-comment-dataset.
\newblock https://github.com/aimansnigdha/Bangla-Abusive-Comment-Dataset, 2018.

\bibitem[{\'A}lvarez-Carmona et~al.(2018){\'A}lvarez-Carmona, Guzm{\'a}n-Falc{\'o}n, Montes-y G{\'o}mez, Escalante, Villasenor-Pineda, Reyes-Meza, and Rico-Sulayes]{alvarez2018overview}
Miguel~{\'A} {\'A}lvarez-Carmona, Estefan{\i}a Guzm{\'a}n-Falc{\'o}n, Manuel Montes-y G{\'o}mez, Hugo~Jair Escalante, Luis Villasenor-Pineda, Ver{\'o}nica Reyes-Meza, and Antonio Rico-Sulayes.
\newblock Overview of mex-a3t at ibereval 2018: Authorship and aggressiveness analysis in mexican spanish tweets.
\newblock In \emph{Notebook papers of 3rd sepln workshop on evaluation of human language technologies for iberian languages (ibereval), seville, spain}, volume~6, 2018.

\bibitem[Appadurai(1996)]{appadurai1996modernity}
Arjun Appadurai.
\newblock Modernity at large: Cultural dimensions of globalization.
\newblock \emph{U of Minnesota P}, 1996.

\bibitem[Basile et~al.(2019)Basile, Bosco, Fersini, Nozza, Patti, Pardo, Rosso, and Sanguinetti]{basile2019semeval}
Valerio Basile, Cristina Bosco, Elisabetta Fersini, Debora Nozza, Viviana Patti, Francisco Manuel~Rangel Pardo, Paolo Rosso, and Manuela Sanguinetti.
\newblock Semeval-2019 task 5: Multilingual detection of hate speech against immigrants and women in twitter.
\newblock In \emph{Proceedings of the 13th international workshop on semantic evaluation}, pages 54--63, 2019.

\bibitem[Bhattacharya et~al.(2020)Bhattacharya, Singh, Kumar, Bansal, Bhagat, Dawer, Lahiri, and Ojha]{trac2-dataset}
Shiladitya Bhattacharya, Siddharth Singh, Ritesh Kumar, Akanksha Bansal, Akash Bhagat, Yogesh Dawer, Bornini Lahiri, and Atul~Kr. Ojha.
\newblock Developing a multilingual annotated corpus of misogyny and aggression.
\newblock In \emph{Proceedings of the Second Workshop on Trolling, Aggression and Cyberbullying}, pages 158--168, Marseille, France, May 2020. European Language Resources Association (ELRA).
\newblock URL \url{https://aclanthology.org/2020.trac-1.25/}.

\bibitem[Bilmes(2022)]{bilmes2022submodularity}
Jeff Bilmes.
\newblock Submodularity in machine learning and artificial intelligence.
\newblock \emph{arXiv preprint arXiv:2202.00132}, 2022.

\bibitem[Cao et~al.(2023)Cao, Zhou, Lee, Cabello, Chen, and Hershcovich]{cao2023assessing}
Y~Cao, L~Zhou, S~Lee, L~Cabello, M~Chen, and D~Hershcovich.
\newblock Assessing cross-cultural alignment between chatgpt and human societies: An empirical study. arxiv.
\newblock \emph{Preprint posted online on March}, 31, 2023.

\bibitem[Caselli et~al.(2020)Caselli, Basile, Mitrovi{\'c}, Kartoziya, and Granitzer]{caselli2020feel}
Tommaso Caselli, Valerio Basile, Jelena Mitrovi{\'c}, Inga Kartoziya, and Michael Granitzer.
\newblock I feel offended, don’t be abusive! implicit/explicit messages in offensive and abusive language.
\newblock In \emph{Proceedings of the Twelfth Language Resources and Evaluation Conference}, pages 6193--6202, 2020.

\bibitem[\c{C}\"{o}ltekin(2020)]{coltekin2020lrec}
\c{C}a\u{g}r{\i} \c{C}\"{o}ltekin.
\newblock A corpus of turkish offensive language on social media.
\newblock In \emph{Proceedings of The 12th Language Resources and Evaluation Conference}, pages 6174--6184, Marseille, France, 2020.
\newblock URL \url{https://www.aclweb.org/anthology/2020.lrec-1.758}.

\bibitem[Chan et~al.(2023)Chan, Garc{\'\i}a, Silvestri, O'Donnel, and Palla]{chan2023harmonizing}
Alex~J Chan, Jos{\'e} Luis~Redondo Garc{\'\i}a, Fabrizio Silvestri, Colm O'Donnel, and Konstantina Palla.
\newblock Harmonizing global voices: Culturally-aware models for enhanced content moderation.
\newblock \emph{arXiv preprint arXiv:2312.02401}, 2023.

\bibitem[Chen et~al.(2024)Chen, Deng, Yuan, Ji, and Gu]{chen2024self}
Zixiang Chen, Yihe Deng, Huizhuo Yuan, Kaixuan Ji, and Quanquan Gu.
\newblock Self-play fine-tuning converts weak language models to strong language models.
\newblock \emph{arXiv preprint arXiv:2401.01335}, 2024.

\bibitem[Chowdhury et~al.(2020)Chowdhury, Mubarak, Abdelali, Jung, Jansen, and Salminen]{chowdhury2020multi}
Shammur~Absar Chowdhury, Hamdy Mubarak, Ahmed Abdelali, Soon-gyo Jung, Bernard~J Jansen, and Joni Salminen.
\newblock A multi-platform arabic news comment dataset for offensive language detection.
\newblock In \emph{Proceedings of the Twelfth Language Resources and Evaluation Conference}, pages 6203--6212, 2020.

\bibitem[Cobbe et~al.(2021)Cobbe, Kosaraju, Bavarian, Chen, Jun, Kaiser, Plappert, Tworek, Hilton, Nakano, et~al.]{cobbe2021training}
Karl Cobbe, Vineet Kosaraju, Mohammad Bavarian, Mark Chen, Heewoo Jun, Lukasz Kaiser, Matthias Plappert, Jerry Tworek, Jacob Hilton, Reiichiro Nakano, et~al.
\newblock Training verifiers to solve math word problems.
\newblock \emph{arXiv preprint arXiv:2110.14168}, 2021.

\bibitem[daanVeer(2020)]{Korean-HateSpeech-dataset}
daanVeer.
\newblock Korean hatespeech dataset.
\newblock https://github.com/daanVeer/HateSpeech\_dataset, 2020.

\bibitem[Davidson et~al.(2017)Davidson, Warmsley, Macy, and Weber]{hateoffensive}
Thomas Davidson, Dana Warmsley, Michael Macy, and Ingmar Weber.
\newblock Automated hate speech detection and the problem of offensive language.
\newblock In \emph{Proceedings of the 11th International AAAI Conference on Web and Social Media}, ICWSM '17, pages 512--515, 2017.

\bibitem[de~Paula and Schlicht(2021)]{de2021ai}
Angel Felipe~Magnossao de~Paula and Ipek~Baris Schlicht.
\newblock Ai-upv at iberlef-2021 detoxis task: Toxicity detection in immigration-related web news comments using transformers and statistical models.
\newblock \emph{arXiv preprint arXiv:2111.04530}, 2021.

\bibitem[de~Pelle and Moreira(2017)]{Pelle2017}
Rogers~P. de~Pelle and Viviane~P. Moreira.
\newblock Offensive comments in the brazilian web: a dataset and baseline results.
\newblock 2017.

\bibitem[Delanoy(2020{\natexlab{a}})]{delanoy2020culture}
Werner Delanoy.
\newblock What is culture.
\newblock \emph{The Cambridge handbook of intercultural communication}, pages 17--34, 2020{\natexlab{a}}.

\bibitem[Delanoy(2020{\natexlab{b}})]{delanoy_2020}
Werner Delanoy.
\newblock \emph{What Is Culture?}, page 17–34.
\newblock Cambridge Handbooks in Language and Linguistics. Cambridge University Press, 2020{\natexlab{b}}.
\newblock \doi{10.1017/9781108555067.003}.

\bibitem[Fazio and Zanna(1981)]{fazio1981direct}
Russell~H Fazio and Mark~P Zanna.
\newblock Direct experience and attitude-behavior consistency.
\newblock In \emph{Advances in experimental social psychology}, volume~14, pages 161--202. Elsevier, 1981.

\bibitem[Fersini et~al.(2018)Fersini, Rosso, Anzovino, et~al.]{fersini2018overview}
Elisabetta Fersini, Paolo Rosso, Maria Anzovino, et~al.
\newblock Overview of the task on automatic misogyny identification at ibereval 2018.
\newblock \emph{Ibereval@ sepln}, 2150:\penalty0 214--228, 2018.

\bibitem[Fung et~al.(2022)Fung, Chakraborty, Guo, Rambow, Muresan, and Ji]{fung2022normsage}
Yi~R Fung, Tuhin Chakraborty, Hao Guo, Owen Rambow, Smaranda Muresan, and Heng Ji.
\newblock Normsage: Multi-lingual multi-cultural norm discovery from conversations on-the-fly.
\newblock \emph{arXiv preprint arXiv:2210.08604}, 2022.

\bibitem[Gao et~al.(2020)Gao, Biderman, Black, Golding, Hoppe, Foster, Phang, He, Thite, Nabeshima, et~al.]{gao2020pile}
Leo Gao, Stella Biderman, Sid Black, Laurence Golding, Travis Hoppe, Charles Foster, Jason Phang, Horace He, Anish Thite, Noa Nabeshima, et~al.
\newblock The pile: An 800gb dataset of diverse text for language modeling.
\newblock \emph{arXiv preprint arXiv:2101.00027}, 2020.

\bibitem[Google(2023)]{gemini}
Google.
\newblock Gemini.
\newblock https://deepmind.google/technologies/gemini/\#introduction, 2023.

\bibitem[HASOC(2020)]{HASOC2020}
HASOC.
\newblock Hasoc2020.
\newblock https://hasocfire.github.io/hasoc/2020/index.html, 2020.

\bibitem[Hofstede(1984)]{hofstede1984culture}
Geert Hofstede.
\newblock \emph{Culture's consequences: International differences in work-related values}, volume~5.
\newblock sage, 1984.

\bibitem[Hu et~al.(2021)Hu, Shen, Wallis, Allen-Zhu, Li, Wang, Wang, and Chen]{hu2021lora}
Edward~J Hu, Yelong Shen, Phillip Wallis, Zeyuan Allen-Zhu, Yuanzhi Li, Shean Wang, Lu~Wang, and Weizhu Chen.
\newblock Lora: Low-rank adaptation of large language models.
\newblock \emph{arXiv preprint arXiv:2106.09685}, 2021.

\bibitem[Husain(2020)]{husain2005osact4}
F~Husain.
\newblock Osact4 shared task on offensive language detection: Intensive preprocessing-based approach. arxiv 2020.
\newblock \emph{arXiv preprint arXiv:2005.07297}, 2020.

\bibitem[Ji et~al.(2023)Ji, Qiu, Chen, Zhang, Lou, Wang, Duan, He, Zhou, Zhang, et~al.]{ji2023ai}
Jiaming Ji, Tianyi Qiu, Boyuan Chen, Borong Zhang, Hantao Lou, Kaile Wang, Yawen Duan, Zhonghao He, Jiayi Zhou, Zhaowei Zhang, et~al.
\newblock Ai alignment: A comprehensive survey.
\newblock \emph{arXiv preprint arXiv:2310.19852}, 2023.

\bibitem[Jiang et~al.(2019)Jiang, Gao, He, Kang, Sun, Zhang, Si, and Liu]{jiang2019detect}
Zhuoren Jiang, Zhe Gao, Guoxiu He, Yangyang Kang, Changlong Sun, Qiong Zhang, Luo Si, and Xiaozhong Liu.
\newblock Detect camouflaged spam content via stoneskipping: Graph and text joint embedding for chinese character variation representation.
\newblock In \emph{Proceedings of the 2019 Conference on Empirical Methods in Natural Language Processing (EMNLP2019)}. ACM, 2019.

\bibitem[Jin et~al.(2023)Jin, Chandra, Verma, Hu, De~Choudhury, and Kumar]{jin2023better}
Yiqiao Jin, Mohit Chandra, Gaurav Verma, Yibo Hu, Munmun De~Choudhury, and Srijan Kumar.
\newblock Better to ask in english: Cross-lingual evaluation of large language models for healthcare queries.
\newblock \emph{arXiv e-prints}, pages arXiv--2310, 2023.

\bibitem[Johnson et~al.(2022)Johnson, Pistilli, Men{\'e}dez-Gonz{\'a}lez, Duran, Panai, Kalpokiene, and Bertulfo]{johnsonghost}
Rebecca~L Johnson, Giada Pistilli, Natalia Men{\'e}dez-Gonz{\'a}lez, Leslye Denisse~Dias Duran, Enrico Panai, Julija Kalpokiene, and Donald~Jay Bertulfo.
\newblock The ghost in the machine has an american accent: value conflict in gpt-3.
\newblock \emph{arXiv preprint arXiv:2203.07785}, 2022.

\bibitem[Kaddoura and Henno(2024)]{kaddoura2024dataset}
Sanaa Kaddoura and Safaa Henno.
\newblock Dataset of arabic spam and ham tweets.
\newblock \emph{Data in Brief}, 52\penalty0 (10990):\penalty0 4, 2024.

\bibitem[Kaggle(2019)]{Jigsaw-Multilingual-Toxicity}
Kaggle.
\newblock Jigsaw-multilingual-toxicity.
\newblock https://www.kaggle.com/code/tarunpaparaju/jigsaw-multilingual-toxicity-eda-models, 2019.

\bibitem[Kaggle(2021)]{turkish-tweets}
Kaggle.
\newblock 5k turkish tweets with incivil content.
\newblock https://www.kaggle.com/datasets/kbulutozler/5k-turkish-tweets-with-incivil-content, 2021.

\bibitem[Kaggle(2022)]{turkish-offensive-language-detection}
Kaggle.
\newblock turkish offensive language detection.
\newblock https://www.kaggle.com/datasets/toygarr/turkish-offensive-language-detection, 2022.

\bibitem[Karayi{\u{g}}it et~al.(2021)Karayi{\u{g}}it, Ac{\i}, and Akda{\u{g}}l{\i}]{karayiugit2021detecting}
Habibe Karayi{\u{g}}it, {\c{C}}i{\u{g}}dem~{\.I}nan Ac{\i}, and Ali Akda{\u{g}}l{\i}.
\newblock Detecting abusive instagram comments in turkish using convolutional neural network and machine learning methods.
\newblock \emph{Expert Systems with Applications}, 174:\penalty0 114802, 2021.

\bibitem[Kova{\v{c}} et~al.(2023)Kova{\v{c}}, Sawayama, Portelas, Colas, Dominey, and Oudeyer]{kovavc2023large}
Grgur Kova{\v{c}}, Masataka Sawayama, R{\'e}my Portelas, C{\'e}dric Colas, Peter~Ford Dominey, and Pierre-Yves Oudeyer.
\newblock Large language models as superpositions of cultural perspectives.
\newblock \emph{arXiv preprint arXiv:2307.07870}, 2023.

\bibitem[Lee et~al.(2022)Lee, Lim, Lee, Jo, Kim, Yoon, and Han]{lee-etal-2022-k}
Jean Lee, Taejun Lim, Heejun Lee, Bogeun Jo, Yangsok Kim, Heegeun Yoon, and Soyeon~Caren Han.
\newblock K-{MH}a{S}: A multi-label hate speech detection dataset in {K}orean online news comment.
\newblock In \emph{Proceedings of the 29th International Conference on Computational Linguistics}, pages 3530--3538, Gyeongju, Republic of Korea, October 2022. International Committee on Computational Linguistics.
\newblock URL \url{https://aclanthology.org/2022.coling-1.311}.

\bibitem[Leite et~al.(2020)Leite, Silva, Bontcheva, and Scarton]{leite2020toxic}
Joao~A Leite, Diego~F Silva, Kalina Bontcheva, and Carolina Scarton.
\newblock Toxic language detection in social media for brazilian portuguese: New dataset and multilingual analysis.
\newblock \emph{arXiv preprint arXiv:2010.04543}, 2020.

\bibitem[Lewis et~al.(2020)Lewis, Perez, Piktus, Petroni, Karpukhin, Goyal, K{\"u}ttler, Lewis, Yih, Rockt{\"a}schel, et~al.]{lewis2020retrieval}
Patrick Lewis, Ethan Perez, Aleksandra Piktus, Fabio Petroni, Vladimir Karpukhin, Naman Goyal, Heinrich K{\"u}ttler, Mike Lewis, Wen-tau Yih, Tim Rockt{\"a}schel, et~al.
\newblock Retrieval-augmented generation for knowledge-intensive nlp tasks.
\newblock \emph{Advances in Neural Information Processing Systems}, 33:\penalty0 9459--9474, 2020.

\bibitem[Li et~al.(2023)Li, Yu, Zhou, Schick, Zettlemoyer, Levy, Weston, and Lewis]{li2023self}
Xian Li, Ping Yu, Chunting Zhou, Timo Schick, Luke Zettlemoyer, Omer Levy, Jason Weston, and Mike Lewis.
\newblock Self-alignment with instruction backtranslation.
\newblock \emph{arXiv preprint arXiv:2308.06259}, 2023.

\bibitem[Lin et~al.(2024)Lin, Ji, Tiedemann, Martins, and Schütze]{lin2024mala500}
Peiqin Lin, Shaoxiong Ji, Jörg Tiedemann, André F.~T. Martins, and Hinrich Schütze.
\newblock Mala-500: Massive language adaptation of large language models, 2024.

\bibitem[Lin and Chen(2023)]{lin2023taiwan}
Yen-Ting Lin and Yun-Nung Chen.
\newblock Taiwan llm: Bridging the linguistic divide with a culturally aligned language model.
\newblock \emph{arXiv preprint arXiv:2311.17487}, 2023.

\bibitem[Liu et~al.(2023{\natexlab{a}})Liu, Bubeck, Eldan, Kulkarni, Li, Nguyen, Ward, and Zhang]{liu2023tinygsm}
Bingbin Liu, Sebastien Bubeck, Ronen Eldan, Janardhan Kulkarni, Yuanzhi Li, Anh Nguyen, Rachel Ward, and Yi~Zhang.
\newblock Tinygsm: achieving> 80\% on gsm8k with small language models.
\newblock \emph{arXiv preprint arXiv:2312.09241}, 2023{\natexlab{a}}.

\bibitem[Liu et~al.(2023{\natexlab{b}})Liu, Koto, Baldwin, and Gurevych]{liu2023multilingual}
Chen~Cecilia Liu, Fajri Koto, Timothy Baldwin, and Iryna Gurevych.
\newblock Are multilingual llms culturally-diverse reasoners? an investigation into multicultural proverbs and sayings.
\newblock \emph{arXiv preprint arXiv:2309.08591}, 2023{\natexlab{b}}.

\bibitem[Loper and Bird(2002)]{loper2002nltk}
Edward Loper and Steven Bird.
\newblock Nltk: The natural language toolkit.
\newblock \emph{arXiv preprint cs/0205028}, 2002.

\bibitem[Marion et~al.(2023)Marion, {\"U}st{\"u}n, Pozzobon, Wang, Fadaee, and Hooker]{marion2023less}
Max Marion, Ahmet {\"U}st{\"u}n, Luiza Pozzobon, Alex Wang, Marzieh Fadaee, and Sara Hooker.
\newblock When less is more: Investigating data pruning for pretraining llms at scale.
\newblock \emph{arXiv preprint arXiv:2309.04564}, 2023.

\bibitem[Masoud et~al.(2023)Masoud, Liu, Ferianc, Treleaven, and Rodrigues]{masoud2023cultural}
Reem~I Masoud, Ziquan Liu, Martin Ferianc, Philip Treleaven, and Miguel Rodrigues.
\newblock Cultural alignment in large language models: An explanatory analysis based on hofstede's cultural dimensions.
\newblock \emph{arXiv preprint arXiv:2309.12342}, 2023.

\bibitem[Moon et~al.(2020)Moon, Cho, and Lee]{moon-etal-2020-beep}
Jihyung Moon, Won~Ik Cho, and Junbum Lee.
\newblock {BEEP}! {K}orean corpus of online news comments for toxic speech detection.
\newblock In \emph{Proceedings of the Eighth International Workshop on Natural Language Processing for Social Media}, pages 25--31, Online, July 2020. Association for Computational Linguistics.
\newblock URL \url{https://www.aclweb.org/anthology/2020.socialnlp-1.4}.

\bibitem[Moussa{\"\i}d et~al.(2013)Moussa{\"\i}d, K{\"a}mmer, Analytis, and Neth]{moussaid2013social}
Mehdi Moussa{\"\i}d, Juliane~E K{\"a}mmer, Pantelis~P Analytis, and Hansj{\"o}rg Neth.
\newblock Social influence and the collective dynamics of opinion formation.
\newblock \emph{PloS one}, 8\penalty0 (11):\penalty0 e78433, 2013.

\bibitem[Mubarak et~al.(2022)Mubarak, Al-Khalifa, and Al-Thubaity]{mubarak2022overview}
Hamdy Mubarak, Hend Al-Khalifa, and AbdulMohsen Al-Thubaity.
\newblock Overview of osact5 shared task on arabic offensive language and hate speech detection.
\newblock In \emph{Proceedinsg of the 5th Workshop on Open-Source Arabic Corpora and Processing Tools with Shared Tasks on Qur'an QA and Fine-Grained Hate Speech Detection}, pages 162--166, 2022.

\bibitem[Myung et~al.(2024)Myung, Lee, Zhou, Jin, Putri, Antypas, Borkakoty, Kim, Perez-Almendros, Ayele, et~al.]{myung2024blend}
Junho Myung, Nayeon Lee, Yi~Zhou, Jiho Jin, Rifki~Afina Putri, Dimosthenis Antypas, Hsuvas Borkakoty, Eunsu Kim, Carla Perez-Almendros, Abinew~Ali Ayele, et~al.
\newblock Blend: A benchmark for llms on everyday knowledge in diverse cultures and languages.
\newblock \emph{arXiv preprint arXiv:2406.09948}, 2024.

\bibitem[Naous et~al.(2023)Naous, Ryan, and Xu]{naous2023having}
Tarek Naous, Michael~J Ryan, and Wei Xu.
\newblock Having beer after prayer? measuring cultural bias in large language models.
\newblock \emph{arXiv preprint arXiv:2305.14456}, 2023.

\bibitem[Nguyen et~al.(2023{\natexlab{a}})Nguyen, Razniewski, Varde, and Weikum]{nguyen2023extracting}
Tuan-Phong Nguyen, Simon Razniewski, Aparna Varde, and Gerhard Weikum.
\newblock Extracting cultural commonsense knowledge at scale.
\newblock In \emph{Proceedings of the ACM Web Conference 2023}, pages 1907--1917, 2023{\natexlab{a}}.

\bibitem[Nguyen et~al.(2023{\natexlab{b}})Nguyen, Zhang, Li, Aljunied, Tan, Cheng, Chen, Deng, Yang, Liu, et~al.]{nguyen2023seallms}
Xuan-Phi Nguyen, Wenxuan Zhang, Xin Li, Mahani Aljunied, Qingyu Tan, Liying Cheng, Guanzheng Chen, Yue Deng, Sen Yang, Chaoqun Liu, et~al.
\newblock Seallms--large language models for southeast asia.
\newblock \emph{arXiv preprint arXiv:2312.00738}, 2023{\natexlab{b}}.

\bibitem[Niszczota and Janczak(2023)]{niszczota2023large}
Pawe{\l} Niszczota and Mateusz Janczak.
\newblock Large language models can replicate cross-cultural differences in personality.
\newblock \emph{arXiv preprint arXiv:2310.10679}, 2023.

\bibitem[OpenAI(2023{\natexlab{a}})]{chatgpt}
OpenAI.
\newblock Chatgpt.
\newblock https://chat.openai.com/, 2023{\natexlab{a}}.

\bibitem[OpenAI(2023{\natexlab{b}})]{openai2023gpt4}
OpenAI.
\newblock Gpt-4 technical report, 2023{\natexlab{b}}.

\bibitem[Ousidhoum et~al.(2019)Ousidhoum, Lin, Zhang, Song, and Yeung]{ousidhoum-etal-multilingual-hate-speech-2019}
Nedjma Ousidhoum, Zizheng Lin, Hongming Zhang, Yangqiu Song, and Dit-Yan Yeung.
\newblock Multilingual and multi-aspect hate speech analysis.
\newblock In \emph{Proceedings of EMNLP}. Association for Computational Linguistics, 2019.

\bibitem[Pereira-Kohatsu et~al.(2019)Pereira-Kohatsu, Quijano-S{\'a}nchez, Liberatore, and Camacho-Collados]{pereira2019detecting}
Juan~Carlos Pereira-Kohatsu, Lara Quijano-S{\'a}nchez, Federico Liberatore, and Miguel Camacho-Collados.
\newblock Detecting and monitoring hate speech in twitter.
\newblock \emph{Sensors}, 19\penalty0 (21):\penalty0 4654, 2019.

\bibitem[Pipatanakul et~al.(2023)Pipatanakul, Jirabovonvisut, Manakul, Sripaisarnmongkol, Patomwong, Chokchainant, and Tharnpipitchai]{pipatanakul2023typhoon}
Kunat Pipatanakul, Phatrasek Jirabovonvisut, Potsawee Manakul, Sittipong Sripaisarnmongkol, Ruangsak Patomwong, Pathomporn Chokchainant, and Kasima Tharnpipitchai.
\newblock Typhoon: Thai large language models.
\newblock \emph{arXiv preprint arXiv:2312.13951}, 2023.

\bibitem[Pitenis et~al.(2020)Pitenis, Zampieri, and Ranasinghe]{pitenis2020offensive}
Zeses Pitenis, Marcos Zampieri, and Tharindu Ranasinghe.
\newblock Offensive language identification in greek.
\newblock \emph{arXiv preprint arXiv:2003.07459}, 2020.

\bibitem[Plaza-del Arco et~al.(2021)Plaza-del Arco, Montejo-R{\'a}ez, Lopez, and Mart{\'\i}n-Valdivia]{plaza2021offendes}
Flor~Miriam Plaza-del Arco, Arturo Montejo-R{\'a}ez, L~Alfonso~Urena Lopez, and Mar{\'\i}a-Teresa Mart{\'\i}n-Valdivia.
\newblock Offendes: A new corpus in spanish for offensive language research.
\newblock In \emph{Proceedings of the International Conference on Recent Advances in Natural Language Processing (RANLP 2021)}, pages 1096--1108, 2021.

\bibitem[Rao et~al.(2023)Rao, Khandelwal, Tanmay, Agarwal, and Choudhury]{rao2023ethical}
Abhinav Rao, Aditi Khandelwal, Kumar Tanmay, Utkarsh Agarwal, and Monojit Choudhury.
\newblock Ethical reasoning over moral alignment: A case and framework for in-context ethical policies in llms.
\newblock \emph{arXiv preprint arXiv:2310.07251}, 2023.

\bibitem[Romim et~al.(2021)Romim, Ahmed, Talukder, and Saiful~Islam]{romim2021hate}
Nauros Romim, Mosahed Ahmed, Hriteshwar Talukder, and Md~Saiful~Islam.
\newblock Hate speech detection in the bengali language: A dataset and its baseline evaluation.
\newblock In \emph{Proceedings of International Joint Conference on Advances in Computational Intelligence: IJCACI 2020}, pages 457--468. Springer, 2021.

\bibitem[Rosenthal et~al.(2020)Rosenthal, Atanasova, Karadzhov, Zampieri, and Nakov]{rosenthal2020solid}
Sara Rosenthal, Pepa Atanasova, Georgi Karadzhov, Marcos Zampieri, and Preslav Nakov.
\newblock Solid: A large-scale semi-supervised dataset for offensive language identification.
\newblock \emph{arXiv preprint arXiv:2004.14454}, 2020.

\bibitem[Ross et~al.(2016)Ross, Rist, Carbonell, Cabrera, Kurowsky, and Wojatzki]{ross2016hatespeech}
Bj{\"o}rn Ross, Michael Rist, Guillermo Carbonell, Benjamin Cabrera, Nils Kurowsky, and Michael Wojatzki.
\newblock {Measuring the Reliability of Hate Speech Annotations: The Case of the European Refugee Crisis}.
\newblock In Michael Bei{\ss}wenger, Michael Wojatzki, and Torsten Zesch, editors, \emph{Proceedings of NLP4CMC III: 3rd Workshop on Natural Language Processing for Computer-Mediated Communication}, volume~17 of \emph{Bochumer Linguistische Arbeitsberichte}, pages 6--9, Bochum, sep 2016.

\bibitem[R{\"o}ttger et~al.(2022)R{\"o}ttger, Seelawi, Nozza, Talat, and Vidgen]{rottger2022multilingual}
Paul R{\"o}ttger, Haitham Seelawi, Debora Nozza, Zeerak Talat, and Bertie Vidgen.
\newblock Multilingual hatecheck: Functional tests for multilingual hate speech detection models.
\newblock \emph{arXiv preprint arXiv:2206.09917}, 2022.

\bibitem[Sharif and Hoque(2022)]{sharif2022tackling}
Omar Sharif and Mohammed~Moshiul Hoque.
\newblock Tackling cyber-aggression: Identification and fine-grained categorization of aggressive texts on social media using weighted ensemble of transformers.
\newblock \emph{Neurocomputing}, 490:\penalty0 462--481, 2022.

\bibitem[Shen et~al.(2023)Shen, Jin, Huang, Liu, Dong, Guo, Wu, Liu, and Xiong]{shen2023large}
Tianhao Shen, Renren Jin, Yufei Huang, Chuang Liu, Weilong Dong, Zishan Guo, Xinwei Wu, Yan Liu, and Deyi Xiong.
\newblock Large language model alignment: A survey.
\newblock \emph{arXiv preprint arXiv:2309.15025}, 2023.

\bibitem[Shi et~al.(2024)Shi, Li, Zhang, Ziems, Horesh, de~Paula, Yang, et~al.]{shi2024culturebank}
Weiyan Shi, Ryan Li, Yutong Zhang, Caleb Ziems, Raya Horesh, Rog{\'e}rio~Abreu de~Paula, Diyi Yang, et~al.
\newblock Culturebank: An online community-driven knowledge base towards culturally aware language technologies.
\newblock \emph{arXiv preprint arXiv:2404.15238}, 2024.

\bibitem[Song et~al.(2021)Song, Ryu, Lee, and Park]{song2021large}
Hoyun Song, Soo~Hyun Ryu, Huije Lee, and Jong~C Park.
\newblock A large-scale comprehensive abusiveness detection dataset with multifaceted labels from reddit.
\newblock In \emph{Proceedings of the 25th Conference on Computational Natural Language Learning}, pages 552--561, 2021.

\bibitem[Survey(2022)]{wvs}
World~Values Survey.
\newblock World values survey.
\newblock https://www.worldvaluessurvey.org/wvs.jsp, 2022.

\bibitem[Suzgun et~al.(2022)Suzgun, Scales, Sch{\"a}rli, Gehrmann, Tay, Chung, Chowdhery, Le, Chi, Zhou, et~al.]{suzgun2022challenging}
Mirac Suzgun, Nathan Scales, Nathanael Sch{\"a}rli, Sebastian Gehrmann, Yi~Tay, Hyung~Won Chung, Aakanksha Chowdhery, Quoc~V Le, Ed~H Chi, Denny Zhou, et~al.
\newblock Challenging big-bench tasks and whether chain-of-thought can solve them.
\newblock \emph{arXiv preprint arXiv:2210.09261}, 2022.

\bibitem[Touvron et~al.(2023)Touvron, Martin, Stone, Albert, Almahairi, Babaei, Bashlykov, Batra, Bhargava, Bhosale, et~al.]{touvron2023llama}
Hugo Touvron, Louis Martin, Kevin Stone, Peter Albert, Amjad Almahairi, Yasmine Babaei, Nikolay Bashlykov, Soumya Batra, Prajjwal Bhargava, Shruti Bhosale, et~al.
\newblock Llama 2: Open foundation and fine-tuned chat models, 2023.
\newblock \emph{URL https://arxiv. org/abs/2307.09288}, 2023.

\bibitem[Vargas et~al.(2022)Vargas, Carvalho, Rodrigues~de G{\'o}es, Pardo, and Benevenuto]{vargas-etal-2022-hatebr}
Francielle Vargas, Isabelle Carvalho, Fabiana Rodrigues~de G{\'o}es, Thiago Pardo, and Fabr{\'\i}cio Benevenuto.
\newblock {H}ate{BR}: A large expert annotated corpus of {B}razilian {I}nstagram comments for offensive language and hate speech detection.
\newblock In \emph{Proceedings of the Thirteenth Language Resources and Evaluation Conference}, pages 7174--7183, Marseille, France, June 2022. European Language Resources Association.
\newblock URL \url{https://aclanthology.org/2022.lrec-1.777}.

\bibitem[Walters and Wilder(2023)]{walters2023fabrication}
William~H Walters and Esther~Isabelle Wilder.
\newblock Fabrication and errors in the bibliographic citations generated by chatgpt.
\newblock \emph{Scientific Reports}, 13\penalty0 (1):\penalty0 14045, 2023.

\bibitem[Wang et~al.(2023{\natexlab{a}})Wang, Yang, Huang, Yang, Majumder, and Wei]{wang2023improving}
Liang Wang, Nan Yang, Xiaolong Huang, Linjun Yang, Rangan Majumder, and Furu Wei.
\newblock Improving text embeddings with large language models.
\newblock \emph{arXiv preprint arXiv:2401.00368}, 2023{\natexlab{a}}.

\bibitem[Wang et~al.(2023{\natexlab{b}})Wang, Li, Chen, Song, Lin, Cao, Liu, and Sui]{wang2023making}
Peiyi Wang, Lei Li, Liang Chen, Feifan Song, Binghuai Lin, Yunbo Cao, Tianyu Liu, and Zhifang Sui.
\newblock Making large language models better reasoners with alignment.
\newblock \emph{arXiv preprint arXiv:2309.02144}, 2023{\natexlab{b}}.

\bibitem[Wang et~al.(2023{\natexlab{c}})Wang, Zhou, and Sachan]{wang2023let}
Ruida Wang, Wangchunshu Zhou, and Mrinmaya Sachan.
\newblock Let's synthesize step by step: Iterative dataset synthesis with large language models by extrapolating errors from small models.
\newblock \emph{arXiv preprint arXiv:2310.13671}, 2023{\natexlab{c}}.

\bibitem[Wang et~al.(2023{\natexlab{d}})Wang, Jiao, Huang, Dai, Huang, Tu, and Lyu]{wang2023not}
Wenxuan Wang, Wenxiang Jiao, Jingyuan Huang, Ruyi Dai, Jen-tse Huang, Zhaopeng Tu, and Michael~R Lyu.
\newblock Not all countries celebrate thanksgiving: On the cultural dominance in large language models.
\newblock \emph{arXiv preprint arXiv:2310.12481}, 2023{\natexlab{d}}.

\bibitem[Waseem and Hovy(2016)]{waseem2016hateful}
Zeerak Waseem and Dirk Hovy.
\newblock Hateful symbols or hateful people? predictive features for hate speech detection on twitter.
\newblock In \emph{Proceedings of the NAACL student research workshop}, pages 88--93, 2016.

\bibitem[Wiegand et~al.(2018)Wiegand, Siegel, and Ruppenhofer]{wiegand2018overview}
Michael Wiegand, Melanie Siegel, and Josef Ruppenhofer.
\newblock Overview of the germeval 2018 shared task on the identification of offensive language.
\newblock 2018.

\bibitem[Xu et~al.(2023)Xu, Liu, Yan, Xu, Si, Zhou, Yi, Gao, Sang, Zhang, Zhang, Peng, Huang, and Zhou]{xu2023cvalues}
Guohai Xu, Jiayi Liu, Ming Yan, Haotian Xu, Jinghui Si, Zhuoran Zhou, Peng Yi, Xing Gao, Jitao Sang, Rong Zhang, Ji~Zhang, Chao Peng, Fei Huang, and Jingren Zhou.
\newblock Cvalues: Measuring the values of chinese large language models from safety to responsibility.
\newblock \emph{arXiv 2307.09705}, 2023.

\bibitem[Yao et~al.(2023)Yao, Yi, Wang, Wang, and Xie]{yao2023instructions}
Jing Yao, Xiaoyuan Yi, Xiting Wang, Jindong Wang, and Xing Xie.
\newblock From instructions to intrinsic human values--a survey of alignment goals for big models.
\newblock \emph{arXiv preprint arXiv:2308.12014}, 2023.

\bibitem[Yu et~al.(2023)Yu, Jiang, Shi, Yu, Liu, Zhang, Kwok, Li, Weller, and Liu]{yu2023metamath}
Longhui Yu, Weisen Jiang, Han Shi, Jincheng Yu, Zhengying Liu, Yu~Zhang, James~T Kwok, Zhenguo Li, Adrian Weller, and Weiyang Liu.
\newblock Metamath: Bootstrap your own mathematical questions for large language models.
\newblock \emph{arXiv preprint arXiv:2309.12284}, 2023.

\bibitem[Zampieri et~al.(2020)Zampieri, Nakov, Rosenthal, Atanasova, Karadzhov, Mubarak, Derczynski, Pitenis, and {\c{C}}{\"o}ltekin]{zampieri2020semeval}
Marcos Zampieri, Preslav Nakov, Sara Rosenthal, Pepa Atanasova, Georgi Karadzhov, Hamdy Mubarak, Leon Derczynski, Zeses Pitenis, and {\c{C}}a{\u{g}}r{\i} {\c{C}}{\"o}ltekin.
\newblock Semeval-2020 task 12: Multilingual offensive language identification in social media (offenseval 2020).
\newblock \emph{arXiv preprint arXiv:2006.07235}, 2020.

\bibitem[Zhou et~al.(2022)Zhou, Deng, Mi, Li, Wang, Huang, Jiang, Liu, and Meng]{zhou2022towards}
Jingyan Zhou, Jiawen Deng, Fei Mi, Yitong Li, Yasheng Wang, Minlie Huang, Xin Jiang, Qun Liu, and Helen Meng.
\newblock Towards identifying social bias in dialog systems: Frame, datasets, and benchmarks.
\newblock \emph{arXiv preprint arXiv:2202.08011}, 2022.

\bibitem[Zhu et~al.(2023)Zhu, Wang, Zhou, Wang, Chen, Wang, Yang, Ye, Zhang, Gong, et~al.]{zhu2023promptbench}
Kaijie Zhu, Jindong Wang, Jiaheng Zhou, Zichen Wang, Hao Chen, Yidong Wang, Linyi Yang, Wei Ye, Yue Zhang, Neil~Zhenqiang Gong, et~al.
\newblock Promptbench: Towards evaluating the robustness of large language models on adversarial prompts.
\newblock \emph{arXiv preprint arXiv:2306.04528}, 2023.

\end{thebibliography}
